\documentclass{article}

\PassOptionsToPackage{nameinlink}{cleveref}
\PassOptionsToPackage{table,xcdraw}{xcolor}

\usepackage{iclr2026_conference,times}
\iclrfinalcopy

\usepackage[table,xcdraw]{xcolor}
\definecolor{mydarkblue}{rgb}{0,0.08,0.45} %

\usepackage{adjustbox}
\usepackage{pgfplots}
\pgfplotsset{compat=1.17}
\pgfdeclarelayer{background}
\pgfsetlayers{background,main}
\usetikzlibrary{shapes.arrows}

\usepackage[nice]{nicefrac}
\usepackage{booktabs}
\usepackage{multirow}
\usepackage{algorithm}
\usepackage{algpseudocode}
\usepackage{wrapfig}
\usepackage{graphicx}
\usepackage{subcaption}
\usepackage{colortbl}
\usepackage{arydshln}
\usepackage{multirow}
\usepackage{bbding}
\usepackage{rotating} %

\newcommand{\val}[3]{$#1#2$\scalebox{.75}{${\pm}#3$}}

\usepgfplotslibrary{fillbetween}
\usepgfplotslibrary[groupplots]
\usepgflibrary {shadings}
\usetikzlibrary{backgrounds}

\usetikzlibrary{decorations.text}

\newlength{\figurewidth}
\newlength{\figureheight}

\usepackage{xspace}

\usepackage[pagebackref,breaklinks,colorlinks=true,allcolors=mydarkblue]{hyperref}

\usepackage{safecolours}

\colorlet{scBlue}{tol-colour1}
\colorlet{scRed}{tol-colour2}
\colorlet{scGreen}{tol-colour3}
\colorlet{scYellow}{tol-colour4}
\colorlet{scCyan}{tol-colour5}
\colorlet{scPurple}{tol-colour6}

\usepackage{tikz}

\usepackage{fontawesome5}

\usepackage{shortex-light} %

\newcommand{\todo}[1]{\emph{\color{scRed}#1}}

\newcommand{\trainset}{\mcD_{\text{train}}}
\newcommand{\testset}{\mcD_{\text{test}}}

\newcommand{\ntrain}{\ensuremath{n_{\text{train}}}}
\newcommand{\nval}{\ensuremath{n_{\text{val}}}}
\newcommand{\ntest}{\ensuremath{n_{\text{test}}}}
\newcommand{\nclass}{\ensuremath{c}}

\newcommand{\MAP}{\ensuremath{\operatorname{MAP}}}
\newcommand{\BALD}{\ensuremath{\operatorname{BALD}}}
\newcommand{\EPIG}{\ensuremath{\operatorname{EPIG}}}
\newcommand{\GGN}{\ensuremath{\operatorname{GGN}}}
\newcommand{\CE}{\ensuremath{\textsc{CE}}}
\newcommand{\img}{\ensuremath{\textsc{Img}}}
\newcommand{\txt}{\ensuremath{\textsc{Txt}}}
\newcommand{\enc}{\ensuremath{\textsc{Enc}}}
\newcommand{\imgtotxt}{\ensuremath{\img \rightarrow \txt}}
\newcommand{\txttoimg}{\ensuremath{\img \leftarrow \txt}}

\newcommand{\xstar}{\ensuremath{x^\star}}
\newcommand{\bxstar}{\ensuremath{\bx^\star}}
\newcommand{\ystar}{\ensuremath{y^\star}}
\newcommand{\bystar}{\ensuremath{\by^\star}}

\newcommand{\Var}{\mathbb{V}\mathrm{ar}}

\newcommand{\cossim}[2]{\ensuremath{\operatorname{S}_{\textsc{cos}}(#1, #2)}}

\newcommand{\softmax}[1]{\ensuremath{\operatorname{softmax}\left(#1\right)}}

\tikzset{
	oracle/.pic={
		\begin{scope}
			\clip (-0.4,-0.195) rectangle (0.4, 0.4);
			\fill[thick] (0, 0) circle (0.3);
		\end{scope}

		\fill[rounded corners=0.075cm] (-0.3, -0.36) rectangle (0.3, -0.255);
		\fill[rounded corners=0.075cm] (-0.225, -0.3) rectangle (0.225, -0.225);

		\begin{scope}[yshift=-0.2375cm, xshift=-0.17cm, scale=1.5]
			\fill[white] (0.18, 0.18) -- (0.195, 0.225) -- (0.24, 0.24) -- (0.195, 0.255) -- (0.18, 0.3) -- (0.165, 0.255) -- (0.12, 0.24) -- (0.165, 0.225) -- cycle;
		\end{scope}
	},
	robothead/.pic={

		\fill[thick, rounded corners=0.1cm] (-0.3,-0.3) rectangle ++(0.6,0.6) ;

		\fill[white, thick] (-0.16,0.1) circle (0.08);
		\fill[white, thick] (0.16,0.1) circle (0.08);
		\fill[thick] (-0.16,0.1) circle (0.04);
		\fill[thick] (0.16,0.1) circle (0.04);

		\fill[white, thick, rounded corners=0.04cm] (-0.16,-0.16) rectangle ++(0.32,0.1);
		\draw[thick] (-0.08,-0.16) -- (-0.08,-0.06);
		\draw[thick] (0,-0.16) -- (0,-0.06);
		\draw[thick] (0.08,-0.16) -- (0.08,-0.06);

		\draw[thick] (0,0.3) -- (0,0.5);
		\fill[thick] (0,0.5) circle (0.04);

		\fill[thick, rounded corners=0.04cm] (-0.4,0.1) rectangle ++(0.08,0.2);
		\fill[thick, rounded corners=0.04cm] (0.32,0.1) rectangle ++(0.08,0.2);
	},
	expert/.pic={

		\fill[thick, rounded corners=0.1cm] (-0.3,-0.3) rectangle ++(0.6,0.6) ;

		\fill[white, thick] (-0.16,0.1) circle (0.08);
		\fill[white, thick] (0.16,0.1) circle (0.08);
		\fill[thick] (-0.16,0.1) circle (0.04);
		\fill[thick] (0.16,0.1) circle (0.04);

		\draw[thick, rounded corners=0.03cm, black] (-0.26, 0.02) rectangle ++(0.18, 0.16); %
		\draw[thick, rounded corners=0.03cm, black] (0.08, 0.02) rectangle ++(0.18, 0.16); %
		\draw[thick, black] (-0.08,0.1) -- (0.08,0.1); %
		\draw[thick, black] (-0.34,0.2) -- (-0.26,0.1); %
		\draw[thick, black] (0.34,0.2) -- (0.26,0.1); %

		\fill[white, thick, rounded corners=0.04cm] (-0.16,-0.16) rectangle ++(0.32,0.1);
		\draw[thick] (-0.08,-0.16) -- (-0.08,-0.06);
		\draw[thick] (0,-0.16) -- (0,-0.06);
		\draw[thick] (0.08,-0.16) -- (0.08,-0.06);

		\draw[thick] (0,0.3) -- (0,0.5);
		\fill[thick] (0,0.5) circle (0.04);

		\fill[thick, rounded corners=0.04cm] (-0.4,0.1) rectangle ++(0.08,0.2);
		\fill[thick, rounded corners=0.04cm] (0.32,0.1) rectangle ++(0.08,0.2);
	},	
	cross/.pic = {
		\draw[rotate = 45] (-#1,0) -- (#1,0);
		\draw[rotate = 45] (0,-#1) -- (0, #1);
	},
	pics/gauss/.style args={#1/#2}{
		code = {
			\begin{scope}
				\clip (-0.5,0) rectangle (1.5,1.5);
				\fill[black!10] (-0.5,0) rectangle (1.5,1.5);
				\fill[thick, domain=-10:10, samples=50] plot[smooth] (\x, {1/(#2*sqrt(2*pi))*exp(-(\x-#1)^2/(2*#2^2))});
				\draw[] (-0.5,0) rectangle (1.5,1.5);
			\end{scope}
		}
	}
} 
\crefname{appendix}{App.}{Apps.}
\crefname{section}{Sec.}{Secs.}

\algnewcommand\algorithmicforeach{\textbf{for each}}
\algdef{S}[FOR]{ForEach}[1]{\algorithmicforeach\ #1\ \algorithmicdo}

\newcolumntype{o}{>{\columncolor{scCyan!30}}c}

\newcommand{\legendline}[1]{(\protect\tikz[baseline=-0.5ex]{\protect\draw[thick, #1] (0,0) -- (0.5,0);})}

\newcommand{\legendhist}[1]{\textcolor{#1}{\rule{1.5ex}{1.5ex}}}
\renewcommand{\paragraph}[1]{\medskip\noindent\textbf{#1}~~}

\newcommand{\highlight}[2]{\protect\tikz[baseline]\protect\node[fill=#1, rounded corners=2pt, inner sep=1pt, anchor=base, font=\strut]{\,#2\,};}

\usepackage{xspace}
\newcommand{\eg}{\textit{e.g.}\xspace}
\newcommand{\ie}{\textit{i.e.}\xspace}
\newcommand{\ia}{\textit{i.a.}\xspace}
\newcommand{\cf}{\textit{cf.}\xspace}
\newcommand{\etc}{\textit{etc.}\xspace}
\newcommand{\etal}{\textit{et~al.}\xspace}
\newcommand{\wrt}{w.r.t.\xspace}

\renewcommand{\paragraph}[1]{\textbf{#1}~~}

\title{Post-hoc\,Probabilistic\,Vision-Language\,Models}

\newcommand{\smallquad}{\hspace{0.5em}} 

\author{Anton Baumann$^{1,2,*}$ \quad Rui Li$^{3}$ \quad Marcus Klasson$^{8}$ \quad Santeri Mentu$^{3,4}$\\
\textbf{Shyamgopal Karthik}$^{1,2,5,6}$ \quad \textbf{Zeynep Akata}$^{1,2,6,7}$ \quad \textbf{Arno Solin}$^{3,4}$ \quad \textbf{Martin Trapp}$^{9,10}$\\[0.5em]
\small
$^1$Technical University of Munich \smallquad
$^2$Helmholtz Munich \smallquad
$^3$ELLIS Institute Finland \& Aalto University \smallquad \\
\small
$^4$Finnish Center for Artificial Intelligence \smallquad
$^5$University of Tübingen \smallquad
$^6$Munich Center for Machine Learning \\
\small
$^7$Munich Data Science Institute \smallquad
$^8$Ericsson Research \smallquad 
$^9$KTH Royal Institute of Technology \smallquad
$^{10}$Digital Futures
}

\begin{document}
\crefname{table}{Table}{Tables}
\crefname{appendix}{App.}{Apps.}
\maketitle

\begin{abstract}
    Vision-language models (VLMs), such as CLIP and SigLIP, have found remarkable success in classification, retrieval, and generative tasks.
	For this, VLMs deterministically map images and text descriptions to a joint latent space in which their similarity is assessed using the cosine similarity. 
	However, a deterministic mapping of inputs fails to capture uncertainties over concepts arising from domain shifts when used in downstream tasks. 
	In this work, we propose post-hoc uncertainty estimation in VLMs that does not require additional training.
	Our method leverages a Bayesian posterior approximation over the last layers in VLMs and analytically quantifies uncertainties over cosine similarities.
    We demonstrate its effectiveness for uncertainty quantification and support set selection in active learning.   
	Compared to baselines, we obtain improved and well-calibrated predictive uncertainties, interpretable uncertainty estimates, and sample-efficient active learning.
    Our results show promise for safety-critical applications of large-scale models.
\end{abstract}

\begin{center}
    \centering
    \captionsetup{type=figure}
\newcommand{\arcarrow}[8]%
{   \pgfmathsetmacro{\rin}{#1}
    \pgfmathsetmacro{\rmid}{#2}
    \pgfmathsetmacro{\rout}{#3}
    \pgfmathsetmacro{\astart}{#4}
    \pgfmathsetmacro{\aend}{#5}
    \pgfmathsetmacro{\atip}{#6}
    \fill[#7,draw=white,line width=1pt] (\astart:\rin) arc (\astart:\aend:\rin) -- (\aend+\atip:\rmid) -- (\aend:\rout) arc (\aend:\astart:\rout) -- (\astart+\atip:\rmid) -- cycle;
    \path[decoration={text along path, text={#8}, text align={align=center}, raise=-0.5ex},decorate] (\astart+\atip:{\rmid-.05}) arc (\astart+\atip:\aend+\atip:{\rmid-.05});
}
\pgfmathdeclarefunction{gauss}{2}{%
  \pgfmathparse{1/(#2*sqrt(2*pi))*exp(-((x-#1)^2)/(2*#2^2))}%
}
\resizebox{\textwidth}{!}{%
\begin{tikzpicture}[
        box/.style={minimum width=0.485\textwidth,minimum height=2.9cm,rounded corners=4pt,text width=3cm,align=center,fill=white}, 
        label/.style = {font={\footnotesize},anchor=north,inner sep=0pt,align=center}, 
    ]

  \node[box, minimum width=0.45\columnwidth, draw=black!30,fill=blue!03] (part1) at (0,0) {};
  \node[box,minimum width=0.38\columnwidth,draw=black!30,fill=blue!03] (part3) at (12cm,0) {};
  \node[box,minimum width=0.38\columnwidth,draw=black!30,fill=blue!03] (part2) at ($(part1.east)!.5!(part3.west)$) {};

  \node[label] (label1) at ([yshift=-0.1cm]part1.north) {{\bf Post-hoc Probabilistic Model} };
  \node[label] (label2) at ([yshift=-0.1cm]part2.north) {{\bf Uncertainty Quantification} };
  \node[label] (label3) at ([yshift=-0.1cm]part3.north) {{\bf Active Learning}};

  \node[
    draw=black!30,
    fill=black!30,
    single arrow,
    shape border rotate=0,
    minimum height=1cm,
    single arrow head extend=0.25cm,
    rounded corners=1pt
  ] at ($(part1.east)!.5!(part2.west)$) {};
  \node[
    draw=black!30,
    fill=black!30,
    single arrow,
    shape border rotate=0,
    minimum height=1cm,
    single arrow head extend=0.25cm,
    rounded corners=1pt
  ] at ($(part2.east)!.5!(part3.west)$) {};

  \pic[fill=gray, draw=gray, scale=1.5, transform shape] (vlm) at (-2, 0) {robothead};
  \node[align=center,font=\scriptsize] at (-2,-0.8) { Pre-trained VLM \\(CLIP, SigLIP)};

  \draw[->,thick] (-1.2,0) -- node[sloped,above,align=left]{\scriptsize Laplace approximation} (1.3,0);
  \pic[fill=scCyan, draw=scCyan, scale=1.5, transform shape] (bayes-vlm) at (2, 0) {expert};
  \node[align=center] (bayes) at (2,-0.7) {\scriptsize BayesVLM};
  \begin{scope}[xshift=6.5cm,yshift=0.5cm]
  \node[draw=scRed, thick, anchor=center,rounded corners=2pt,fill=white] (im0) at (-1.0,-.25) 
    {\includegraphics[width=1cm]{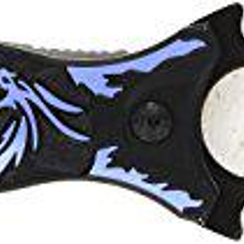}};
  \node[draw=scGreen, thick, anchor=center,rounded corners=2pt,fill=white] (im1) at (1.0,-.25) 
    {\includegraphics[width=1cm]{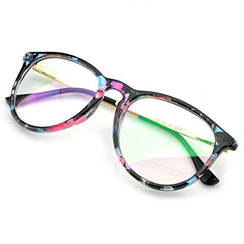}};
  \node[align=right,font=\scriptsize,below of=im0] {Glasses: 0.45 \\ Scissors: 0.55};
  \node[align=right,font=\scriptsize,below of=im1] {Glasses: 0.99 \\ Scissors: 0.01};
  \node[align=center,font=\scriptsize] at (-1, -1.75) {\bfseries \textcolor{scRed}{Uncertain}};
  \node[align=center,font=\scriptsize] at (1, -1.75) {\bfseries \textcolor{scGreen}{Certain}};
  
  \end{scope}

  \begin{scope}[xshift=12cm,yshift=0.5cm]
  
  \node[align=center,draw=scRed, thick, anchor=east, rounded corners=1pt, fill=white] 
    at (-1.25,0.2)
    {\includegraphics[width=0.5cm]{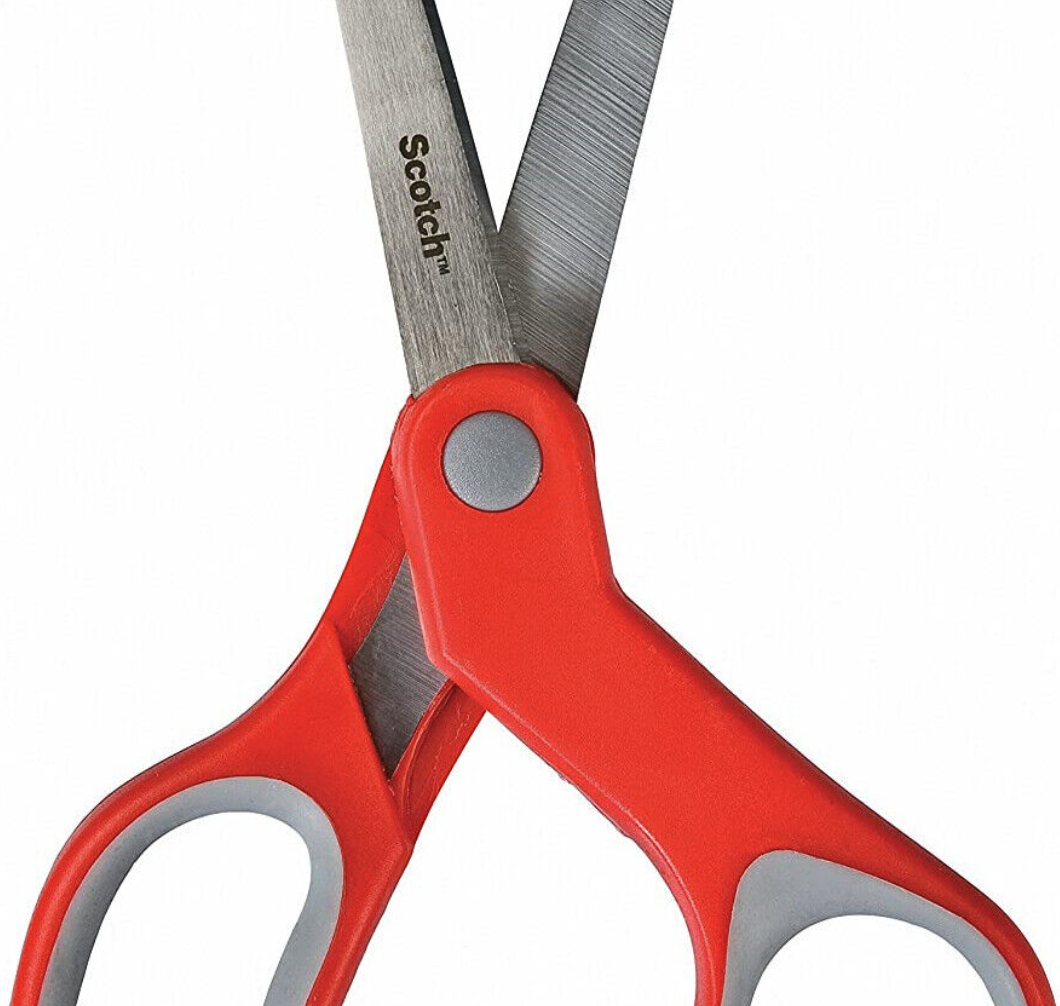}};
  \node[align=center,draw=scRed, thick, anchor=east, rounded corners=1pt, fill=white] 
    at (-1.25,-0.6)
    {\includegraphics[width=0.5cm]{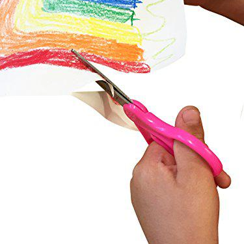}};
  \node[align=center,draw=scRed, thick, anchor=east, rounded corners=1pt, fill=white] 
    at (-1.25,-1.4)
    {\includegraphics[width=0.5cm]{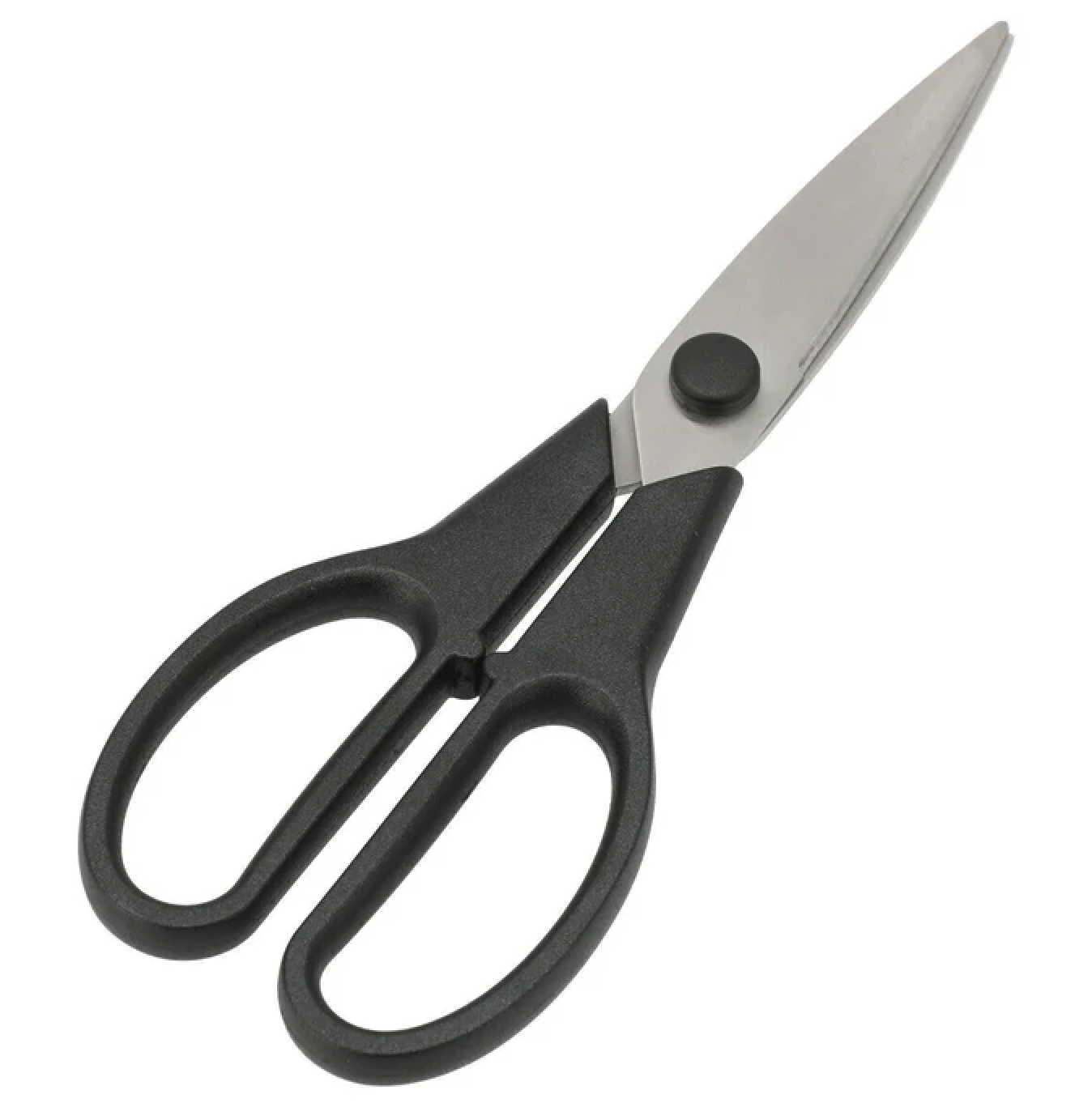}};;

  \draw[->, thick] (-1,-0.6) -- (0.5,-0.6) node[midway, above] {\scriptsize Update model} node[midway, below,align=center,font=\tiny] {Use {\bfseries uncertainty} to \\ select fine tuning data};

  \node[align=center,draw=scGreen, thick, anchor=center, fill=white, rounded corners=2pt] (im2) at (1.5,-.25) {\includegraphics[width=1cm]{pics/homeoffice-product_test_mutual_info_highest_3.png}};
  \node[align=right,font=\scriptsize,below of=im2]  {Glasses: 0.2 \\ Scissors: 0.8};
  \node[align=center,font=\scriptsize] at (1.5, -1.75) {\bfseries \textcolor{scGreen}{More certain now}};
  \end{scope}

\end{tikzpicture}}
    \vspace*{-1.3em} %
    \captionof{figure}{We introduce an efficient and effective post-hoc method to provide uncertainty estimates for vision-language models (\eg, CLIP, SigLIP) using a Laplace approximation. We demonstrate that uncertainty estimates derived from this approximation improve the calibration of these models on several zero-shot classification benchmarks (\cref{sec:results_uncertainties}) and are effective in active learning (\cref{sec:results_activelearning}).}
    \label{fig:teaser}
\end{center}%
\section{Introduction}

\label{sec:intro}

\renewcommand{\thefootnote}{} 
\footnotetext{$^{*}$Work partially done during an internship at Aalto University.}
\renewcommand{\thefootnote}{\arabic{footnote}} %

Pre-trained large-scale vision-language models (VLMs) \citep{bordes2024introduction,zhang2024vision}, such as CLIP~\citep{radford2021learning} and SigLIP~\citep{zhai2023sigmoid}, have achieved remarkable success in tasks like zero-shot classification, retrieval, and generation, driven by training on billion-scale data sets~\citep{gadre2024datacomp,schuhmann2022laion}. 
However, when employing large-scale machine learning models reliably in real-world settings and on downstream applications, we expect them not only to provide accurate predictions but also to enable us to quantify their predictive uncertainties.
Obtaining efficient and effective uncertainty estimates is particularly relevant for safety-critical applications, as well as when making decisions based on those estimates, such as in active learning.

Previous work on uncertainty quantification for VLMs has primarily focused on calibration~\citep{guo2017calibration,tu2023a}, test-time adaptation~\citep{ayhan2018test,farina2024frustratingly,yoon2024ctpt,lafon2025vilu}, fine-tuning~\citep{fort2021exploring,tu2023a,ju2025exploiting}, or training probabilistic VLMs from scratch~\citep{pcmepp,prolip}. However, each of those approaches has limitations regarding their applicability in real-world settings.
For example, calibration methods cannot capture epistemic uncertainties, adapter and retraining-based methods come with substantial computational demands and require retraining in streaming/active learning settings, and test-time adaptation methods significantly increase inference costs.

To manifest efficient and effective uncertainty quantification for the reliable application of VLMs, we identify the following desiderata:
The method should be applicable to any VLM architecture (\emph{model-agnostic}), and uncertainties should be obtained in a \emph{post-hoc} manner without retraining the model from scratch. 
During inference, it should have low to no computational overhead (\emph{efficient}) and capture relevant sources of uncertainties (\emph{effective}). 
Finally, the method should extract uncertainties from the original VLM without adding new layers or adapters that require training (\emph{training-free}).

The Bayesian framework provides a principled way to model epistemic and aleatoric uncertainties, and has shown promise as a `toolbox' for uncertainty quantification in deep learning \citep{papamarkou2024position}.
Consider \cref{fig:active_learning}, which shows results on the EuroSAT data set \citep{helber2019eurosat}, a land use and land cover classification task based on Sentinel-2 satellite images, for the popular OpenCLIP model (\legendhist{scPurple}).
We observe that the Bayesian counterpart (\legendhist{scCyan}) results in less overconfident predictions before active learning (compare quadrant \texttt{b} and \texttt{a}) and substantially reduces the error in the predictions after active learning, compared to the fine-tuned OpenCLIP model (\legendhist{scPurple}).
Much of the misclassification of the OpenCLIP model after active learning can be attributed to its overconfident behaviour before and after active learning, indicating the benefits of using a Bayesian formulation.

\setlength{\figurewidth}{.5\textwidth}
\setlength{\figureheight}{.5\textwidth}
\begin{figure}[t]
  \centering
\begin{minipage}{.49\textwidth}
  \centering
\begin{tikzpicture}
    \node (a) at (-0.4\figurewidth,0) {};
    \node (b) at (0.4\figurewidth,0) {};
    \draw[|-|] (a) -- node[above, midway] {\scriptsize Before Active Learning (Zero-shot)} (b);
\end{tikzpicture}\\
\begin{tikzpicture}
    \begin{axis}[
        height=3cm,
        width=1.05\figurewidth, 
        axis lines = middle,
        tick label style={font=\tiny},
        ytick=\empty,
        y axis line style={draw=none},
        ybar,
        bar width=3.4pt,
        axis background/.style={fill=scYellow!10},
    ]
        \addplot+[ybar, fill=scPurple, draw=white, fill opacity=0.5, draw opacity=0.7, very thin] table [x index=0, y index=1, col sep=space] {figs/marginal_entropy_hist_data/histogram_bins_before_map.dat};
        \addplot+[ybar, fill=scBlue, draw=white, fill opacity=0.5, draw opacity=0.7, very thin] table [x index=0, y index=1, col sep=space] {figs/marginal_entropy_hist_data/histogram_bins_before_la.dat};
        \node[fill=scYellow!10,fill opacity=0.8, rounded corners, inner sep=0pt] at (1.55,200) {\scriptsize \textbf{Entropy} (error $>0.5$) };
    \end{axis}
\end{tikzpicture}\\
\pgfplotsset{
      width=0.6\figurewidth,
      enlargelimits=false,
      tick label style={font=\tiny},
      axis equal image,
      xlabel={\scriptsize Uncertainty (Entropy)}, 
      xlabel near ticks,
      xlabel style={yshift=4pt},      
      ylabel near ticks,
      xtick={0, 0.5, 1}, 
      ytick={0, 0.5, 1},
      xticklabels={0, 1, 2},
      yticklabels={0, 0.5, 1},
      ytick pos=left,
      xtick pos=bottom,
      ytick align=outside,
      xtick align=outside, 
      xmin=0, xmax=1.05, ymin=0, ymax=1.05
}
\begin{tikzpicture}
  \begin{axis}[
          ylabel={\scriptsize Error},
    ]
    \addplot graphics[xmin=-0.21, xmax=1.20, ymin=-0.18, ymax=1.20] {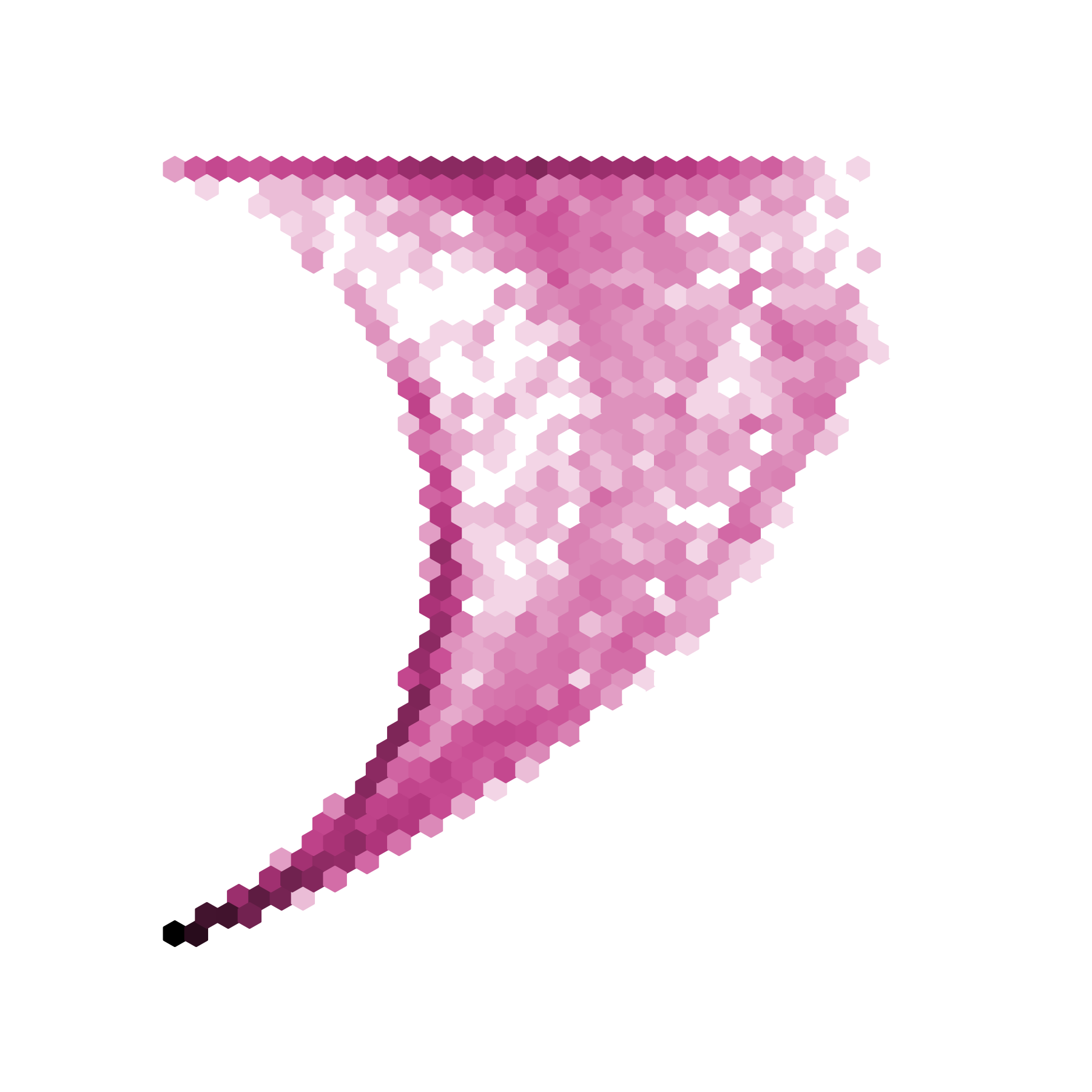};
    \fill[scYellow, fill opacity=0.1] (axis cs:0,1.05) rectangle (1.05,0.5);
    \draw[thin,gray] (axis cs:0,0.5) -- (axis cs:1.05,0.5);
    \draw[thin,gray] (axis cs:0.5,0) -- (axis cs:0.5,1.05);
    \node at (axis cs:0.1,0.75) {\scriptsize (b)};
    \node at (axis cs:0.96,0.75) {\scriptsize (a)};
    \node at (axis cs:0.1,0.25) {\scriptsize (c)};
    \node at (axis cs:0.96,0.25) {\scriptsize (d)};
  \end{axis}
\end{tikzpicture}
\begin{tikzpicture}
  \begin{axis}[
    ]
    \addplot graphics[xmin=-0.21, xmax=1.20, ymin=-0.18, ymax=1.20] {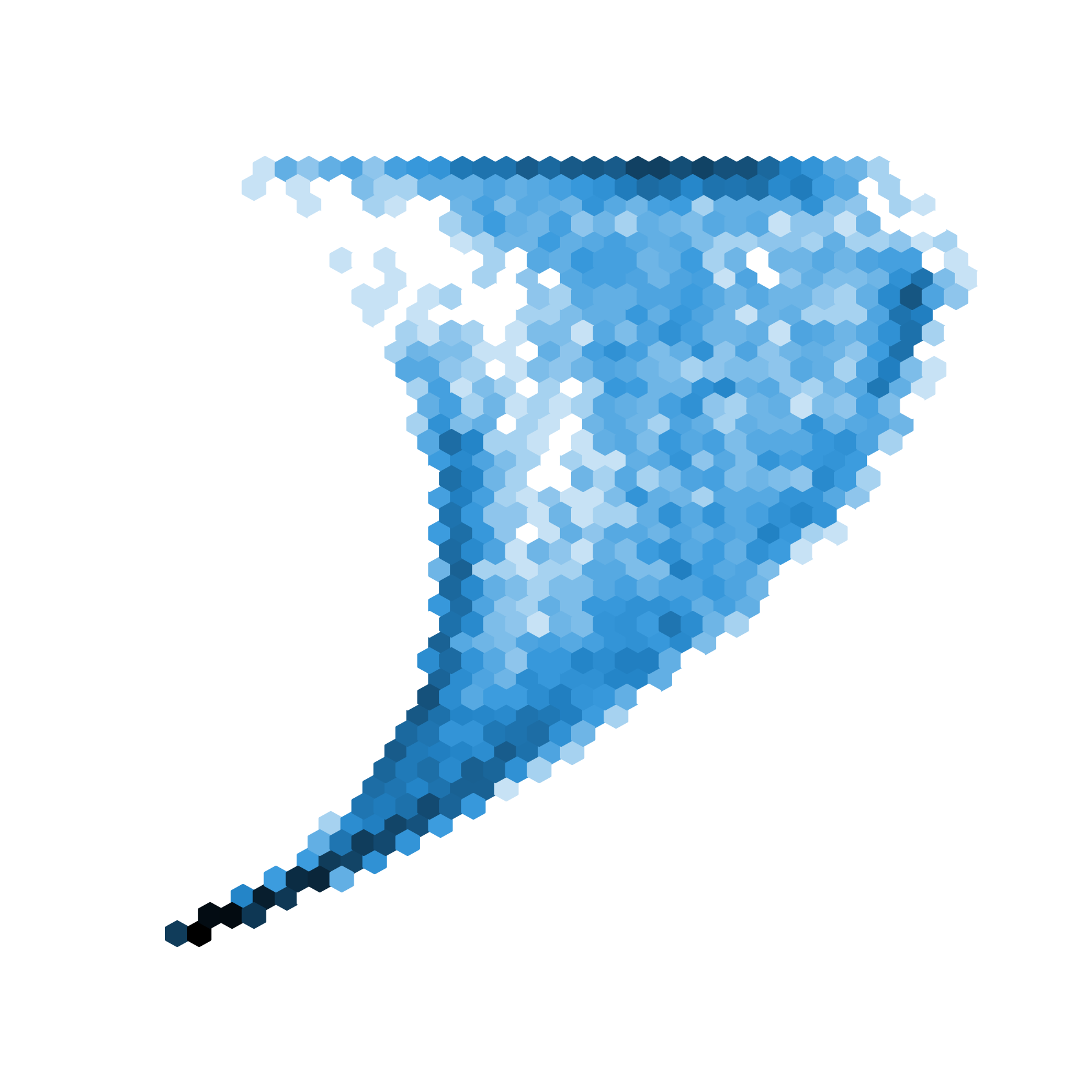};
    \fill[scYellow, fill opacity=0.1] (axis cs:0,1.05) rectangle (1.05,0.5);
    \draw[thin,gray] (axis cs:0,0.5) -- (axis cs:1.05,0.5);
    \draw[thin,gray] (axis cs:0.5,0) -- (axis cs:0.5,1.05);
    \draw[-latex, thick] (axis cs:0.25,0.75) -- (axis cs:0.75,0.75);
  \end{axis}
\end{tikzpicture}
\end{minipage}
\begin{minipage}{.49\textwidth}
  \centering
\begin{tikzpicture}
    \node (a) at (-0.4\figurewidth,0) {};
    \node (b) at (0.4\figurewidth,0) {};
    \draw[|-|] (a) -- node[above, midway] {\scriptsize After Active Learning} (b);
\end{tikzpicture}\\
\begin{tikzpicture}
    \begin{axis}[
        height=3cm,
        width=1.05\figurewidth, 
        axis lines = middle,
        tick label style={font=\tiny},
        ytick=\empty,
        y axis line style={draw=none},
        axis background/.style={fill=scYellow!10},
        ybar,
        bar width=3.4pt,
    ]
        \addplot+[ybar, fill=scPurple, draw=white, fill opacity=0.5, draw opacity=0.7, very thin] table [x index=0, y index=1, col sep=space] {figs/marginal_entropy_hist_data/error_histogram_bins_after_map.dat};
        \addplot+[ybar, fill=scBlue, draw=white, fill opacity=0.5, draw opacity=0.7, very thin] table [x index=0, y index=1, col sep=space] {figs/marginal_entropy_hist_data/error_histogram_bins_after_la.dat};
        \node[fill=scYellow!10,fill opacity=0.8, rounded corners, inner sep=0pt,align=center] at (0.76,7) {\scriptsize \textbf{Error} (entropy $<1$)\\[-0.4em]\scriptsize (counts in log-scale) };
    \end{axis}
\end{tikzpicture}\\
\pgfplotsset{
      width=0.6\figurewidth,
      enlargelimits=false,
      tick label style={font=\tiny},
      axis equal image,
      xlabel={}, 
      xlabel near ticks,
      xlabel style={yshift=4pt},
      ylabel near ticks,
      xtick={0, 0.5, 1}, 
      ytick={0, 0.5, 1},
      xticklabels={0, 1, 2},
      yticklabels={0, 0.5, 1},
      ytick pos=left,
      xtick pos=bottom,
      ytick align=outside,
      xtick align=outside, 
      xmin=0, xmax=1.05, ymin=0, ymax=1.05
}
\begin{tikzpicture}
  \begin{axis}[
      xlabel={\scriptsize Uncertainty (Entropy)},
      ylabel={\scriptsize Error},
    ]
   \addplot graphics[xmin=-0.21, xmax=1.20, ymin=-0.18, ymax=1.20] {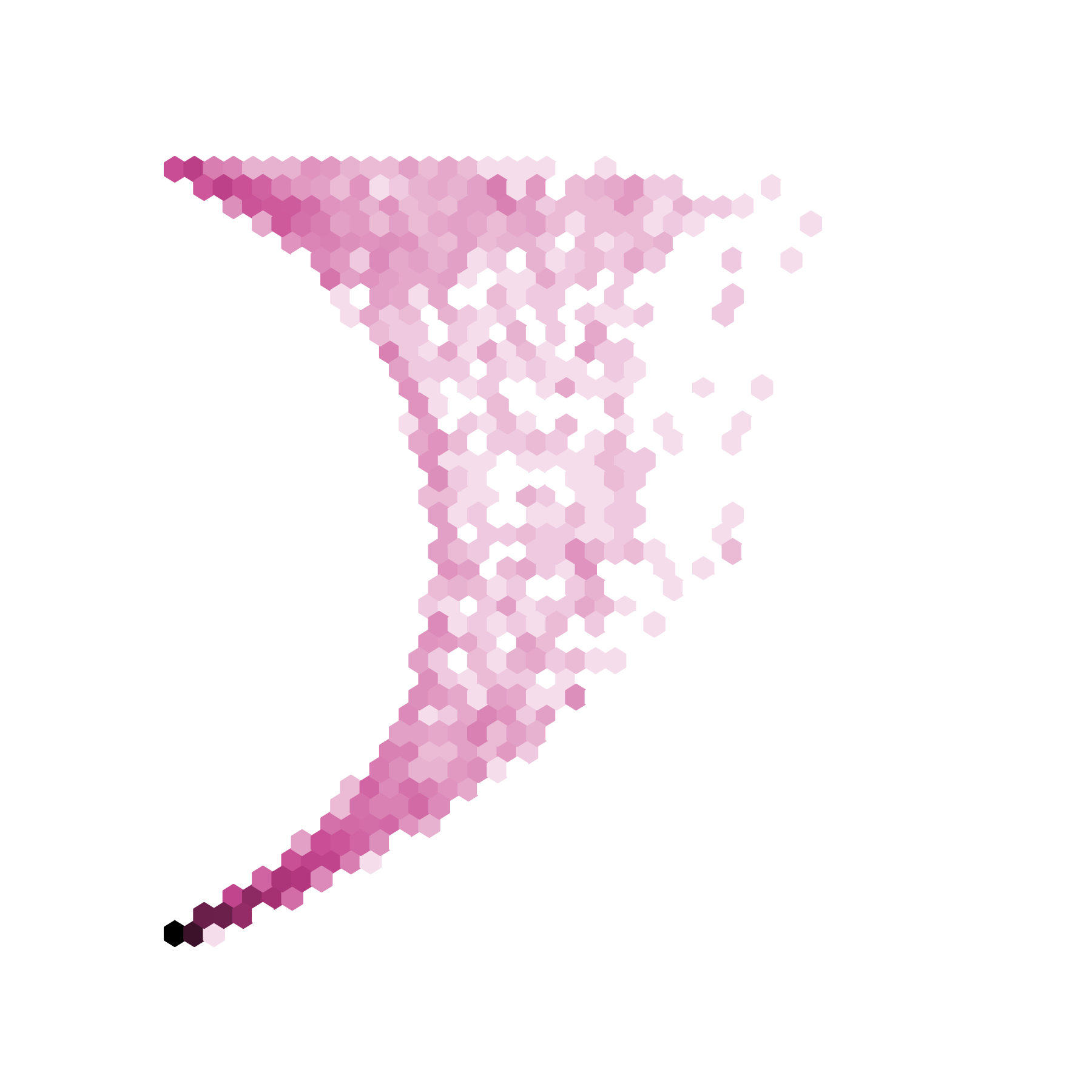};
    \fill[scYellow, fill opacity=0.1] (axis cs:0,1.05) rectangle (0.5,0);
    \draw[thin,gray] (axis cs:0,0.5) -- (axis cs:1.05,0.5);
    \draw[thin,gray] (axis cs:0.5,0) -- (axis cs:0.5,1.05);
    \node at (axis cs:0.1,0.75) {\scriptsize (b)};
    \node at (axis cs:0.96,0.75) {\scriptsize (a)};
    \node at (axis cs:0.1,0.25) {\scriptsize (c)};
    \node at (axis cs:0.96,0.25) {\scriptsize (d)};
  \end{axis}
\end{tikzpicture}
\begin{tikzpicture}
  \begin{axis}[
      xlabel={\scriptsize Uncertainty (Entropy)}, 
    ]
    \addplot graphics[xmin=-0.21, xmax=1.20, ymin=-0.18, ymax=1.20] {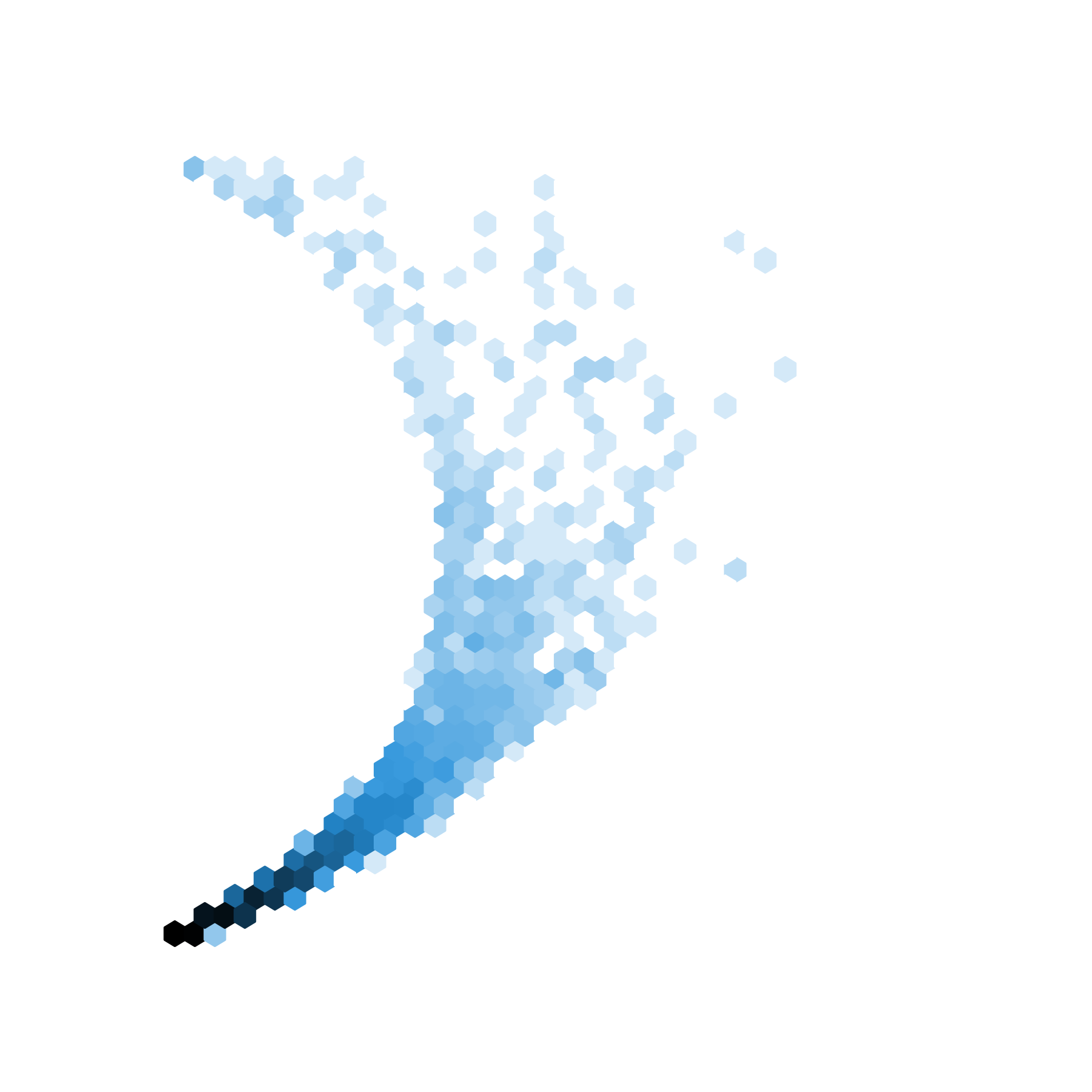};
    \fill[scYellow, fill opacity=0.1] (axis cs:0,1.05) rectangle (0.5,0);
    \draw[thin,gray] (axis cs:0,0.5) -- (axis cs:1.05,0.5);
    \draw[thin,gray] (axis cs:0.5,0) -- (axis cs:0.5,1.05);
    \draw[-latex, thick] (axis cs:0.75,0.75) -- (axis cs:0.35,0.25);
  \end{axis}
\end{tikzpicture}
\end{minipage}   \vspace*{-0.7em}
  \caption{Predictive error vs.\ uncertainty (entropy) on the EuroSAT data set \citep{helber2019eurosat} for the OpenCLIP ViT-H-14 model. \textbf{The zero-shot} comparison (left side) of the original model (\legendhist{scPurple}) and its Bayesian counterpart (\legendhist{scCyan}) indicates that our Bayesian model exhibits better calibration and substantially reduces overconfident predictions. 
  \textbf{Active Learning} results (right side) show that those improvements lead to a substantially reduced misclassification rate after adaptation; quadrant (b).
  }
  \label{fig:active_learning}
    \vspace*{-8pt}
\end{figure}

This work proposes BayesVLM, an efficient and effective post-hoc uncertainty quantification method for pre-trained VLMs that adheres to the outlined desiderata.
We leverage a Laplace approximation \citep{mackay1992information} to the Bayesian posterior, thereby eliminating the need for additional training, architectural changes, or modifications to the training objective.
For this, we introduce independent probabilistic models for each modality, adhering to the \iid assumption and enabling efficient posterior inference.
Further, we derive an analytical expression for the distribution over cosine similarities for efficient uncertainty propagation.
We evaluate our approach on zero-shot classification benchmarks and for uncertainty-aware active fine-tuning~\citep{gal2017deep,hubotter2024transductiveactivelearningtheory}, finding improvements in performance over baselines in both scenarios.
Lastly, we assess the efficiency and robustness of our approach (BayesVLM) and find that BayesVLM provides efficient, effective, and robust uncertainty estimates, even when the VLM is pre-trained on proprietary data.
\paragraph{Contributions} 
The overall contributions are illustrated in \cref{fig:teaser} and can be summarised as follows:
{\em (i)}~we propose BayesVLM, an efficient and effective post-hoc method for uncertainty quantification in pre-trained VLMs, without architecture changes or further training (\cref{sec:method});
{\em (ii)}~we present the first direct Bayesian formulation of vision-language models and derive an analytical expression of the distribution over cosine similarities for efficient uncertainty propagation (\cref{sec:bayesvlm});
{\em (iii)}~we demonstrate the efficacy of BayesVLM in both zero-shot and active learning settings, showing improvements over baselines in both settings. And we assess its efficiency and robustness, finding that BayesVLM provides robust estimates while introducing little to no computational overhead (\cref{sec:experiments}).\looseness-1

\section{Related work}\label{sec:related_work}
\paragraph{Vision-language models}
Models like CLIP~\citep{radford2021learning} and SigLIP~\citep{zhai2023sigmoid}, trained on massive data sets such as LAION~\citep{schuhmann2022laion}, have become widespread in various applications, including zero-shot classification, generative modeling~\citep {rombach2022high,podell2023sdxl}, and retrieval~\citep{saito2023pic2word,karthik2023vision}. This work presents an effective post-hoc approach to uncertainty estimation for these pre-trained VLMs.

\paragraph{Uncertainty in vision-language models} Quantifying uncertainties in VLMs has observed increasing interest, with approaches involving learning probabilistic embeddings, for example, by learning additional probabilistic adapters \citep{prolip,lafon2025vilu} or through pre-training/fine-tuning with a probabilistic loss \citep{pcmepp,ju2025exploiting}. In addition, recent approaches also explored training-free uncertainty quantification, \eg, through test-time augmentation~\citep{ayhan2018test} or zero-shot out-of-distribution detection~\citep{fu2025clipscope}. Another key approach is to solely focus on calibration through methods such as temperature scaling~\citep{guo2017calibration}. Further related works are discussed in \cref{app:related_work}.
In contrast, we present a training-free post-hoc approach that does not require architectural changes, but estimates the Bayesian posterior of a pre-trained model and efficiently propagates uncertainty arising from the Bayesian posterior to the VLM output.

\paragraph{Active learning}
The goal of active learning~\citep{ren2021survey,settles2009active} is to improve model performance by `actively' selecting additional informative data through an acquisition function \citep{holub2008entropy, sener2018active}. A particularly relevant area is Bayesian active learning~\citep{mackay1992information,gal2017deep}, where acquisition functions leverage model uncertainties. Notable examples include the BALD score~\citep{houlsby2011bayesian} and EPIG~\citep{bickfordsmith2023prediction}, both of which are functions of information gain. While such methods are gaining traction in large language models~\citep{hubotter2025efficiently}, they remain relatively underexplored for VLMs, where ad-hoc strategies are more prevalent. In our work, we bridge this gap.

\section{Methods}
\label{sec:method}
\paragraph{Notation}
We denote vectors by bold lower-case letters (\eg, $\bx, \ba$) and use bold upper-case letters for matrices (\eg, $\bX, \bP$). Further, sets are denoted in upper-case calligraphic letters (\eg, $\mcD, \mcI$) and model parameters or hyperparameters are denoted using Greek letters (\eg, $\alpha, \btheta$). In particular, let $\bx^\img_i \in \reals^{p_{\img}}$ and $\bx^\txt_j \in \reals^{p_{\txt}}$ denote the $i$\textsuperscript{th} image and $j$\textsuperscript{th} text description, respectively.
Further, let $\phi\colon \reals^{p_{\img}} \to \reals^{d_{\img}}$ and $\psi\colon \reals^{p_{\txt}} \to \reals^{d_{\txt}}$ denote the image and text encoders of the VLM, where $p_\img$ and $p_\txt$ are the respective input dimensionalities and $d_{\img}$, $d_{\txt}$ is the dimensionality of the respective feature space.
Then, by denoting the linear image and text projections as  $\bP \in \reals^{d \times d_{\img}}$ and $\bQ \in \reals^{d \times d_{\txt}}$ respectively, the feature embeddings in the joint space can be written as $\bg=\bP \phi(\bx^\img)$ and $\bh=\bQ\psi(\bx^\txt)$.
We write $\bG$ and $\bH$ to denote the matrices of stacked image and text embeddings, respectively, whose rows correspond to the individual $\bg_i$ and $\bh_j$.
Lastly, we use the hat symbol to denote unit-length normalised vectors, \eg, $\hat{\bg} = \nicefrac{\bg}{\|\bg\|}.$
The notation is listed in full in \cref{app:notation}.
\subsection{Background}\label{sec:background}
\paragraph{Language-image pre-training}
We consider VLMs trained by minimising the InfoNCE loss \citep{oord2018representation} (\eg, CLIP~\citep{radford2021learning}) and present additional experiments for the SigLIP loss \citep{zhai2023sigmoid}.
Specifically, the InfoNCE loss is defined as the sum of two cross-entropy terms, one for each relational direction---image to text ($\mcL_{\CE}(\bX^\img, \bX^\txt)$) and text to image ($\mcL_{\CE}(\bX^\txt,\bX^\img)$).
The total loss is defined as follows $\mcL_{\text{InfoNCE}}(\bX^\img, \bX^\txt)=$
\begin{equation}\label{eq:cliploss}
	-\underbrace{\frac{1}{2n}  {\textstyle \sum_{i=1}^n} \log \frac{ \exp( t\hat{\bg}_i\transpose \hat{\bh}_i ) }{  {\textstyle \sum_{j=1}^n} \exp(t\hat{\bg}_i\transpose \hat{\bh}_j) }}_{\imgtotxt,\,\mcL_{\CE}(\bX^\img, \bX^\txt)} - \underbrace{ \frac{1}{2n}  {\textstyle \sum_{i=1}^n} \log \frac{ \exp( t\hat{\bh}_i\transpose \hat{\bg}_i ) }{  {\textstyle \sum_{j=1}^n}\exp(t\hat{\bh}_i\transpose \hat{\bg}_j) }}_{\txttoimg,\,\mcL_{\CE}(\bX^\txt, \bX^\img)},
\end{equation}
where $t$ is a learnable temperature parameter, $n$ denotes the number of image-text pairs, and $\hat{\bg}$ and $\hat{\bh}$ are the unit-length normalised embeddings.
This contrastive loss function encourages embeddings for matching image-text pairs to be similar while simultaneously pushing unrelated image-text pairs away from each other \citep{oord2018representation}. 
In practice, evaluating this loss is infeasible on billions of data points.
The common practice adopted is to evaluate it on a sufficiently large batch.
Recently, the SigLIP loss \citep{zhai2023sigmoid} was proposed as an alternative to the InfoNCE loss, a binary classification loss over cosine similarities. Further details on SigLIP are given in \cref{app:background_vlms}.

\paragraph{Laplace approximation}
Given a data set $\mcD=\{(\bx_i, \by_i)\}_{i=1}^n$ and denote the neural network parameters as $\btheta$. In Bayesian deep learning, we aim to estimate the posterior distribution, \ie, 
\begin{equation} \label{eq:posterior}
	 p(\btheta \mid \mcD) = \frac{p(\btheta) \prod_{i=1}^n p(\by_i \mid \bx_i, \btheta) }{\int_{\btheta} p(\btheta)  \prod_{i=1}^n p(\by_i \mid \bx_i, \btheta) \,\dee \btheta} = \frac{\highlight{gray!20}{prior} \times \highlight{gray!20}{likelihood} }{\highlight{scRed!20}{marginal likelihood}}.
\end{equation}
Since the marginal likelihood involves an intractable high-dimensional integral, we approximate the posterior.
We adopt the Laplace approximation (LA) \citep{mackay1992information}, a post-hoc method that has been increasingly used in the Bayesian deep learning community~\citep{daxberger2021laplace,li2024streamlining,meronen2024dnn,ritter2018scalable,roy2022uncertainty,scannell2024functionspace}.

Specifically, LA fits a Gaussian distribution to the posterior, centred at the MAP estimate of a \textit{pre-trained} model, and is therefore `post-hoc'.
The \emph{prior} is implicitly defined by the L2 regularisation (weight decay) commonly used during training~\citep{radford2021learning,zhai2023sigmoid}, and corresponds to a diagonal Gaussian prior on the parameters, \ie, $p(\btheta)=\mcN(\mathbf{0}, \lambda^{-1} \bI)$. 
The \emph{likelihood} is defined by the training loss. 
The final approximate posterior is given as $p(\btheta \mid \mcD) \approx \mcN(\btheta_{\MAP}, \bSigma)$ where $\btheta_{\MAP}$ is the MAP estimate and $\bSigma = \rbra{-\nabla^2_{\btheta} \log p(\mcD \mid \btheta) \rvert_{\btheta = \btheta_{\MAP}} + \lambda \bI}^{-1}$ is the Hessian of the negative log joint evaluated at $\btheta_{\MAP}$. A detailed derivation is given in \cref{app:background_laplace}.

\begin{figure*}[t]
	\centering
	\resizebox{\textwidth}{!}{
\begin{tikzpicture}
			
  \newcommand{\encoder}[4]{%
    \begin{scope}[xshift=#3cm, yshift=#4cm]
      \fill[#1] (0,0) -- (0,3*#2) -- (1,2*#2) -- (1,-2*#2) -- (0,-3*#2) -- cycle;
    \end{scope}
  }
  
  \newcommand{\dist}[6]{%
    \begin{scope}[xshift=#2cm, yshift=#3cm, fill opacity=0.25]
      \draw[rotate=#4, thick, #1!60, fill=#1!60, #6] (0, 0) ellipse (#5);
      \draw[rotate=#4, thick, #1!90, fill=#1!90, scale=0.5, #6] (0, 0) ellipse (#5);
      \draw[rotate=#4, thick, draw=#1, fill=#1, scale=0.5*0.25, #6] (0, 0) ellipse (#5);
    \end{scope}
  }
  
  \pgfmathdeclarefunction{gauss}{2}{%
    \pgfmathparse{1/(#2*sqrt(2*pi))*exp(-((x-#1)^2)/(2*#2^2))}%
  }

  \node[draw=scYellow, thick, anchor=east] at (-2.8,0) {\includegraphics[width=1cm]{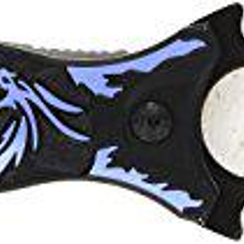}};
  \node[draw=scBlue, thick, anchor=east] at (-1.4,0) {\includegraphics[width=1cm]{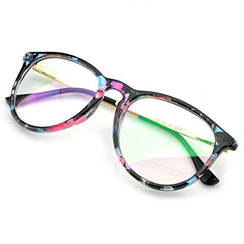}};
  \node[draw=scRed, thick, anchor=east] at (0,0) {\includegraphics[width=1cm]{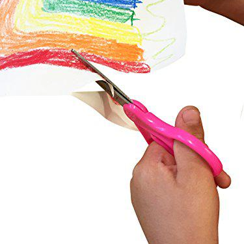}};

  \node[draw=scPurple, thick, dashed, align=center, anchor=east] at (0,-1.5) {\small An image of \{reading glasses\}.};
  \node[draw=scGreen, thick, dashed, align=center, anchor=east] at (0,-2.25) {\small An image of \{scissors\}.};
  \node[draw=scCyan, thick, dashed, align=center, anchor=east] at (0,-3) {\small An image of \{crayon pencils\}.};

  \encoder{black!20, draw=black!60}{0.25}{1}{0}
  \node at (1.5,0) {\small $\phi$};
  \encoder{black!20, draw=black!60}{0.25}{1}{-2.25}
  \node at (1.5,-2.25) {\small $\psi$};

  \draw[-latex, scYellow] (-3.4, 0.6) to [out=15,in=140] (0.95,0.25);
  \draw[-latex, scBlue] (-2,-0.6) to [out=-15, in=220] (0.95,-.25);
  \draw[-latex, scRed] (0,0) to [out=0, in=180] (0.95,0);
  
  \draw[-latex, scPurple] (0,-1.5) to [out=0,in=140] (0.95,-2);
  \draw[-latex, scGreen] (0,-2.25) to [out=0,in=180] (0.95,-2.25);
  \draw[-latex, scCyan] (0,-3) to [out=0,in=220] (0.95,-2.5);

  \begin{scope}[yshift=-0.125cm, xshift=4.5cm]
    \node at (-0.8,0.125) {\small $\phi(\bx^{\img}_i) \, \times$};
    \node at (0.5, 1) {\scriptsize $\distNorm(\bP,
    \bSigma_{\img})$ };

    \pic[scale=0.25, fill=scRed] at (0, 0.375) {gauss=0.5/0.5};
    \pic[scale=0.25, fill=scRed] at (0, 0) {gauss=-0.5/0.5};
    \pic[scale=0.25, fill=scRed] at (0, -0.375) {gauss=0.5/0.3};
    \pic[scale=0.25, fill=scRed] at (0.5, 0.375) {gauss=1/0.2};
    \pic[scale=0.25, fill=scRed] at (0.5, 0) {gauss=0.25/0.4};
    \pic[scale=0.25, fill=scRed] at (0.5, -0.375) {gauss=0/0.3};
    \pic[scale=0.25, fill=scRed] at (0.5*2, 0.375) {gauss=0/0.3};
    \pic[scale=0.25, fill=scRed] at (0.5*2, 0) {gauss=1.5/0.7};
    \pic[scale=0.25, fill=scRed] at (0.5*2, -0.375) {gauss=-0.5/0.5};
  \end{scope}
  
  \begin{scope}[yshift=-2.375cm, xshift=4.5cm]
    \node at (-0.8,0.125) {\small $\psi(\bx^{\txt}_i) \, \times$};
    \node at (0.5, 1) {\scriptsize $\distNorm(\bQ,
    \bSigma_{\txt})$ };

    \pic[scale=0.25, fill=scRed] at (0, 0.375) {gauss=1/0.5};
    \pic[scale=0.25, fill=scRed] at (0, 0) {gauss=-0.5/0.2};
    \pic[scale=0.25, fill=scRed] at (0, -0.375) {gauss=-0.5/0.5};
    \pic[scale=0.25, fill=scRed] at (0.5, 0.375) {gauss=1/0.4};
    \pic[scale=0.25, fill=scRed] at (0.5, 0) {gauss=0.5/0.2};
    \pic[scale=0.25, fill=scRed] at (0.5, -0.375) {gauss=-0.1/0.3};
    \pic[scale=0.25, fill=scRed] at (0.5*2, 0.375) {gauss=1/0.5};
    \pic[scale=0.25, fill=scRed] at (0.5*2, 0) {gauss=.5/0.7};
    \pic[scale=0.25, fill=scRed] at (0.5*2, -0.375) {gauss=0/0.5};
  \end{scope}
  
  \draw[-latex, thick] (2,0) -- (2.9, 0);
  \draw[-latex, thick] (2,-2.25) -- (2.9, -2.25);
  
  \begin{scope}[xshift=4cm]

    \begin{scope}[xshift=0.5cm]
      \dist{scRed}{3.2}{-0.3}{0}{0.75cm and 0.5cm}{} %
      \dist{scYellow}{3}{0}{45}{0.75cm and 0.25cm}{} %
      \dist{scBlue}{3.8}{0.4}{-30}{0.4cm and 0.2cm}{} %
      
      \begin{scope}[yshift=-2.25cm]
        \dist{scCyan}{3.2}{-.4}{-45}{0.4cm and 0.5cm}{dashed} %
        \dist{scPurple}{3.7}{0.3}{22}{0.5cm and 0.3cm}{dashed} %
        \dist{scGreen}{2.8}{0.1}{15}{0.25cm and 0.15cm}{dashed} %
      \end{scope}
    \end{scope}

    \draw[-latex, thick] (2,0) -- (2.9, 0);
    \draw[-latex, thick] (2,-2.25) -- (2.9, -2.25);
    
    \draw[-latex, thick] (5, 0) -- (5.9, 0);
    \draw[-latex, thick] (5, 0) -- (5.9, -1.025);
    \draw[-latex, thick] (5, 0) -- (5.9, -2.15);
    
    \draw[-latex, thick, dashed] (5, -2.25) -- (5.9, -0.1);
    \draw[-latex, thick, dashed] (5, -2.25) -- (5.9, -1.225);
    \draw[-latex, thick, dashed] (5, -2.25) -- (5.9, -2.25);

    \begin{scope}[xshift=6cm]
      
      \begin{scope}[yshift=-0.5cm]
        \begin{axis}[
          no markers, domain=0:10, samples=100,
          height=2.5cm, width=4cm,
          xtick=\empty, ytick=\empty,
          axis line style={draw=scPurple, dashed, thick},
          axis background/.style={fill=scPurple!10}
          ]
          \addplot [very thick, scRed] {gauss(2,1.25)};
          \addplot [very thick, scYellow] {gauss(4,0.75)};
          \addplot [very thick, scBlue] {gauss(8,0.4)};
        \end{axis}
      \end{scope}
      
      \begin{scope}[yshift=-1.625cm]
        \begin{axis}[
          no markers, domain=0:10, samples=100,
          height=2.5cm, width=4cm,
          xtick=\empty, ytick=\empty,
          axis line style={draw=scGreen, dashed, thick},
          axis background/.style={fill=scGreen!10}
          ]
          \addplot [very thick, scBlue] {gauss(2,0.25)};
          \addplot [very thick, scRed] {gauss(5.5,1)};
          \addplot [very thick, scYellow] {gauss(7,0.5)};
        \end{axis}
      \end{scope}
      
      \begin{scope}[yshift=-2.75cm]
        \begin{axis}[
          no markers, domain=0:10, samples=100,
          height=2.5cm, width=4cm,
          xtick=\empty, ytick=\empty,
          axis line style={draw=scCyan, dashed, thick},
          axis background/.style={fill=scCyan!10}
          ]
          \addplot [very thick, scRed] {gauss(6,1.75)};
          \addplot [very thick, scYellow] {gauss(2,1)};
          \addplot [very thick, scBlue] {gauss(1.5,0.5)};
        \end{axis}
      \end{scope}
    \end{scope}

    \node[align=center, anchor=north] at (3.7, 2) {\small Embeddings Become\\ \small Probabilistic  (\cref{eq:feature_dist})};
    \node[align=center, anchor=north] at (7.2, 2) {\small Cosine Similarities Become\\ \small Probabilistic (\cref{eq:cosine_sim_dist}) };
  \end{scope}
  
  \node[, anchor=north] at (-2, 2) {\small Inputs};
  \node[anchor=north, align=center] at (1.5, 2) {\small Encoders \\ \small \faLock};
  
  \node[align=center, anchor=north] at (4.5, 2) {\small Laplace Approx.\\ \small (\cref{eq:hessian})};
  
\end{tikzpicture}
 	}
	\caption{
		{\bfseries Illustration of uncertainty propagation in BayesVLMs:}
		We estimate uncertainties over the last layers of both encoders using a Laplace approximation, which induces probabilistic feature embeddings. We then approximate the distribution over cosine similarities by estimating the expected value and variance. 
        The cosine similarity distribution is then propagated to the VLM output. %
	}
	\label{fig:uncertainty_pipeline}
\end{figure*}

\subsection{BayesVLM: Post-hoc probabilistic VLMs}
\label{sec:bayesvlm}
To estimate predictive uncertainties in a post-hoc fashion for VLMs, we independently estimate the posterior of the image projection $\bP$ and text projection $\bQ$. 
For CLIP, we reformulate the contrastive loss to obtain tractable likelihoods for $\bP$ and $\bQ$, enabling separate posterior inference. 
We then approximate the Hessian of the log-likelihood and show how the resulting posteriors induce a distribution over cosine similarities. 
Finally, we derive a Gaussian approximation of this distribution for efficient downstream inference.
Our BayesVLM pipeline is illustrated in \cref{fig:uncertainty_pipeline}.

\paragraph{Estimate posterior: Likelihood approximation} 
The first step in formulating our Bayesian model, BayesVLM, is to define its likelihood function.
When doing so, we encounter the following key challenges: popular loss functions for VLMs, such as the InfoNCE loss (\cref{eq:cliploss}), entangle modalities and data points.
While this is a desirable behaviour when learning multi-modal models, it breaks the usual \iid assumption made in Bayesian models.
Specifically, we have that
\begin{equation}
  (\bx_i^\img, \bx_i^\txt) \overset{\textcolor{scRed}{\text{non-\iid}}}{\sim}  p(\bx_i^\img, \bx_i^\txt \mid \bX^\img_{\setminus i}, \bX^\txt_{\setminus i}, \btheta),
\end{equation}
which hinders straightforward application of the Bayesian framework, as data is only conditionally independent.
For that purpose, we are instead assuming two independent probabilistic models, one for each modality, with likelihood functions corresponding to the conditional probability for each modality rather than their joint, \ie,
\begin{equation}
  \bx^\img_i \distiid p(\bx^\img_i \mid \bX^\txt, \btheta), \qquad
  \bx^\txt_i \distiid p(\bx^\txt_i \mid \bX^\img, \btheta).
\tag{\iid assumption}
\end{equation}
Consequently, in case of the InfoNCE loss, each likelihood function is given by its respective modality-specific sub-loss term, \ie, in case of the probabilistic model for the image modality, we have $\mcL_{\CE}(\bX^\img, \bX^\txt)$, and corresponds to a categorical distribution.
A similar approximation is also applied to SigLIP. 
Crucially, defining independent probabilistic models for each modality additionally necessitates independence between the encoders.
For example, when treating the projection layers $\bP$ and $\bQ$ probabilistically, we obtain that:
\begin{equation}
  \bP \indep \bQ \tag{Consequence of \iid assumption}.
\end{equation}
Following the \iid assumption, the probabilistic model for the image modality is 
\vspace*{-6pt}
\begin{center}
	\centering
	\begin{tikzpicture}
 	\node [] (a) at (0,0) {$\bx^{\img}_i$};
 	\node [] (b) at (6,0) {$\hat{\bg}_i=\frac{\bP\phi(\bx^{\img}_i)}{\|\bP\phi(\bx^{\img}_i) \|}$};
 	\draw [->,shorten >=8pt,shorten <=8pt] (a) -- node [above,font=\small] {image encoder $\phi(\cdot)$ and} node[below,font=\small] {image projection layer $\bP$} (b);	
 	\node [] (c) at (12,0) {$\hat{\bH}\hat{\bg}_i$,};
 	\draw [->,shorten >=8pt,shorten <=8pt] (b) -- node [above,font=\small] {\textbf{given} text embeddings $\hat{\bH}$} node[below,font=\small] {compute logits} (c);
\end{tikzpicture}
\end{center}
\vspace*{-6pt}
and the likelihood becomes a categorical distribution (see \cref{app:iid-assumption} for formulation)
\[
\log p(\bX^\img \mid \bX^\txt, \btheta)  = \log {\textstyle \prod_{i=1}^n} p(\bx_i^\img \mid \bX^\txt, \btheta) = \log {\textstyle \prod_{i=1}^n}  [\text{softmax}(\hat{\bH}\hat{\bg}_i)]_i.
\]
The probabilistic model for text input can be obtained similarly.
We can now apply the LA to this probabilistic model to estimate the approximate posterior.

\emph{Why is this still a reasonable approximation?} For VLMs, it is important to capture interactions between modalities, and assuming independence seems problematic at first. However, as we are using a local post-hoc posterior estimation through the LA, we are effectively introducing an independence conditionally on the MAP estimate of the (joint) contrastive loss. Thus, crucially, even though we assume independence between modalities, we can still capture interactions between modalities.
Note that this assumption is also important for computational reasons, as it helps us derive a computationally efficient approach.
Our empirical assessment of the Hessian block structure, as discussed in \cref{app:subsec:independence_assumption}, shows that cross-modal curvature terms are moderate in magnitude, indicating low to moderate cross-modal dependencies.
A detailed discussion is given in \cref{app:exact-model,app:iid-assumption}.

\paragraph{Estimate posterior: Hessian approximation}
Computing the full Hessian of the negative log-likelihood for the posterior covariance in the Laplace approximation is infeasible, as its size scales quadratically with the number of model parameters, making both its estimation and subsequent predictions computationally prohibitive. We, therefore, adopt the Generalised Gauss–Newton (GGN) approximation~\citep{schraudolph2002fast}, which requires the Jacobian of the outputs with respect to the parameters. For linear projection layers, this Jacobian can be derived in closed form. For the image and text encoders, however, the parameter count is prohibitively large, so we treat them as deterministic and approximate the posterior only over the projection matrices $\bP$ and $\bQ$.

To further reduce computational and memory costs, we use the Kronecker-factored (KFAC) Generalised Gauss–Newton (GGN) approximation~\citep{ritter2018scalable,martens2015optimizing}, which expresses the Hessian as a Kronecker product of two smaller matrices. 
This preserves a richer posterior structure than diagonal approximations. 
Following \citep{ritter2018scalable}, the KFAC GGN approximation for the Hessian of $\bP$ is 
\begin{equation}\label{eq:kfacggn}
	\underbrace{\left(\nicefrac{1}{\sqrt{n}} {\textstyle \sum_{i=1}^n} \phi(\bx_i^{\img}) {\phi(\bx_i^{\img})}^{\top}\right)}_{\bA_{\img}}  \otimes \underbrace{\left(\nicefrac{1}{\sqrt{n}} {\textstyle \sum_{i=1}^n} \bJ_\img(\bx^\img_i)\transpose \bLambda_{\img} \, \bJ_\img(\bx^\img_i)\right)}_{\bB_{\img}},
\end{equation}
where $\bJ_\img(\bx^\img_i)=\nicefrac{\partial \hat{\bH}\frac{\bg_i}{\|\bg_i\|}}{\partial \bg_i}$ and $\bLambda_{\img}=\diag(\boldsymbol{\pi}) - \boldsymbol{\pi} \boldsymbol{\pi}\transpose$, with $\pi_c = \nicefrac{\exp(f_c)}{\sum_{c'} \exp(f_{c'})}$, $\hat{\bg}_i\transpose \hat{\bh}_c \defines f_c$.
As estimating the Kronecker factors $\bA$ and $\bB$ over the training data set is infeasible, following prior work \citep{ritter2018scalable}, we leverage a subset of the data and include a pseudo-data count $\tau$ to compensate for the reduced sample size.
The posterior covariance over $\bP$ is approximated as \looseness-1
\begin{equation}\label{eq:hessian}
    \bSigma_{\img} = \left(\tau(\bA_{\img} \otimes \bB_{\img}) + \lambda \bI \right)^{-1} \approx \underbrace{\left(\sqrt{\tau}\bA_{\img} + \sqrt{\lambda}\bI \right)^{-1}}_{\widetilde{\bA}_{\img}^{-1}} \otimes \underbrace{\left(\sqrt{\tau}\bB_{\img} + \sqrt{\lambda}\bI \right)^{-1}}_{\widetilde{\bB}_{\img}^{-1}} .
\end{equation}
Note that the Kronecker factors $\bA$ and $\bB$ can be understood as model statistics under the training data. 
After having the Gaussian posterior over $\bP$ and $\bQ$, as Gaussians are closed under linear transformations, the distribution over $\bg$ (and $\bh$) can be obtained analytically:
\begin{align}
	\hspace*{-0.5em}p(\bg \mid \mcD) &= \mcN\left(\bP_{\text{MAP}}\phi(\bx^\img), \left(\phi(\bx^\img)\transpose\widetilde{\bA}_{\img}^{-1}\phi(\bx^\img)\right) \widetilde{\bB}_{\img}^{-1}\right). \label{eq:feature_dist} %
\end{align}
Analogous results hold for the text projection $\bQ$ and text embedding $\bh$, which we omit here for brevity.
See \cref{app:posterior-derivation} for detailed derivations, and \cref{alg:BayesVLM-Precomputation} outlines the steps described above. 
\paragraph{Make predictions: Cosine similarities approximation}
Given a posterior distribution over the model parameters, evaluating the VLM on an image-text pair yields \emph{random} image and text embeddings rather than deterministic ones, inducing a distribution over their cosine similarity. While the cosine similarity remains well-defined, it becomes a random variable and is generally not Gaussian. The default prediction method, Monte Carlo estimation, requires costly sampling. To improve efficiency, we propose \emph{ProbCosine}, a Gaussian approximation of the cosine similarity distribution based on the first two moments of the image and text embeddings.

Let the Gaussian distributions for the probabilistic image and text embeddings have means $\bmu_{\bg} = (\mu_{\bg,1}, \dots, \mu_{\bg,d})$ and $\bmu_{\bh} = (\mu_{\bh,1}, \dots, \mu_{\bh,d})$, and diagonal covariances $\bSigma_{\bg} = \diag(\sigma^2_{\bg,1}, \dots, \sigma^2_{\bg,d})$ and $\bSigma_{\bh} = \diag(\sigma^2_{\bh,1}, \dots, \sigma^2_{\bh,d})$.
Given the cosine similarity $\cossim{\bx}{\by} = \nicefrac{\bx\transpose\by}{\|\bx\|\|\by\|}$ between two vectors, the expected cosine similarity under the distribution of $\bg$ and $\bh$ is approximately:
\begin{equation}
	\EE[ \cossim{\bg}{\bh} ] \approx \frac{\sum^d_i \mu_{\bg,i}\mu_{\bh,i}}{\sqrt{\sum_i \mu^2_{\bg,i} + \sigma^2_{\bg,i} }\sqrt{\sum_i \mu^2_{\bh,i} + \sigma^2_{\bh,i} }}	 , \label{eq:cosine_expected}
\end{equation}
where we use the fact that $\EE[x^2] = \mu^2_x + \sigma^2_x$ and $\EE[\|\bx\|] \leq \sqrt{\sum_{i} \mu^2_{\bx,i} + \sigma^2_{\bx,i} }$ by applying the %
triangle inequality.
We can obtain the second moment (variance) $\Var[\cossim{\bg}{\bh}]$ similarly, which is given as:
\begin{equation}
	\Var[\cossim{\bg}{\bh}] = \frac{\sum_i \sigma^2_{\bg,i}(\sigma^2_{\bh,i} + \mu^2_{\bh,i})+\sigma^2_{\bh,i}\mu^2_{\bg,i}}{\sum_i \mu^2_{\bg,i} + \sigma^2_{\bg,i} \sum_i \mu^2_{\bh,i} + \sigma^2_{\bh,i} } . \label{eq:cosine_variance}
\end{equation}
Henceforth, the local Gaussian approximation to the distribution over cosine similarities is:
\begin{equation}
	p(\cossim{\bg}{\bh}) \approx \mcN\rbra{\EE[ \cossim{\bg}{\bh} ], \Var[\cossim{\bg}{\bh}]}. \label{eq:cosine_sim_dist}
\end{equation}
Finally, the predictive distribution $p(y \mid \bx)$, \eg, in a zero-shot classification setting, is calculated with %
the probit approximation \citep{ghosal2022infinitesimal,gibbs1998bayesian}.
Hence, our approach allows for the direct propagation of model uncertainties to the class conditional.
As shown in \cref{fig:cosine_similarity} (\cref{app:additional_results}), compared to ground truth, our approximation qualitatively results in a low approximation error.
A detailed derivation can be found in \cref{app:distribution_cosine_similarity}, and \cref{alg:BayesVLM-Inference} outlines the steps described above.

\subsection{Application: Probabilistic active few-shot learning}\label{sec:active_learning}
Active learning~\citep{ren2021survey,settles2009active} naturally evaluates uncertainty quality by selecting informative samples via predictive uncertainties.
We assess BayesVLM with Bayesian acquisition functions and adaptive target-region selection.
Given unseen test data $\mcX_{\text{test}}=\{\bx_i^\star\}_{i=1}^{n_{\text{test}}}$ with unknown labels, the goal is to choose a labeled subset $\{(\bx_j,y_j)\}_{j=1}^m$ with $\bx_j,y_j\sim p(\bx,y)$ that best reduces label uncertainty on $\mcX_{\text{test}}$.
We first bias selection toward the query-set predictive distribution, then rank support candidates by influence or informativeness.

\paragraph{Target region selection}
Following \citet{margatina2021active,hubotter2025efficiently}, we first apply $k$-NN in feature space to pre-select support candidates near the test data, focusing on training points likely useful for the downstream task and reducing acquisition-function cost.  
Because features are stochastic, we compute either the expected cosine similarity (\cref{eq:cosine_expected}) or the 2-Wasserstein distance between image-feature distributions. Details of the calculations are given in \cref{app:subsec:targeted_selection}.

\paragraph{Acquisition functions}
We consider the BALD \citep{gal2017deep} and EPIG \citep{bickfordsmith2023prediction} scores as acquisition functions and assess their viability on downstream tasks.
Both acquisition functions can utilise model uncertainties estimated by the LA, but they differ conceptually in terms of which uncertainties are targeted. See \cref{app:subsec:acquisition_functions} for details.

\paragraph{Online Laplace approximation}
We maintain a Laplace posterior over the image-projection matrix $\bP$ and update it online by {\em (i)}~a gradient step on $\bP$ and {\em (ii)}~updating the Kronecker factors.  
The prior precision can optionally be re-estimated after each step \citep{lin2023online}; see \cref{app:subsec:online_laplace}.

\newcommand{\ACC}{ACC $\uparrow$}
\newcommand{\NLPD}{NLPD $\downarrow$}
\newcommand{\ECE}{ECE $\downarrow$}
\newcommand{\increase}[1]{\scriptsize $(\textcolor{scGreen}{{\uparrow}\,#1})$}
\newcommand{\decrease}[1]{\scriptsize $(\textcolor{scRed}{{\downarrow}\,#1})$}

\begin{table*}[t]
  \centering
  \vspace*{-2pt}
  \small  %
    \caption{{\bfseries Does BayesVLM provide useful uncertainty estimates in zero-shot settings? Yes.} With the OpenCLIP ViT-B-32 model, our BayesVLM performs on par with CLIP and temp. scaling on ACC (\%) and NLPD, while being better calibrated according to the ECE. }
  \label{tab:uq_results_clip}  
  \setlength{\tabcolsep}{3pt}  %
\vspace*{-6pt}
\resizebox{\textwidth}{!}{

\begin{tabular}{l|l|ccccccc}
\toprule
Metrics & Methods & \sc Flowers-102 & \sc Food-101 & \sc CIFAR-10 & \sc CIFAR-100 & \sc ImageNet-R & \sc UCF101 & \sc SUN397 \\
\hline
    \multirow{4}{*}{ACC $\uparrow$} & CLIP~\citep{radford2021learning} & \val{\bf}{68.99}{0.5899} & \val{}{80.21}{0.2507} & \val{\bf}{93.61}{0.2446} & \val{\bf}{73.76}{0.4399} & \val{}{74.52}{0.5032} & \val{}{59.82}{0.7971} & \val{}{67.18}{0.3333}\\
 & CLIP (temp. scaling) & \val{\bf}{68.99}{0.5899} & \val{}{80.21}{0.2507} & \val{\bf}{93.61}{0.2446} & \val{\bf}{73.76}{0.4399} & \val{}{74.52}{0.5032} & \val{}{59.82}{0.7971} & \val{}{67.18}{0.3333}\\
 & TTA~\citep{farina2024frustratingly} & \val{\bf}{68.87}{0.5905} & \val{\bf}{81.68}{0.2435} & \val{}{88.54}{0.3185} & \val{}{65.64}{0.4749} & \val{\bf}{78.29}{0.4760} & \val{\bf}{63.07}{0.7847} & \val{\bf}{68.58}{0.3295}\\
\rowcolor{scCyan!30}
\cellcolor{white} & BayesVLM
  & \val{\bf}{68.87}{0.4630}   %
  & \val{}{80.43}{0.3968}      %
  & \val{\bf}{93.62}{0.2444}   %
  & \val{\bf}{73.63}{0.4406}   %
  & \val{}{74.45}{0.4361}      %
  & \val{}{61.43}{0.4868}      %
  & \val{}{66.96}{0.4703}      %
\\\hline
    \multirow{4}{*}{NLPD $\downarrow$} & CLIP~\citep{radford2021learning} & \val{}{1.90}{0.0486} & \val{}{0.70}{0.0094} & \val{}{0.21}{0.0079} & \val{}{0.97}{0.0173} & \val{}{1.07}{0.0237} & \val{}{1.59}{0.0366} & \val{}{1.16}{0.0131}\\
 & CLIP (temp. scaling) & \val{\bf}{1.67}{0.0373} & \val{}{0.69}{0.0073} & \val{}{0.21}{0.0061} & \val{\bf}{0.94}{0.0138} & \val{}{1.04}{0.0191} & \val{}{1.46}{0.0282} & \val{\bf}{1.11}{0.0100}\\
 & TTA~\citep{farina2024frustratingly}  & \val{}{1.86}{0.0475} & \val{\bf}{0.67}{0.0094} & \val{}{0.35}{0.0092} & \val{}{1.26}{0.0178} & \val{\bf}{0.90}{0.0210} & \val{}{1.50}{0.0363} & \val{}{1.14}{0.0131}\\
\rowcolor{scCyan!30}
\cellcolor{white} & BayesVLM
  & \val{}{1.73}{0.0320}   %
  & \val{\bf}{0.68}{0.0126} %
  & \val{\bf}{0.20}{0.0067} %
  & \val{}{0.95}{0.0152}    %
  & \val{}{1.03}{0.0177}    %
  & \val{\bf}{1.44}{0.0183} %
  & \val{}{1.12}{0.0155}    %
\\\hline
\multirow{4}{*}{ECE $\downarrow$} & CLIP~\citep{radford2021learning} & $6.59$ & $3.91$ & $1.45$ & $6.31$ & $5.20$ & $11.52$ & $8.71$\\
 & CLIP (temp. scaling) & $5.51$ & $4.74$ & $1.88$ & $3.07$ & $4.80$ & $3.61$ & $2.67$\\
 & TTA~\citep{farina2024frustratingly}  & $9.63$ & $4.18$ & $2.02$ & $5.27$ & $2.88$ & $11.75$ & $9.92$\\
\rowcolor{scCyan!30}
\cellcolor{white} & BayesVLM
  & $\bf4.22$  %
  & $\bf1.69$  %
  & $\bf0.72$  %
  & $\bf1.92$  %
  & $\bf1.78$  %
  & $\bf3.57$  %
  & $\bf2.06$  %
\\
\bottomrule
\end{tabular}
     }
\end{table*}

\begin{table*}[t]
  \centering
  \vspace*{-1.5pt}
  \small  %
  \caption{{\bfseries Can ProbCosine improve the zero-shot performance of pre-trained probabilistic models? Yes.} Applying ProbCosine (Ours) to PCME++~\citep{pcmepp} consistently improves zero-shot performance over its standard prediction (Mean) across classification benchmarks and metrics.}
  \label{tab:uq_results_pcmepp} 
  \setlength{\tabcolsep}{3pt}  %
\vspace*{-6pt}
\resizebox{\textwidth}{!}{

\begin{tabular}{l|l|ccccccc}
\toprule
Metrics & Methods & \sc Flowers-102 & \sc Food-101 & \sc CIFAR-10 & \sc CIFAR-100 & \sc ImageNet-R & \sc UCF101 & \sc SUN397 \\
\hline
 & Mean  & \val{\bf}{40.59}{0.0063} & \val{}{65.47}{0.0030} & \val{\bf}{75.16}{0.0043} & \val{\bf}{42.52}{0.0049} & \val{\bf}{42.87}{0.0057} & \val{\bf}{45.97}{0.0035} & \val{\bf}{28.50}{0.0073}\\
\rowcolor{scCyan!30}\multirow{-2}{*}{\cellcolor{white}ACC $\uparrow$} & Ours  & \val{}{40.43}{0.0063} & \val{\bf}{65.54}{0.0030} & \val{\bf}{75.12}{0.0043} & \val{\bf}{42.60}{0.0049} & \val{\bf}{42.83}{0.0057} & \val{\bf}{46.00}{0.0035} & \val{\bf}{28.50}{0.0073}\\\hline
 & Mean  & \val{}{3.22}{0.0471} & \val{}{1.30}{0.0125} & \val{}{0.77}{0.0132} & \val{}{2.28}{0.0216} & \val{}{2.77}{0.0346} & \val{}{2.18}{0.0169} & \val{}{3.83}{0.0550}\\
\rowcolor{scCyan!30}\multirow{-2}{*}{\cellcolor{white}NLPD $\downarrow$} & Ours  & \val{\bf}{3.04}{0.0407} & \val{\bf}{1.25}{0.0109} & \val{\bf}{0.75}{0.0117} & \val{\bf}{2.21}{0.0193} & \val{\bf}{2.59}{0.0301} & \val{\bf}{2.09}{0.0146} & \val{\bf}{3.50}{0.0472}\\\hline
 & Mean  & $8.81$ & $6.78$ & $4.79$ & $10.78$ & $17.38$ & $12.62$ & $26.03$\\
\rowcolor{scCyan!30} \multirow{-2}{*}{\cellcolor{white}ECE $\downarrow$} & Ours  & $\bf2.79$ & $\bf1.54$ & $\bf2.02$ & $\bf4.89$ & $\bf10.82$ & $\bf5.61$ & $\bf19.41$\\

\bottomrule
\end{tabular}
     }
\vspace{-3mm}
\end{table*}

\section{Experiments}\label{sec:experiments}
We outline our setup and address three questions:  
{\em(i)}~Uncertainty quantification: Does BayesVLM provide reliable uncertainty estimates? {\em(ii)}~Active learning: Can we select informative data for fine-tuning using BayesVLM uncertainty estimates? {\em(iii)}~Efficiency and robustness: Does BayesVLM introduce overhead during inference, does it work in closed-source data settings, and how sensitive is its performance to key hyperparameters?
Further setup details and additional results appear in \cref{app:experiment_details,app:additional_results}.\looseness-3

\paragraph{Data sets}
We evaluate zero-shot classification on \textsc{Flowers-102}~\citep{nilsback08flowers}, \textsc{Food-101}~\citep{bossard14food}, \textsc{CIFAR-10/100}~\citep{krizhevsky2009learning}, \textsc{ImageNet-R}~\citep{hendrycks2021imagenetR}, \textsc{UCF101}~\citep{ucf101}, and \textsc{SUN397}~\citep{sun}.  
For active learning, we form a cross-domain setup with test data from a single domain and a training pool spanning all domains, using OfficeHome~\citep{venkateswara2017officehome} (Art, Clipart, Product) and an ImageNet variant with ImageNet-R and ImageNet-Sketch~\citep{wang2019learning}.

\paragraph{Network architectures} In the zero-shot experiments, we used the OpenCLIP~\citep{ilharcoopenclip2021} ViT-B-32 and ViT-L-14, and the SigLIP-B-16 model \citep{zhai2023sigmoid}. In the active learning experiments, we use either CLIP-Huge and SigLIP-Base and fine-tune their projection layers. 

\paragraph{Zero-shot baselines}
We compare with vanilla CLIP/SigLIP, CLIP/SigLIP with temperature scaling~\citep{guo2017calibration,nixon2019measuring}, and test-time augmentation (TTA)~\citep{farina2024frustratingly}.  
Temperature scaling uses the parameter minimising negative log predictive density (NLPD)~\citep{quinonero2005evaluating} on the ImageNet validation set~\citep{deng2009imagenet}.  
We also show ProbCosine can pair with probabilistic VLMs trained from scratch, \eg, ProLIP~\citep{prolip} and PCME++~\citep{pcmepp}.  
Our focus is on training-free uncertainty estimation, not methods requiring extra adaptation~\citep{upadhyay2023probvlm,zhou2025bayesian}.

\paragraph{Acquisition functions} For active learning, we incorporate the uncertainties from BayesVLM into the acquisition functions BALD and EPIG  and compare against 
random and entropy-based selection. Both BALD and EPIG use target region selection with nearest neighbour (NN), which selects a test sample based on the uncertainty score, and then selects its 1-NN of the labelled training samples. We also combine the random and entropy baselines with this targeted selection strategy. 
\paragraph{Hyperparameter settings} 
We estimated the Hessian with 327k image-text pairs (10 CLIP mini-batches) from LAION-400M~\citep{schuhmann2022laion}, and used the same estimate across all experiments. The pseudo-data count~$\tau$ was selected via grid search to minimise NLPD on the ImageNet validation set, and the prior precision~$\lambda$ was set by maximising the marginal likelihood~(\cref{app:laplace_approximation}). The same hyperparameters were used for both zero-shot and active learning experiments.\looseness-2

\paragraph{Evaluation metrics} 
For the zero-shot experiments, we report the mean and standard error of accuracy (ACC), NLPD~\citep{quinonero2005evaluating}, and the expected calibration error (ECE)~\citep{guo2017calibration} computed over the test set. We use a paired $t$-test with $p=0.05$ to bold results with a significant statistical difference. 
Active-learning results use class-weighted accuracy and NLPD.

\subsection{Uncertainty Quantification: Does BayesVLM provide reliable estimates?} \label{sec:results_uncertainties}

\begin{figure}[t]
    \centering
    \setlength{\figurewidth}{.14\textwidth}
    \setlength{\figureheight}{.09\textwidth}
    \newlength{\boxwidth}
    \setlength{\boxwidth}{.1875\textwidth}
    \pgfplotsset{%
        scale only axis,
        tick label style={font=\tiny}, axis lines = middle, axis line style={->}, 
        ylabel near ticks, 
        xlabel near ticks, 
        y tick label style={rotate=45}, 
        x tick label style={rotate=45},
        yticklabel style={/pgf/number format/fixed, 
                          /pgf/number format/precision=2,
                          /pgf/number format/fixed zerofill},
        scaled y ticks=false,
        xtick = {0, 10, 25, 50, 75, 100, 150, 200},
        xmax = 200,
        xticklabels=\empty,
        every axis plot/.append style={thick},
        ylabel = {\scriptsize Weighted ACC $\rightarrow$},
        xlabel = {}, %
        ylabel style={yshift=-0.1cm},
        yticklabel shift={-0.15cm},
        y label style={at={(axis description cs:-0.3,0.5)},anchor=south},
        xticklabel style={scale=.8},
        yticklabel style={scale=.8},
    }
    \begin{subfigure}[b]{.22\textwidth}
      \raggedleft
      \pgfplotsset{title = {\scriptsize \bfseries OH-Art}}

     \end{subfigure}
    \vspace*{-1.4em}    
    \caption{{\bfseries Can we select informative data for fine-tuning using BayesVLM uncertainty estimates? Yes.} On the OfficeHome data set (OH) and ImageNet variants (IN), when using uncertainty-based scores (EPIG \legendline{scCyan} and BALD \legendline{scBlue}) to select the fine-tuning data, we achieve better performance compared with Entropy (targeted) \legendline{scPurple}, Entropy \legendline{scRed}, Random selection (targeted) \legendline{black}, and Random selection \legendline{gray, dashed}. 
    Thus, highlighting the benefits of using uncertainties from BayesVLM.%
    }
    \label{fig:results_activelearning}
    \vspace{-3mm}
\end{figure}

We first test the uncertainty estimates of BayesVLM in the zero-shot setting.
In \cref{tab:uq_results_clip}, we report the zero-shot performance of the CLIP-base model using our post-hoc BayesVLM approach, alongside baseline methods, with a focus on predictive quality and uncertainty calibration. Results for CLIP-Large are provided in \cref{tab:uq_results_clip_large} (\cref{app:zero_shot_results}).
We observe that BayesVLM achieves similar ACC but lower NLPD than the deterministic CLIP across all data sets, showing that BayesVLM is less overconfident when predicting the incorrect class. BayesVLM performs similarly to temp. scaling on ACC and NLPD, but outperforms all baselines on the ECE. 
Although TTA achieves higher ACC on some benchmarks, BayesVLM is significantly better calibrated, which results in more useful uncertainty estimates. 
We conclude that BayesVLM improves model calibration and uncertainty estimation without compromising performance, indicating the effectiveness of our post-hoc strategy.

To test ProbCosine (\cref{sec:bayesvlm}), we applied it to probabilistic embeddings from the pre-trained VLMs  PCME++~\citep{pcmepp} and ProLIP~\citep{prolip} (see \cref{tab:uq_results_prolip}).  
Zero-shot results for PCME++ (\cref{tab:uq_results_pcmepp}) show that PCME++ combined with ProbCosine keeps accuracy while consistently improving calibration, indicating ProbCosine can improve any VLM with Gaussian embeddings. Similarly, ProbCosine improves calibration in most cases when combined with ProLIP (\cref{tab:uq_results_prolip}).

\subsection{Active Learning: Can we select informative data using BayesVLM?} \label{sec:results_activelearning}

To further assess the utility of BayesVLM's uncertainty estimates, we evaluate it in the active learning setting.
We consider a cross-domain setting where the unlabelled target data is from a single domain while the labelled training samples are from multiple domains. 
The goal is to select the most informative samples from the diverse pool for adapting to the target domain, given a maximum budget (subset size) of support set samples. 
We experiment with the OfficeHome~\citep{venkateswara2017officehome} (OH) dataset, where the domains are \{Art, Clipart, Product\}, and an ImageNet-variant (IN) with domains \{R, Sketch\}.
We incorporated the BayesVLM uncertainties into BALD and EPIG and compared against random and entropy-based selection, either from \emph{(i)} the full training pool (Random, Entropy) or with \emph{(ii)} selection from the test set followed by a 1-NN selection (targeted).\looseness-3

As shown in \cref{fig:results_activelearning}, both EPIG and BALD, with BayesVLM uncertainties for data selection, outperform Random and Entropy across various subset sizes and target domains. On OH-Product and IN-Sketch, EPIG and BALD obtain similar weighted ACC as Entropy-targeted. However, EPIG consistently achieves lower NLPD than Entropy-targeted, which shows that the finetuned model is less overconfident on incorrect predictions when trained with samples selected using BayesVLM.  
Similar conclusions can be observed for CLIP-Huge and SigLIP-Base (\cref{fig:results_activelearning_clip_huge,fig:results_activelearning_siglip_base} in \cref{app:additional_results}).

In \cref{fig:active_learning}, we show the change in the predictive error $(1-p(y=y^*\mid x))$ and the predictive uncertainty (entropy) for BayesVLM before (zero-shot) and after active learning on EuroSAT~\citep{helber2019eurosat} using EPIG. We use 200 support points and compare against CLIP with entropy selection. 
BayesVLM reduces overconfident predictions in the zero-shot setting (samples move (b) $\rightarrow$ (a)), and more effectively adapts to the new data set based on the support set (samples move to (c)).

\subsection{Efficiency and robustness: How efficient and robust is BayesVLM?} \label{sec:results_closedsource}
Following the protocol for zero-shot experiments, we assessed the performance of BayesVLM when estimating the Hessian in settings where the training data is not available, an increasingly common setting for modern machine learning models.
In particular, we estimated the Hessian of BayesVLM using the CC12M as a proxy dataset for CLIP models and used the LAION-400M dataset as a proxy for Google's SigLIP model.
We find that BayesVLM provides robust uncertainty estimates for CLIP even when estimated on the proxy dataset, \cf, \cref{tab:uq_results_base_laion_vs_cc12m}, and it remains stable under mild distribution shifts in the proxy dataset (see \cref{app:subsec:proxy_shift}). Moreover, BayesVLM provides competitive results for SigLIP, a VLM model trained on proprietary data (\cref{tab:uq_results_siglip} in \cref{app:zero_shot_results}), is robust \wrt the pseudo-data count $\tau$ (\cref{app:subsec:pseudo_count}), provides interpretable uncertainties under corruptions (\cref{fig:perturbation_plot_main_text}), and and maintains calibration under substantial distribution shift~(\cref{app:subsec:zeroshot_imagenet_adversarial}).\looseness-3

\begin{table*}[h]
  \centering
  \small  %
  \caption{\textbf{Does BayesVLM work in closed-source data settings? Yes.} With OpenCLIP ViT-B-32 trained on LAION-400M and BayesVLM estimated on the proxy dataset CC12M, we find that results are robust and show only slight degradation; statistically significant differences are \textbf{bold} ($p=0.05$).}
  \label{tab:uq_results_base_laion_vs_cc12m}     
  \setlength{\tabcolsep}{4pt}  %
\vspace*{-6pt}
\resizebox{\textwidth}{!}{
\begin{tabular}{l|l|ccccccc}
\toprule
Metrics & Dataset & \sc Flowers-102 & \sc Food-101 & \sc CIFAR-10 & \sc CIFAR-100 & \sc ImageNet-R & \sc UCF101 & \sc SUN397 \\
\hline
\multirow{2}{*}{ACC $\uparrow$}
  & LAION-400M
    & \val{\bf}{68.87}{0.4630} & \val{}{80.43}{0.3968} & \val{}{93.62}{0.2444} & \val{}{73.63}{0.4406} & \val{}{74.45}{0.4361} & \val{}{61.43}{0.4868} & \val{}{66.96}{0.4703} \\
  & CC12M
    & \val{}{68.12}{0.4660} & \val{}{80.35}{0.3974} & \val{}{93.57}{0.2453} & \val{}{73.78}{0.4398} & \val{}{74.32}{0.4369} & \val{}{61.46}{0.4867} & \val{}{66.81}{0.4709} \\
\hline
\multirow{2}{*}{NLPD $\downarrow$}
  & LAION-400M
    & \val{\bf}{1.73}{0.0320} & \val{}{0.68}{0.0126} & \val{}{0.20}{0.0067} & \val{}{0.95}{0.0152} & \val{\bf}{1.03}{0.0177} & \val{}{1.44}{0.0183} & \val{\bf}{1.12}{0.0155} \\
  & CC12M
    & \val{}{1.77}{0.0330} & \val{}{0.68}{0.0129} & \val{}{0.20}{0.0067} & \val{}{0.95}{0.0152} & \val{}{1.03}{0.0180} & \val{}{1.44}{0.0185} & \val{}{1.13}{0.0162} \\
\hline
\multirow{2}{*}{ECE $\downarrow$}
  & LAION-400M
    & $4.22$ & $1.69$ & $0.72$ & $1.92$ & $1.78$ & $\bf3.77$ & $\bf2.06$ \\
  & CC12M
    & $\bf3.84$ & $\bf0.99$ & $\bf0.70$ & $\bf1.43$ & $\bf1.39$ & $3.83$ & $3.89$\\
\bottomrule
\end{tabular} }
\end{table*}

\paragraph{Computational overhead}%
Compared to the deterministic CLIP, BayesVLM adds under 5\% runtime for CLIP-base and less than 1\% for huge models (\cref{tab:runtime} in \cref{app:runtime}).
Inference cost rises only $0.11\%$ GFLOPs for CLIP-base, whereas TTA needs an $80\times$ increase, see \cref{tab:flops} in \cref{app:runtime}.\looseness-3

\begin{wrapfigure}{r}{0.48\textwidth}
  \centering
  \resizebox{\linewidth}{!}{
  \begin{tikzpicture}

    \def\nodewidth{10mm}

    \node[align=center, text width=12cm, font=\Large\strut] at (3.5, 1) {Expected Value};
    \node[align=center, text width=12cm, font=\Large\strut] at (10.5, 1) {Variance};

    \foreach \a [count=\i] in {0,1,2,3,4,5} {
      \node[minimum size=\nodewidth, inner sep=0pt, draw=white] at (\i, 0) {\includegraphics[width=\nodewidth]{figs/perturbation_plot/input_\a.jpg}};
    }

    \def\meanmaxval{30.66}
    \def\meanminval{17.77}
    \def\meanrangeval{(\meanmaxval-\meanminval)}

    \def\varmaxval{1.16}
    \def\varminval{0.72}
    \def\varrangeval{(\varmaxval-\varminval)}

    \def\rangeval{(\maxval-\minval)}

    \def\meanvalues{{{28.667324, 24.377195, 22.944693, 25.072762, 23.980007, 19.805763}, {30.662502, 28.089537, 27.190496, 30.056845, 29.621847, 24.638433}, {26.113857, 22.81041, 21.354885, 22.767145, 22.596605, 18.271564}, {28.06377, 24.358717, 22.911673, 24.593325, 24.247713, 19.534058}, {24.839317, 22.16473, 20.44263, 22.845757, 23.146006, 17.921112}, {19.88999, 20.707981, 19.58666, 21.409832, 21.706045, 17.771017}}}

    \def\variancevalues{{{0.72650445, 0.7961286, 0.8068778, 0.78780437, 0.8230217, 0.89644986}, {0.8148311, 0.8847842, 0.8980044, 0.8795547, 0.9149752, 0.98888284}, {0.94943386, 1.019018, 1.0361282, 1.0189433, 1.0542376, 1.1279718}, {0.9094317, 0.9791411, 0.99509305, 0.97752655, 1.0128663, 1.0866681}, {0.9604758, 1.0298133, 1.0472734, 1.0302707, 1.065446, 1.1389413}, {0.9873369, 1.0568107, 1.0750171, 1.0581907, 1.0934464, 1.167125}}}

    \foreach \y in {1,...,6} {
      \foreach \x in {1,...,6} {
        \pgfmathsetmacro\meanintensity{100*(\meanvalues[\x-1][\y-1]-\meanminval)/\meanrangeval}
        \node[fill=scCyan!\meanintensity, minimum size=\nodewidth, draw=gray!30] at (\y,-\x) {};
      }
    }
    
    \shade[left color=white, right color=scCyan] (0.5,-7.3) rectangle +(6,0.3);
    \node[below] at (0.75,-7.8) {\meanminval};
    \node[below] at (6.25,-7.8) {\meanmaxval};

    \foreach \a [count=\i from 8] in {0,1,2,3,4,5} {
      \node[minimum size=\nodewidth, inner sep=0pt] at (\i, 0) {\includegraphics[width=\nodewidth]{figs/perturbation_plot/input_\a.jpg}};
    }

    \foreach \y in {1,...,6} {
      \foreach \x in {1,...,6} {
        \pgfmathsetmacro\varintensity{100*(\variancevalues[\x-1][\y-1]-\varminval)/\varrangeval}
        \node[fill=scCyan!\varintensity, minimum size=\nodewidth, draw=gray!30] at (\y+7,-\x) {};
      }
    }

    \shade[left color=white, right color=scCyan] (7.5,-7.3) rectangle +(6,0.3);
    \node[below] at (7.75,-7.8) {\varminval};
    \node[below] at (13.25,-7.8) {\varmaxval};
  
    \foreach \a/\b [count=\i from 2] in {
      {An image of a cup},
      {A\textcolor{scRed}{x} image of a cup},
      {A\textcolor{scRed}{x} ima\textcolor{scRed}{x}e of a cup},
      {A\textcolor{scRed}{x} ima\textcolor{scRed}{x}e\textcolor{scRed}{x}of a cup},
      {A\textcolor{scRed}{x} ima\textcolor{scRed}{x}e\textcolor{scRed}{x}\textcolor{scRed}{x}f a cup},
      {A\textcolor{scRed}{x} ima\textcolor{scRed}{x}e\textcolor{scRed}{x}\textcolor{scRed}{x}f a\textcolor{scRed}{x}cup}} {
      \node[minimum size=\nodewidth, align=right, font=\Large] at (-1.5,-\i+1) {\a};
    }
  \end{tikzpicture}
  }
  \caption{Illustration of ProbCosine under increasing corruption. The mean similarity decreases and variance increases with higher levels of corruption, demonstrating effective uncertainty estimation under distribution shift.
  } \label{fig:perturbation_plot_main_text}
\end{wrapfigure}
\paragraph{Probabilistic cosine similarities}
We qualitatively assessed the distribution obtained by ProbCosine on a randomly selected test example from the OfficeHome clipart domain, evaluating the mean and variance of the cosine similarity under increasing corruption in both image and text domains. Text corruption was introduced by randomly replacing characters with `x', and image corruption by randomly adding grey squares. %
\cref{fig:perturbation_plot_main_text} shows the mean and variance of cosine similarities as corruption increases. We observe that the expected cosine similarity generally decreases and variance increases with more corruption, indicating that our approximation effectively captures model uncertainties under distribution shift.
 Note that we observe a slight increase in the cosine similarity after one character has been replaced, indicating that performing predictions solely on the expected cosine similarity can be problematic.
 In this case, the variance over cosine similarities can capture the change in the input, highlighting the importance of capturing and propagating the model uncertainties.

\color{black}

\paragraph{Number of data points for Hessian estimation} 
We evaluated how the number of samples affects Hessian estimation by computing the trace over 10 random subsets of LAION-400M. As shown in \cref{app:fig:hessian-nbr-samples} (\cref{app:hessian_trace}), the traces for both image and text projections quickly converge with low variance, suggesting that 10 mini-batches are sufficient for a stable estimate. \looseness-3

\begin{wrapfigure}{r}{0.48\textwidth}
    \setlength{\figurewidth}{.4\textwidth}
    \setlength{\figureheight}{.1\textwidth}
    \centering
    \vspace{-1em} 
    \pgfplotsset{%
        scale only axis,
        tick label style={font=\tiny}, axis lines = middle, axis line style={->}, 
        ylabel near ticks, 
        xlabel near ticks, 
        y tick label style={rotate=45}, 
        yticklabel style={/pgf/number format/fixed, 
                          /pgf/number format/precision=2,
                          /pgf/number format/fixed zerofill},
        scaled y ticks=false,
        xticklabels=\empty,
        every axis plot/.append style={thick},
        ylabel style={yshift=-0.1cm},
        xlabel style={yshift=+0.1cm},
        yticklabel shift={-0.15cm},
        xticklabel style={scale=.8},
        yticklabel style={scale=.8},
        xlabel={\scriptsize\# Datapoints ($i \times K$)},
        ylabel={\scriptsize$\frac{\tr(\bB_{i\times K})}{\tr(\bB_{5\times K})}$},
        xmin=0.8, xmax=5.2,
        ymin=0.96, ymax=1.02,
        xtick={1,2,3,4,5},
        xticklabels={1$\times$K,2$\times$K,3$\times$K,4$\times$K,5$\times$K},
    }
\begin{tikzpicture}

\begin{axis}[
height=\figureheight,
width=\figurewidth,
]

\addplot[
    scRed, dashed, very thick
]
    coordinates {
        (1,0.971)
        (2,0.978)
        (3,0.989) 
        (4,0.993) 
        (5,1.000) 
    };
    
\addplot[
    scRed, name path=A, thin
]
    coordinates {
        (1,0.971 + 0.0436)
        (2,0.978 + 0.0140)
        (3,0.989 + 0.0117)
        (4,0.993 + 0.0233)
        (5,1.000 + 0.0189)
    };

\addplot[
    scRed, name path=B, thin
]
    coordinates {
        (1,0.971 - 0.0436)
        (2,0.978 - 0.0140)
        (3,0.989 - 0.0117)
        (4,0.993 - 0.0233)
        (5,1.000 - 0.0189)
    };

\addplot [scRed, opacity=0.1] fill between [of=A and B];

\addplot[
    scYellow, name path=A, thin
]
    coordinates {
        (1,0.991 + 0.0163)
        (2,0.998 + 0.00847)
        (3,0.998 + 0.0108)
        (4,1.002 + 0.00798)
        (5,1.000 + 0.00549)
    };
    
\addplot[
    scYellow, name path=B, thin
]
    coordinates {
        (1,0.991 - 0.0163)
        (2,0.998 - 0.00847)
        (3,0.998 - 0.0108)
        (4,1.002 - 0.00798)
        (5,1.000 - 0.00549)
    };
    
\addplot[
    scYellow, very thick, dash dot
]
    coordinates {
        (1,0.991) 
        (2,0.998)
        (3,0.998)
        (4,1.002)
        (5,1.000)
    };
\addplot [scYellow, opacity=0.1] fill between [of=A and B];

\addplot[
    scCyan, name path=A, thin
]
    coordinates {
        (1,1.003 + 0.00712)
        (2,1.002 + 0.00520)
        (3,1.000 + 0.00379)
        (4,1.000 + 0.00360)
        (5,1.000 + 0.00282)
    };
    
\addplot[
    scCyan, name path=B, thin
]
    coordinates {
        (1,1.003 - 0.00712)
        (2,1.002 - 0.00520)
        (3,1.000 - 0.00379)
        (4,1.000 - 0.00360)
        (5,1.000 - 0.00282)
    };
    
\addplot[
    scCyan, very thick
]
    coordinates {
        (1,1.003) 
        (2,1.002) 
        (3,1.000) 
        (4,1.000) 
        (5,1.000) 
    }; 
    
\addplot [scCyan, opacity=0.1] fill between [of=A and B];

\end{axis}
\end{tikzpicture}     \vspace{-2.5em} 
    \caption{Relative trace of the image Hessian $B$-factor for varying base batch sizes
$K$ (2048 \legendline{scRed, dashed}, 8192 \legendline{scYellow, dash dot},
32768 \legendline{scCyan}) and 1–5 random batches.
Error bars show $\pm1$\,std over five trials.}
    \label{fig:ablation_batch_size_image}
\end{wrapfigure}
\paragraph{Number of negative samples}
We vary the batch size $K\!\in\!\{32768,8192,2048\}$ and estimate the posterior from 1–5 random batches, reporting mean$\pm$std over five trials.  
Since the posterior depends on negative samples only via the Hessian $\bB$ (\cf~\cref{eq:kfacggn}), we show the relative trace $\tr(\bB_{i\times K})/\tr(\bB_{5\times K})$, which is expected to be one.  
As observed in \Cref{fig:ablation_batch_size_image} and \cref{app:fig:ablation_batch_size} (\cref{app:subsec:batch_size}), a base batch size of $32768$ stays near $1$ across all batches with minimal variance, indicating stable estimates of the Hessian.

\section{Discussion \& Conclusion}\label{sec:conclusion}

In this work, we introduced a novel approach for post-hoc uncertainty estimation and propagation for large-scale vision language models (VLMs) such as CLIP \citep{radford2021learning} and SigLIP \citep{zhai2023sigmoid}.
For this, we first formulated probabilistic models admissible to a Bayesian treatment and then utilised a post-hoc posterior approximation over the last layer of each encoder.
Moreover, we derived an analytic approximation of the distribution over cosine similarities for efficient uncertainty propagation.
Thus, our approach allows efficient and effective uncertainty quantification without any architectural changes or additional training.
We demonstrated the effectiveness of BayesVLM in zero-shot and active learning settings, showing improvements over baselines, and additionally assessed its robustness and efficiency, showing that BayesVLM is a valuable tool for reliable application of VLMs.\looseness-1

Beyond the settings considered in this work, BayesVLM is also applicable to several related problems. An interesting application is the detection of OOD or failure modes. Having access to uncertainties over the image and text embeddings directly enables such scenarios. For example, credible intervals obtained from our Bayesian treatment provide a principled signal for detecting distribution shift or unreliable predictions.
Another promising direction is uncertainty-aware retrieval. Since retrieval methods rely on similarity scores in a shared embedding space, uncertainties over the embedding projections can be incorporated into the retrieval process. This allows retrieval systems to more reliably detect cases where inputs fall outside the model’s training distribution.

\paragraph{Limitations}
The limitations of our approach are {\em(i)}~we need access to training data to estimate the Hessian, {\em(ii)}~we require that embeddings are Gaussian distributed, {\em(iii)}~ our method only utilises Bayesian projection layers, and {\em(iv)}~we assume independence between image and text projection parameters in the local curvature approximation.
Because training data for many VLMs are closed-source, we also assessed potential performance degradation when estimating the Hessian on proxy datasets and found that BayesVLM yields robust estimates. However, further research is needed in closed-source settings.

\clearpage
\subsection*{Reproducibility statement}
To ensure the reproducibility of our work, we have provided detailed information on our method and experimental setups.
We will discuss the respective details below.

\paragraph{BayesVLM method \& algorithms}
In addition to the details presented in the main text (\cref{sec:method}), we provided detailed derivations in \cref{app:derivations} of {\em (i)}~the likelihood function approximation in \cref{app:exact-model}, {\em (ii)}~the Laplace approximation used in our method in \cref{app:laplace_approximation}, and {\em (iii)}~the distribution over cosine similarities in \cref{app:distribution_cosine_similarity}.
Moreover, we provided algorithmic descriptions of our method in \cref{alg:BayesVLM-Precomputation} and \cref{alg:BayesVLM-Inference}, outlining the precomputation of BayesVLM and the forward inference.
Lastly, we presented detailed descriptions of the active learning algorithm used in our work in \cref{app:teachnical_details} and provided specific details on {\em (i)}~the targeted selection algorithm in \cref{app:subsec:targeted_selection}, {\em (ii)}~the acquisition functions used in this work in \cref{app:subsec:acquisition_functions}, and {\em (iii)}~the online Laplace updates in \cref{app:subsec:online_laplace}.

\paragraph{Experiments}
In addition to the details provided in the main text in \cref{sec:experiments}, we provided extensive additional information in \cref{app:experiment_details}.
Specifically, we {\em (i)}~detail information on the pre-trained models used in this work in \cref{app:subsec:exp_details_pretrained_models}, {\em (ii)}~ present detailed information on the Hessian estimation and respective hyper-parameters in \cref{app:subsec:exp_details_hessian_estimation}, and in
\cref{app:subsec:exp_details_hessian_param_estimation}, and {\em (iii)}~present details on the hyperparameters and setup of the active learning experiments in \cref{app:subsec:exp_details_active_learning}.
We also presented additional experiments and experimental results that extend beyond those presented in the main text in \cref{app:additional_results}.

\paragraph{Implementation}
The code for the experiments is available at: {\small \url{https://aaltoml.github.io/BayesVLM/}}. 
Models and precomputed Hessian estimates can be accessed at: {\small \url{https://huggingface.co/collections/aalto-ml/bayesvlm}}.

\subsubsection*{Acknowledgements}
AS, RL, and SM acknowledge funding from the Research Council of Finland (grant number 339730 and 362408). MT acknowledges funding from the Research Council of Finland (grant number 347279) and support from the Wallenberg AI, Autonomous Systems, and Software Program (WASP), funded by the Knut and Alice Wallenberg Foundation.  
MK and SM acknowledge funding from the Finnish Center for Artificial Intelligence (FCAI).
AB, SK, and ZA acknowledge partial funding by the ERC (853489 - DEXIM) and the Alfried Krupp von Bohlen und Halbach Foundation. SK thanks the International Max Planck Research School for Intelligent Systems (IMPRS-IS).
We acknowledge CSC -- IT Center for Science, Finland, for awarding this project access to the LUMI supercomputer, owned by the EuroHPC Joint Undertaking, hosted by CSC (Finland) and the LUMI consortium through CSC. We acknowledge the computational resources provided by the Aalto Science-IT project. 
Finally, we thank Riccardo Mereu, Jonas Hübotter, and Omar Eldeeb for providing feedback on the manuscript.

{
    \bibliographystyle{iclr2026_conference}

}

\appendix

\clearpage
\section*{Appendix}

The appendix is organized as follows:
\cref{app:notation} summarizes the notation used in the paper.
\cref{app:background} reviews background on vision–language models and the Laplace approximation.
\cref{app:derivations} presents the derivation of the posterior estimation and the efficient computation of distributions over cosine similarity.
\cref{app:teachnical_details} outlines the active learning setup.
\cref{app:experiment_details} details the experimental setup, while additional results are provided in \cref{app:additional_results}.

\paragraph{Use of Large Language Models}
In this paper, LLMs were used only for minor grammatical edits, word polishing, or rephrasing. They did not contribute to research ideation, experiments, or core writing. All suggestions from LLMs were manually verified and edited by the authors prior to final inclusion.

\section{Notation} \label{app:notation}
We will briefly summarise the notation used throughout the paper. See \cref{app:tab:notation_modality} for the modality-specific notation used and \cref{app:tab:notation_general} for an overview of the notation of general operands and operators.
\begin{table}[h]
  \centering
  \caption{Summary of modality-specific notation.}
  \begin{tabular}{@{}lcc@{}}
    \toprule
    Description & Image & Text \\
    \midrule
    Input & $\bx^\img$ & $\bx^\txt$ \\
    Encoder & $\phi(\cdot)$ & $\psi(\cdot)$ \\
    Projection matrix & $\bP$ & $\bQ$ \\
    Embedding & $\bg$ & $\bh$ \\
    Normalised embedding & $\hat{\bg}$ & $\hat{\bh}$ \\
    Stacked embeddings & $\bG$ & $\bH$ \\
    Kronecker factors & $\bA_\img, \bB_\img$ & $\bA_\txt, \bB_\txt$ \\
    Covariance matrix & $\bSigma_\img$ & $\bSigma_\txt$ \\
    Jacobian matrix & $\bJ_\img$ & $\bJ_\txt$ \\
    \bottomrule
  \end{tabular}
  \label{app:tab:notation_modality}
\end{table}
\begin{table}[h]
  \centering
  \caption{Summary of general notation.}
  \begin{tabular}{@{}lc@{}}
    \toprule
    Description & Notation \\
    \midrule
    Number of data points & $n$ \\
    Number of test data points & $n_{\text{test}}$ \\
    Number of support set points & $m$ \\
    Kronecker product & $\otimes$ \\
    Prior precision & $\lambda$ \\
    Pseudo-data count & $\tau$ \\
    \bottomrule
  \end{tabular}
  \label{app:tab:notation_general}
\end{table}

\section{Background} \label{app:background}
This section provides additional background information and an extended discussion of related work.

\subsection{Extended Related Work}\label{app:related_work}

\paragraph{Uncertainty in vision-language models} 
Many efforts have aimed to learn probabilistic embeddings by making architectural changes to the VLMs and pre-training with a probabilistic loss~\citep{pcmepp,prolip,chun2021probabilistic,ji2023map, li2022differentiable,neculai2022probabilistic}. 
To reduce training costs, several works have proposed enabling uncertainty estimation in pre-trained VLMs via additional training of adapters ~\citep{morales2024bayesadapter,upadhyay2023probvlm,lafon2025vilu}, learning distributions of prompts~\citep{cho2024make,lu2022prompt,yang2024bayesian}, model ensembles~\citep{miao2024bayesian}, or test-time adaptation~\citep{zhou2025bayesian}. These works use a proxy data set different from the pre-training set to learn the predictive uncertainties.  
Test-time augmentation is a training-free method used for obtaining input-dependent predictive uncertainties by augmenting the test input~\citep{ayhan2018test,farina2024frustratingly,shanmugam2021better}, which trades off simplicity against higher inference costs. 
Other recent training-free approaches focus on zero-shot out-of-distribution detection in CLIP~\citep{fu2025clipscope} or estimating the distribution on the hypersphere as a von-Mises Fisher distribution \citep{ju2025exploiting}.
Moreover, calibration of VLMs has been studied for mitigating overconfident predictions~\citep{tu2023a,tu2024empirical,yoon2024ctpt} where temperature scaling is a common post-hoc method for calibrating pre-trained models using a held-out validation set~\citep{galil2023can,guo2017calibration}. 
Here, we apply the Laplace approximation to estimate uncertainties directly from the pre-trained VLM without the need for additional training, architectural changes, or training from scratch. 
Our approach estimates a Bayesian posterior distribution with the pre-training data or a proxy data set before test time and has a similar inference speed to the pre-trained VLM. 

\paragraph{Active learning}
In active learning~\citep{ren2021survey,settles2009active}, the model determines through an acquisition function which additional data points are needed to make reliable predictions on a given downstream task. The acquisition function quantifies the informativeness of samples using entropy~\citep{holub2008entropy,safaei2025active,wang2014new} or diversity-based scores~\citep{ash2020deep,agarwal2020contextual}, coresets~\citep{sener2018active}, and parametric models~\citep{sinha2019variational,xie2023active}.
Here, we focus on acquisition functions utilising model uncertainties from Bayesian active learning~\citep{bickfordsmith2023prediction,gal2017deep,houlsby2011bayesian}. A popular method is the BALD score~\citep{gal2017deep,houlsby2011bayesian}, which measures the reduction in epistemic uncertainties of the model. More recently, EPIG was proposed to measure the information gain in the space of predictions rather than parameters~\citep{bickfordsmith2023prediction}, building on MacKay’s foundational work on information-theoretic experimental design \citep{mackay1992information}. While such acquisition functions have gained traction in large language models~\citep{hubotter2025efficiently}, they remain underexplored in VLMs, where ad-hoc strategies like prompt tuning \citep{bang2024active} are more prevalent. This work bridges this gap by adapting Bayesian active learning methods to VLMs.

\subsection{Language-image pre-training} \label{app:background_vlms}
We consider VLMs trained by minimising the InfoNCE loss \citep{oord2018representation} (\eg, CLIP~\citep{radford2021learning}) or the SigLIP loss \citep{zhai2023sigmoid}.
Specifically, the InfoNCE loss is defined as the sum of two cross-entropy terms, one for each relational direction---image to text ($\mcL_{\CE}(\bX^\img, \bX^\txt)$) and text to image ($\mcL_{\CE}(\bX^\txt,\bX^\img)$).
The total loss is defined as follows $\mcL_{\text{InfoNCE}}(\bX^\img, \bX^\txt)=$
\begin{equation}
	-\underbrace{\frac{1}{2n} \sum_{i=1}^n \log \frac{ \exp( t\hat{\bg}_i\transpose \hat{\bh}_i ) }{ \sum_{j=1}^n \exp(t\hat{\bg}_i\transpose \hat{\bh}_j) }}_{\imgtotxt,\,\mcL_{\CE}(\bX^\img, \bX^\txt)} - \underbrace{ \frac{1}{2n} \sum_{i=1}^n \log \frac{ \exp( t\hat{\bh}_i\transpose \hat{\bg}_i ) }{ \sum_{j=1}^n \exp(t\hat{\bh}_i\transpose \hat{\bg}_j) }}_{\txttoimg,\,\mcL_{\CE}(\bX^\txt, \bX^\img)},
\end{equation}
where $t$ is a learnable temperature parameter, $n$ denotes the number of image-text pairs, and $\hat{\bg}$ and $\hat{\bh}$ are the unit-length normalised embeddings.
This contrastive loss function encourages embeddings for matching image-text pairs to be similar while simultaneously pushing unrelated image-text pairs away from each other \citep{oord2018representation}.

Recently, the SigLIP loss \citep{zhai2023sigmoid} has been proposed as an alternative to the InfoNCE loss, aimed at improving numerical stability and training speed.
In contrast to InfoNCE, the SigLIP loss uses a binary classification loss over the cosine similarities, 
\ie, $\mcL_{\text{SigLIP}}(\bX^\img, \bX^\txt) =$
\begin{equation}
	 -\frac{1}{n}\sum_{i=1}^{n} \sum_{j=1}^{n} \log \frac{1}{1 + \exp(z_{ij} (-t \hat{\bg}_i\transpose \hat{\bh}_j + b))} , \label{eq:siglip}
\end{equation}
where $z_{ii} = 1$, $z_{ij} = -1$ if $i \neq j$ and $b$ is a learnable bias term.
For classification settings, the SigLIP loss does not provide normalised class conditional probabilities $p(y \mid \bx)$ but provides binary classification probabilities. 
Henceforth, when fine-tuning a SigLIP pre-trained VLM for classification tasks, one typically uses the cross-entropy loss instead.

\subsection{Laplace approximation}\label{app:background_laplace}
Given a data set $\mcD = \{(\bx_i, \by_i)\}_{i=1}^n$ and denote the model parameters as $\btheta$, in Bayesian deep learning, we aim to estimate the posterior distribution 
\begin{align}
	p(\btheta \mid \mcD) &= \frac{p(\btheta)  \prod_{i=1}^n p(\by_i \mid \bx_i, \btheta) }{\int_{\btheta} p(\btheta) \prod_{i=1}^n p(\by_i \mid \bx_i, \btheta) \dee \btheta}  \\
	&= \frac{\highlight{gray!20}{prior} \times \highlight{gray!20}{likelihood} }{\highlight{scRed!20}{marginal likelihood}} . \nonumber
\end{align}
Unfortunately, computing the denominator (marginal likelihood) is generally intractable (not feasible) as it requires integration over a high-dimensional space \wrt a potentially non-linear function.
A classical approach to circumvent this challenge is to approximate the posterior using a Laplace approximation \cite{mackay1992information}, which has recently gained traction in the Bayesian deep learning community~\cite{ritter2018scalable,daxberger2021laplace,li2024streamlining,meronen2024dnn,roy2022uncertainty,scannell2024functionspace}.

The Laplace approximation hinges on the idea that the posterior distribution is proportional to the joint, \ie,
\begin{equation}
	p(\btheta \mid \mcD) \propto p(\btheta, \mcD) = p(\btheta)  \prod_{i=1}^n p(\by_i \mid \bx_i, \btheta)  \label{eq:joint}
\end{equation}
up to an unknown normalisation constant (the marginal likelihood).
Moreover, using a second-order Taylor expansion of the log joint around the maximum-a-posteriori (MAP) estimate $\btheta_{\MAP}$ (mode of the function) one obtains the unnormalised log density function of a Gaussian centred at $\btheta_{\MAP}$, \ie, $\log p(\btheta, \mcD) \approx$
\begin{equation}
 \log p(\btheta_{\MAP}, \mcD) - \frac{1}{2}(\btheta - \btheta_{\MAP})\transpose \bSigma^{-1} (\btheta - \btheta_{\MAP}), 
\end{equation}
where 
\begin{equation}
	\bSigma = \rbra{-\nabla_{\btheta}^2 \log p(\btheta, \mcD) \rvert_{\btheta = \btheta_{\MAP}}}^{-1} = \rbra{-\nabla_{\btheta}^2 \log p(\mcD \mid \btheta) \rvert_{\btheta = \btheta_{\MAP}} - \nabla_{\btheta}^2 \log p(\btheta) \rvert_{\btheta = \btheta_{\MAP}}  }^{-1} 
\end{equation}
is the Hessian matrix of the log joint (prior $\times$ likelihood) at $\btheta_{\MAP}$.
By matching the marginal likelihood in \cref{eq:posterior} with the normalisation constant of a Gaussian, we obtain the Laplace approximation: 
\begin{equation}
	p(\btheta \mid \mcD) \approx \mcN(\btheta_{\MAP}, \bSigma^{-1}) ,
\end{equation}
with covariance given by the inverse of the Hessian matrix.

As LA fits a Gaussian distribution to the posterior, centred at the MAP estimate of a \textit{pre-trained} model, it is `post-hoc'.
The \emph{prior} is implicitly defined by the L2 regularisation (weight decay) commonly used during training~\cite{radford2021learning,zhai2023sigmoid}, and corresponds to a diagonal Gaussian prior on the parameters, \ie, $p(\btheta)=\mcN(\mathbf{0}, \lambda^{-1} \bI)$. 
The \emph{likelihood} is defined by the training loss. 

\section{Derivations} \label{app:derivations}
This section provides detailed derivations of the equations presented in the main text. 
\cref{app:exact-model} discusses the setting where the \iid assumption is not made and the challenges associated with it.
\cref{app:laplace_approximation} discusses the \iid assumption, the resulting probabilistic model, and the derivations for estimating the posterior.
\cref{app:distribution_cosine_similarity} covers the derivations for efficient prediction, \ie, the distribution over cosine similarities. 

\begin{figure}[t!]
\begin{minipage}[t]{.48\textwidth}
\begin{algorithm}[H]
 \caption{Turn VLM into BayesVLM}\label{alg:BayesVLM-Precomputation}
 \begin{algorithmic}[1]
 \State \textbf{Input:} VLM encoders $\{\img, \txt\}$, training data $\mcD$
  \vspace{0.4em}\ForEach {encoder $\enc \in \{\img, \txt\}$}
    \State Compute $\bA_{\enc}$ factor with \cref{app:eq:afactor}
    \State Compute $\bB_{\enc}$ factor with \cref{app:eq:bfactor}
  \EndFor

  \State Find $\lambda$ by maximising the marginal likelihood (\cref{app:eq:marginal}) 
  \State \emph{(Optional)} Find optimal $\tau$ or set $\tau = 1$
  \ForEach {encoder $\enc \in \{\img, \txt\}$}
    \State Update $\widetilde{\bA}_{\enc} \gets \sqrt{\tau} \bA_{\enc} + \sqrt{\lambda}\bI$
    \State Update $\widetilde{\bB}_{\enc} \gets \sqrt{\tau} \bB_{\enc} + \sqrt{\lambda}\bI$
  \EndFor
  \State \textbf{Return:} $\{(\widetilde{\bA}_{\img}, \widetilde{\bB}_{\img}), (\widetilde{\bA}_{\txt}, \widetilde{\bB}_{\txt})\}$
 \end{algorithmic}
\end{algorithm}
\end{minipage}
\hfill
\begin{minipage}[t]{.48\textwidth}

\def\NoNumber#1{{\def\alglinenumber##1{}\State #1}\addtocounter{ALG@line}{-1}}
\begin{algorithm}[H]
 \caption{Compute Predictions}\label{alg:BayesVLM-Inference}
 \begin{algorithmic}[1]
  \State \textbf{Input:} BayesVLM, $(x_{\img},x_{\txt})$ 
  
  \vspace{0.4em}\noindent Compute embeddings using \cref{eq:feature_dist}, \ie,
  \State \quad $\mu_{\bg} \gets \bP_{\MAP}\,\phi(\bx_{\img})$
  \State \quad $\Sigma_{\bg} \gets \left(\phi(\bx^\img)^\top\widetilde{\bA}_{\img}^{-1}\phi(\bx^\img)\right) \widetilde{\bB}_{\img}^{-1}$
  \State \quad $\mu_{\bh} \gets \bQ_{\MAP}\,\psi(\bx_{\txt})$
  \State \quad $\Sigma_{\bh} \gets \left(\psi(\bx^\txt)^\top\widetilde{\bA}_{\txt}^{-1}\psi(\bx^\txt)\right) \widetilde{\bB}_{\txt}^{-1}$
  
  \vspace{0.4em}\noindent Apply ProbCosine, \ie,
  \State \quad Compute $ \EE[\cossim{\bg}{\bh}]$ with \cref{eq:cosine_expected}
  \State \quad Compute $ \Var[\cossim{\bg}{\bh}]$ with \cref{eq:cosine_variance}

  \vspace{0.4em}\noindent Apply probit approximation \citep{gibbs1998bayesian}, \ie,
  \State \textbf{Return:} $\softmax{\frac{t\,\EE[ \cossim{\bg}{\bh}]}{\sqrt{1 + \sfrac{\pi}{8} * t^2\,\Var[\cossim{\bg}{\bh}]}}}$

 \end{algorithmic}
 \end{algorithm}
\end{minipage}
\end{figure}

\subsection{What Happens without the \iid assumption} \label{app:exact-model}
In this section, we derive the Laplace approximation when we don't make the \iid assumption.
We will show this results in multiple computationally expensive or infeasible terms in the posterior covariance, and the posterior obtained by our \iid assumption keeps the computationally feasible term.

We start by reformulating the InfoNCE loss.
Given a dataset with $n$ image-text pairs $(\bx_i^\img, \bx_i^\txt)$, the InfoNCE loss is defined as $\mcL_{\text{InfoNCE}}(\bX^\img, \bX^\txt)=$
\begin{equation}
	-\underbrace{\frac{1}{2n} \sum_{i=1}^n \log \frac{ \exp( t\hat{\bg}_i\transpose \hat{\bh}_i ) }{ \sum_{j=1}^n \exp(t\hat{\bg}_i\transpose \hat{\bh}_j) }}_{\mcL^\img_{\CE}(\bX^\img, \bX^\txt)} - \underbrace{ \frac{1}{2n} \sum_{i=1}^n \log \frac{ \exp( t\hat{\bh}_i\transpose \hat{\bg}_i ) }{ \sum_{j=1}^n \exp(t\hat{\bh}_i\transpose \hat{\bg}_j) }}_{\mcL^\txt_{\CE}(\bX^\txt, \bX^\img)},
\end{equation}
where $t$ is a learnable temperature parameter, $\hat{\bg}$ and $\hat{\bh}$ are the unit-length normalised image and text embeddings. 
Evaluating this loss in practice is infeasible on billions of data points. 
Therefore, the common practice adopted in VLMs, such as CLIP, is to evaluate it on a sufficiently large batch. 
Specifically, denote a batch of image-text pairs as $\mcB = \{\bX^\img_{\mcB}, \bX^\img_{\mcB}\}$.
Then the InfoNCE loss over the whole data set is approximated by:
\[
\mcL_{\text{InfoNCE}}(\bX^\img, \bX^\txt) &\approx \sum_{\mcB} \mcL_{\text{InfoNCE}}(\bX^\img_{\mcB}, \bX^\txt_{\mcB}) .
\] 
For each batch, we can view the InfoNCE loss as two separate classification losses, one over image inputs and the other over text inputs.
To avoid clutter, we drop the temperature parameter from now on.
Looking at the loss for the image inputs $\mcL^\img_{\CE}(\bX^\img_{\mcB}, \bX^\txt_{\mcB})$, we can reformulate it as follows:
\[
\mcL^\img_{\CE}(\bX^\img_{\mcB}, \bX^\txt_{\mcB}) &= -\frac{1}{2|\mcB|} \sum_{i=1}^{|\mcB|}\log \frac{ \exp( \hat{\bg}_i\transpose \hat{\bh}_i ) }{ \sum_{j=1}^{|\mcB|} \exp(\hat{\bg}_i\transpose \hat{\bh}_j) } \\
&= -\frac{1}{2|\mcB|} \sum_{i=1}^{|\mcB|}\log \left[\operatorname{softmax}\left(\left[\hat{\bg}_i\transpose \hat{\bh}_1, \hat{\bg}_i\transpose \hat{\bh}_2, \ldots, \hat{\bg}_i\transpose \hat{\bh}_{|\mcB|}\right] \right)\right]_i \\
&= -\frac{1}{2|\mcB|} \sum_{i=1}^{|\mcB|}\log \left[\operatorname{softmax}\left(\hat{\bH}\hat{\bg}_i\right)\right]_i,
\]
where $\left[\operatorname{softmax}(\bz)\right]_i \triangleq \frac{\exp(z_i)}{\sum_j \exp(z_j)}$ is the $i$-th output of softmax function. We can see that the loss is equivalent to the cross-entropy loss on the following model, where label $\by^\img_i$ is a one-hot encoded vector with $i$-th element equal to one,
\begin{center}
	\centering
	\begin{tikzpicture}
 	\node [] (a) at (0,0) {$\bx^{\img}_i$};
 	\node [] (b) at (6,0) {$\hat{\bg}_i=\frac{\bP\phi(\bx^{\img}_i)}{\|\bP\phi(\bx^{\img}_i) \|}$};
 	\draw [->,shorten >=8pt,shorten <=8pt] (a) -- node [above,font=\small] {Image encoder $\phi(\cdot)$ and} node[below,font=\small] {image projection layer $\bP$} (b);	
 	\node [] (c) at (12,0) {$\hat{\bH}\hat{\bg}_i.$};
 	\draw [->,shorten >=8pt,shorten <=8pt] (b) -- node [above,font=\small] {use text embeddings $\hat{\bH}$} node[below,font=\small] {to compute logit} (c);
\end{tikzpicture}
\end{center}

Similarly, the text loss $\mcL^\txt_{\CE}(\bX^\txt, \bX^\img)$ can be viewed as cross-entropy loss on the following model where label $\by^\txt_i$ is a one-hot encoded vector with $i$-th element equal to one,
\begin{center}
	\centering
	\begin{tikzpicture}
 	\node [] (a) at (0,0) {$\bx^{\txt}_i$};
 	\node [] (b) at (6,0) {$\hat{\bh}_i=\frac{\bQ\psi(\bx^{\txt}_i)}{\|\bQ\psi(\bx^{\txt}_i) \|}$};
 	\draw [->,shorten >=8pt,shorten <=8pt] (a) -- node [above,font=\small] {Text encoder $\psi(\cdot)$ and} node[below,font=\small] {text projection layer $\bQ$} (b);	
 	\node [] (c) at (12,0) {$\hat{\bG}\hat{\bh}_i.$};
 	\draw [->,shorten >=8pt,shorten <=8pt] (b) -- node [above,font=\small] {use image embeddings $\hat{\bG}$} node[below,font=\small] {to compute logit} (c);
\end{tikzpicture}
\end{center}

Under this view, VLMs trained with the InfoNCE loss can be viewed as using the following equivalent model and loss:
\[
&f(\bx_{i}^\img, \bx_{i}^\txt \mid \bX_{\setminus i}^\img,  \bX_{\setminus i}^\txt, \btheta) = \left[\hat{\bH}\hat{\bg}_i, \hat{\bG}\hat{\bh}_i\right], \\
&\ell_i^{\img, \txt} = - \underbrace{\log [\text{softmax}\left(\hat{\bH}\hat{\bg}_i\right)]_i}_{\ell_i^\img} - \underbrace{\log [\text{softmax}\left(\hat{\bG}\hat{\bh}_i\right)]_i}_{\ell_i^\txt} \\
&\mcL^\img_{\CE}(\bX^\img_{|\mcB|}, \bX^\txt_{|\mcB|}) = \frac{1}{2|\mcB|} \sum_{i=1}^{|\mcB|} \ell_i^{\img, \txt}.
\]

Because data is only conditionally independent in this model, \ie,
\[
  (\bx_i^\img, \bx_i^\txt) \sim p(\bx_i^\img, \bx_i^\txt \mid \bX^\img_{\setminus i}, \bX^\txt_{\setminus i}, \btheta),
\]
the usual \iid assumption made in Bayesian models is violated. 
Note that performing Bayesian inference over non-\iid data in general settings is an active research field \citep{ralaivola09a}.
Nevertheless, we can still consider applying the Laplace approximation in this case.
Crucially, note that Laplace approximation is derived through a second-order Taylor approximation of the negative log joint $-\log p(\mcD \mid \btheta) p(\btheta)$, which only requires the negative log joint to be a twice-differentiable function. 
Therefore, we can still consider the Laplace for local posterior approximation at the MAP estimation. 
The interpretation of the underlying probabilistic model, however, may be more challenging in those cases.

We will now derive the negative log likelihood Hessian for the image projection layer $\bP$. 
Define shorthand $f_{\bP, \bQ}(\bx_i)=f(\bx_{i}^\img, \bx_{i}^\txt \mid \bX_{\setminus i}^\img,  \bX_{\setminus i}^\txt, \btheta)$, the GGN approximation for the Hessian over image projection layer $\bP$ is given as
\[
\frac{\partial^2 \ell^{\img, \txt}_i}{\partial^2 \bP} &\approx \frac{\partial f_{\bP, \bQ}(\bx_i)}{\partial \bP}\transpose \frac{\partial^2 \ell_i}{\partial^2 f_{\bP, \bQ}(\bx_i)} \frac{\partial f_{\bP, \bQ}(\bx_i)}{\partial \bP},
\]
where
\[
\frac{\partial f_{\bP, \bQ}(\bx_i)}{\partial \bP}\transpose &= \begin{bmatrix}
	\displaystyle \left(\frac{\partial \hat{\bH}\hat{\bg}_i}{\partial \bP}\right)\transpose & \displaystyle \left(\frac{\partial \hat{\bG}\hat{\bh}_i}{\partial \bP}\right)\transpose
\end{bmatrix}, \\ 
\frac{\partial^2 \ell_i^{\img, \txt}}{\partial^2 f_{\bP, \bQ}(\bx_i)} &= 
\begin{bmatrix}
	\displaystyle \frac{\partial^2 \ell_i^{\img, \txt}}{\partial^2 \hat{\bH}\hat{\bg}_i} & \displaystyle \frac{\partial^2 \ell_i^{\img, \txt}}{\partial \hat{\bH}\hat{\bg}_i \partial \hat{\bG}\hat{\bh}_i} \\[2em]
	\displaystyle \frac{\partial^2 \ell_i^{\img, \txt}}{\partial \hat{\bG}\hat{\bh}_i \partial \hat{\bH}\hat{\bg}_i } & \displaystyle \frac{\partial^2 \ell_i^{\img, \txt}}{\partial^2 \hat{\bG}\hat{\bh}_i }.
\end{bmatrix}
\]
When writing out the matrix multiplication, we have:
\[
\frac{\partial^2 \ell^{\img, \txt}_i}{\partial^2 \bP} &\approx \frac{\partial f_{\bP, \bQ}(\bx_i)}{\partial \bP}\transpose \frac{\partial^2 \ell_i}{\partial^2 f_{\bP, \bQ}(\bx_i)} \frac{\partial f_{\bP, \bQ}(\bx_i)}{\partial \bP} \\
 &= \underbrace{\left(\frac{\partial \hat{\bH}\hat{\bg}_i}{\partial \bP}\right)\transpose}_{\reals^{d \times |\mcB|}} \underbrace{\frac{\partial^2 \ell^{\img}_i}{\partial^2 \hat{\bH}\hat{\bg}_i}}_{\reals^{|\mcB| \times |\mcB|}} \frac{\partial \hat{\bH}\hat{\bg}_i}{\partial \bP} \\
 &\quad  + \textcolor{scRed}{\left(\frac{\partial \hat{\bH}\hat{\bg}_i}{\partial \bP}\right)\transpose \frac{\partial^2 \ell^{\txt}_i}{\partial^2 \hat{\bH}\hat{\bg}_i} \frac{\partial \hat{\bH}\hat{\bg}_i}{\partial \bP}} 
 + \textcolor{scRed}{\left(\frac{\partial \hat{\bG}\hat{\bh}_i}{\partial \bP}\right)\transpose  \frac{\partial^2 \ell_i^{\img, \txt}}{\partial \hat{\bG}\hat{\bh}_i \partial \hat{\bH}\hat{\bg}_i }\frac{\partial \hat{\bH}\hat{\bg}_i}{\partial \bP}} \\
  &\quad + \textcolor{scRed}{ \left(\frac{\partial \hat{\bH}\hat{\bg}_i}{\partial \bP}\right)\transpose  \frac{\partial^2 \ell_i^{\img, \txt}}{\partial \hat{\bH}\hat{\bg}_i \partial \hat{\bG}\hat{\bh}_i}\frac{\partial \hat{\bG}\hat{\bh}_i}{\partial \bP} }
 + \textcolor{scRed}{\left(\frac{\partial \hat{\bG}\hat{\bh}_i}{\partial \bP}\right)\transpose \frac{\partial^2 \ell_i^{\img, \txt}}{\partial^2 \hat{\bG}\hat{\bh}_i }\frac{\partial \hat{\bG}\hat{\bh}_i}{\partial \bP} }
\]
Here only the first term $\left(\frac{\partial \hat{\bH}\hat{\bg}_i}{\partial \bP}\right)\transpose \frac{\partial^2 \ell^{\img}_i}{\partial^2 \hat{\bH}\hat{\bg}_i} \frac{\partial \hat{\bH}\hat{\bg}_i}{\partial \bP}$ can be computed efficiently while terms in \textcolor{scRed}{red} are intractable or computationally expensive. 
The approximated posterior for $\bP$ obtained in our BayesVLM corresponds to dropping the computationally expensive or infeasible terms in the exact model.

\subsection{Estimating the posterior for BayesVLM with Laplace approximation} \label{app:laplace_approximation}
We now introduce the procedure for estimating the posterior of BayesVLM using the Laplace approximation in this section.
We start by introducing the \iid assumption we made and the resulting probabilistic model for BayesVLM in \cref{app:iid-assumption}.
Then, we give the derivation for the posterior approximation for BayesVLM in \cref{app:posterior-derivation}.

\subsubsection{\iid assumption and the resulting probabilistic model}\label{app:iid-assumption}
To efficiently estimate the approximated posterior using the Laplace approximation and obtain a clear probabilistic model underlying it, we assume two independent probabilistic models, one for each modality.
Specifically, for each modality, we assume data are \iid given the observations from the other modality:
\begin{equation}
  \bx^\img_i \distiid p(\bx^\img_i \mid \bX^\txt, \btheta), \qquad
  \bx^\txt_i \distiid p(\bx^\txt_i \mid \bX^\img, \btheta).
\tag{\iid assumption}
\end{equation}

Following this assumption, the image encoder $\phi(\cdot)$ and text encoder $\psi(\cdot)$ will become independent, and image projection layer $\bP$ and text projection layer $\bQ$ will become independent as well:
\begin{equation}
  \phi(\cdot) \indep \psi(\cdot), \quad \bP \indep \bQ \tag{Consequence from \iid assumption}.
\end{equation}

Under these assumptions, we can untangle the interaction between two modalities and approximate their respective likelihoods as categorical distributions. 

When the modalities become independent, for image input $\bx_i^\img$, we can only look at the image loss defined as 
\[
\mcL^\img_{\CE}(\bX^\img_{\mcB}, \bX^\txt_{\mcB}) &= -\frac{1}{2|\mcB|} \sum_{i=1}^{|\mcB|}\log \frac{ \exp( \hat{\bg}_i\transpose \hat{\bh}_i ) }{ \sum_{j=1}^{|\mcB|} \exp(\hat{\bg}_i\transpose \hat{\bh}_j) } \\
&= -\frac{1}{2|\mcB|} \sum_{i=1}^{|\mcB|}\log \left[\operatorname{softmax}\left(\left[\hat{\bg}_i\transpose \hat{\bh}_1, \hat{\bg}_i\transpose \hat{\bh}_2, \ldots, \hat{\bg}_i\transpose \hat{\bh}_{|\mcB|}\right] \right)\right]_i \\
&= -\frac{1}{2|\mcB|} \sum_{i=1}^{|\mcB|}\log \left[\operatorname{softmax}\left(\hat{\bH}\hat{\bg}_i\right)\right]_i,
\]
where $\left[\operatorname{softmax}(\bz)\right]_i \triangleq \frac{\exp(z_i)}{\sum_j \exp(z_j)}$ is the $i$-th output of softmax function. 
This corresponds to the cross-entropy loss on the following model, where label $\by^\img_i$ is a one-hot encoded vector with $i$-th element equal to one,
\begin{center}
	\centering
	\begin{tikzpicture}
 	\node [] (a) at (0,0) {$\bx^{\img}_i$};
 	\node [] (b) at (6,0) {$\hat{\bg}_i=\frac{\bP\phi(\bx^{\img}_i)}{\|\bP\phi(\bx^{\img}_i) \|}$};
 	\draw [->,shorten >=8pt,shorten <=8pt] (a) -- node [above,font=\small] {Image encoder $\phi(\cdot)$ and} node[below,font=\small] {image projection layer $\bP$} (b);	
 	\node [] (c) at (12,0) {$\hat{\bH}\hat{\bg}_i.$};
 	\draw [->,shorten >=8pt,shorten <=8pt] (b) -- node [above,font=\small] {\textbf{given} text embeddings $\hat{\bH}$} node[below,font=\small] {compute logit} (c);
\end{tikzpicture}
\end{center}
Therefore, for image input, the corresponding model is
\[
f(\bx_i^\img \mid \bX^\txt, \btheta) = \hat{\bH} \hat{\bg}_i,
\]
with the corresponding log likelihood
\[
\log p(\bX^\img \mid \bX^\txt, \btheta)  &= \log \prod_{i=1}^{n} p(\bx_i^\img \mid \bX^\txt, \btheta) \\
&= \log \prod_{i=1}^{n}  \left[\text{softmax}\left(\hat{\bH}\hat{\bg}_i\right)\right]_i.
\]
Similarly, for text input, the corresponding model is
\[
f(\bx_i^\txt \mid \bX^\img, \btheta) = \hat{\bG} \hat{\bh}_i,
\]
with the corresponding log likelihood
\[
\log p(\bX^\txt \mid \bX^\img, \btheta) &= \log \prod_{i=1}^{n} p(\bx_i^\txt \mid \bX^\img, \btheta) \\
&= \log \prod_{i=1}^{n} \left[\text{softmax}\left(\hat{\bG}\hat{\bh}_i\right)\right]_i.
\]
\emph{Why is this still a reasonable approximation?} For VLMs, it is important to capture interactions between modalities, and assuming independence seems problematic at first. However, as we are using a local post-hoc posterior estimation through the Laplace approximation, we are effectively introducing an independence conditionally on the MAP estimate of the (joint) contrastive loss. Thus, crucially, even though we assume independence between modalities, we can still capture interactions between modalities.
Note that this assumption is also important for computational reasons, as it helps us derive a computationally efficient approach.

\subsubsection{Posterior Approximation with LA}\label{app:posterior-derivation}
Now that we have a well-defined probabilistic model and likelihood, we apply the Laplace approximation to it.

\paragraph{Why only treat $\bP$ and $\bQ$ probabilistically} In the Laplace approximation, for the posterior covariance, we need to compute the Hessian of the log likelihood.
This is computationally infeasible for large models and large datasets, and a common approximation is Generalised Gauss–Newton (GGN) approximation~\citep{schraudolph2002fast}.
Use shorthand $f_{\btheta}(\bx)$ for the model and denote the log likelihood as $\ell(y, f_{\btheta}(\bx))$, the GGN approximates to the Hessian is given by 
\[
\nabla_{\btheta}^2  \ell(y, f_{\btheta}(\bx))  \approx  GGN(\btheta) \triangleq \frac{\partial f_{\btheta}(\bx)}{\partial \btheta}\transpose \frac{\partial^2 \ell(y, f_{\btheta}(\bx))}{\partial f_{\btheta}(\bx)^2}\frac{\partial f_{\btheta}(\bx)}{\partial \btheta} \label{eq:ggn-definition}
\]
Note that in GGN approximation, we need to compute the Jacobian of the model output \wrt to the model parameters $\frac{\partial f_{\btheta}(\bx)}{\partial \btheta}$.
This is computationally infeasible for image and text encoders due to the large number of output dimensions.
For image projection and text projection, this challenge can be bypassed as the Jacobian can be obtained analytically.
Therefore, we treat the vision and image encoder as fixed and apply the Laplace approximation only for the image projection and text projection $\bP$ and $\bQ$. 

\paragraph{KFAC GGN approximation to Hessian} To estimate the Hessian of the log likelihood for $\bP$ and $\bQ$, we use Kronecker-factored approximate curvature (KFAC), which expresses the Hessian as a Kronecker product of two smaller matrices. 
This significantly reduces computational and memory costs while preserving a richer posterior structure than diagonal approximations. 
Following \citep{ritter2018scalable}, the KFAC GGN approximation for $-\nabla^2_{\bP} \log p(\bX^\img \mid \bX^\txt, \bP)$ is 
\begin{equation}
		\underbrace{\left(\frac{1}{\sqrt{n}} \sum_{i=1}^n \phi(\bx_i^{\img}) {\phi(\bx_i^{\txt})}^{\top}\right)}_{\bA_{\img}}  \otimes \underbrace{\left(\frac{1}{\sqrt{n}}\sum_{i=1}^n \bJ_\img(\bx^\img_i) \transpose \bLambda_{\img} \, \bJ_\img(\bx^\img_i)\right)}_{\bB_{\img}},
\end{equation}
and the KFAC GGN approximation for $-\nabla^2_{\bQ} \log p(\mcD \mid \bQ)$ is
\begin{equation}
		\underbrace{\left(\frac{1}{\sqrt{n}} \sum_{i=1}^n \psi(\bx_i^{\txt}) {\psi(\bx_i^{\txt})}^{\top}\right)}_{\bA_{\txt}}  \otimes \underbrace{\left(\frac{1}{\sqrt{n}}\sum_{i=1}^n \bJ_\txt(\bx^\txt_i) \transpose\bLambda_{\txt} \, \bJ_\txt(\bx^\txt_i)\right)}_{\bB_{\txt}},
\end{equation}
where $\bJ_\img(\bx^\img_i)=\frac{\partial \hat{\bH}\frac{\bg_i}{\|\bg_i\|}}{\partial \bg_i}$ and $\bLambda_{\img}=\diag(\boldsymbol{\pi}) - \boldsymbol{\pi} \boldsymbol{\pi}\transpose$, with $\pi_c = \frac{\exp(f_c)}{\sum_{c'} \exp(f_{c'})}$, $\hat{\bg}_i\transpose \hat{\bh}_c \defines f_c$.

As estimating the Kronecker factors over billions of data is computationally infeasible, following \citep{ritter2018scalable}, we leverage a subset of the data and include a pseudo-data count $\tau$ to compensate for the reduced sample size.
Putting everything together, the posterior covariance over $\bP$ and $\bQ$ are approximated as 
\begin{equation}\label{eq:hessian_image}
    \bSigma_{\img} = \left(\tau(\bA_{\img} \otimes \bB_{\img}) + \lambda \bI \right)^{-1} \approx \underbrace{\left(\sqrt{\tau}\bA_{\img} + \sqrt{\lambda}\bI \right)^{-1}}_{\widetilde{\bA}_{\img}^{-1}} \otimes \underbrace{\left(\sqrt{\tau}\bB_{\img} + \sqrt{\lambda}\bI \right)^{-1}}_{\widetilde{\bB}_{\img}^{-1}},
\end{equation}
\begin{equation}\label{eq:hessian_text}
    \bSigma_{\txt} = \left(\tau(\bA_{\txt} \otimes \bB_{\txt}) + \lambda \bI \right)^{-1} \approx \underbrace{\left(\sqrt{\tau}\bA_{\txt} + \sqrt{\lambda}\bI \right)^{-1}}_{\widetilde{\bA}_{\txt}^{-1}} \otimes \underbrace{\left(\sqrt{\tau}\bB_{\txt} + \sqrt{\lambda}\bI \right)^{-1}}_{\widetilde{\bB}_{\txt}^{-1}}, 
\end{equation}
where the respective factors are given as:
\[
  \bA_\img &= \frac{1}{\sqrt{n}} \sum_{i=1}^n \phi(\bx^\img_i)\phi(\bx^\img_i)\transpose \nonumber \\
  \bA_\txt &= \frac{1}{\sqrt{n}} \sum_{i=1}^n \psi(\bx^\txt_i)\psi(\bx^\txt_i)\transpose , \label{app:eq:afactor}
\]
\[
  \bB_\img &= \frac{1}{\sqrt{n}} \sum^n_{i=1} \bJ_\img(\bx^\img_i)\transpose \bLambda_{\img} \, \bJ_\img(\bx^\img_i) \nonumber \\
  \bB_\txt &= \frac{1}{\sqrt{n}} \sum^n_{i=1} \bJ_\txt(\bx^\txt_i)\transpose \bLambda_{\txt} \, \bJ_\txt(\bx^\txt_i) , \label{app:eq:bfactor}
\]

\paragraph{Jacobian computation} Here we derive the Jacobians $\bJ_\img(\bx_i^\img)$ and $\bJ_\txt(\bx_i^\txt)$ used in the KFAC GGN approximation.

Recall $\hat\bg_i$ and $\hat\bh_j$ denote the normalized image and text embedding, respectively.
Let $\hat\bH$ denote the matrix of normalized text embeddings with $\hat\bh_j$ as its columns and $\hat\bG$ the matrix of normalized image embeddings with $\hat\bg_i$ as its columns.
Then, for the InfoNCE likelihood, which depends on the dot product between the normalised embedding in the batch, we compute the Jacobian for the image encoder as follows:
\[
\bJ_\img^\text{InfoNCE}(\bx_i^\img) &= \frac{\partial \hat{\bH}\hat{\bg}_i }{\partial \bg_i} \\
&= \hat{\bH} \frac{\partial }{\partial \bg_i} \frac{{\bg}_i }{\|{\bg}_i \|} \\
&= \hat{\bH} \frac{\|{\bg}_i\| - {\bg}_i \frac{\partial \|{\bg}_i\|}{\partial {\bg}_i }}{\|{\bg}_i \|^2} \\
&= \hat{\bH} \frac{\|{\bg}_i\| - \frac{{\bg}_i {\bg}_i\transpose}{\|{\bg}_i\|}}{\|{\bg}_i \|^2} \\
&= \hat{\bH} \left( \frac{\bone}{\|{\bg}_i \|} - \frac{{\bg}_i {\bg}_i\transpose}{\|{\bg}_i\|^3} \right) .
\]
Analogously, we obtain the Jacobian for the text encoder given as:
\[
\bJ_\txt^\text{InfoNCE}(\bx_i^\txt) = \hat{\bG}\left( \frac{\bone}{\|{\bh}_i \|} - \frac{{\bh}_i {\bh}_i\transpose}{\|{\bh}_i\|^3} \right) .
\]

For SigLIP, we obtain the following Jacobians:
\[
\bJ_\img^\text{SigLIP}(\bx_i^\img) &= \frac{\partial \hat{\bg}_i }{\partial \bg_i} = \left(\frac{\bone}{\|{\bg}_i \|} - \frac{{\bg}_i {\bg}_i\transpose}{\|{\bg}_i\|^3}\right) ,
\]
and
\[
J_\txt^\text{SigLIP}(\bx_i^\txt) &= \frac{\partial \hat{\bh}_i }{\partial \bh_i} = \left(\frac{\bone}{\|{\bh}_i \|} - \frac{{\bh}_i {\bh}_i\transpose}{\|{\bh}_i\|^3}\right) .
\]

\paragraph{Hessian of likelihood \wrt model output computation} Here we derive the loss Hessian \wrt model output $\bLambda_{\img}$ and $\bLambda_{\txt}$.
For InfoNCE loss used in CLIP, the zero-shot classifier induced computes unnormalised logits for each class $c$, represented by $\hat{\bg}_i\transpose \hat{\bh}_c \defines f_c$. 
By applying the softmax function, we calculate the probabilities for each class $c$ as $\pi_c = \frac{\exp(f_c)}{\sum_{c'} \exp(f_{c'})}$. 
The likelihood Hessian of the cross-entropy loss for this classifier is represented by
\[
\bLambda_{\img}^\text{InfoNCE} = \diag(\boldsymbol{\pi}) - \boldsymbol{\pi} \boldsymbol{\pi}\transpose .
\]
Similarly, the likelihood Hessian for the text encoder follows analogous principles in the text-to-image direction. 
For a more detailed derivation of the likelihood Hessian, we refer to~\cite[Ch.~3.5]{rasmussen2006gaussian}. 
Rearranging terms in the analytical expression for $\bJ_\img\transpose \bLambda_{\img}^\text{InfoNCE} \bJ_\img$ facilitates space-efficient computation of the GGN approximation.

The SigLIP loss is defined as follows
\begin{align}
&\mathcal{L}_{\text{SigLIP}}(\bX^\img, \bX^\txt) \\
&= -\frac{1}{n} \sum_{i=1}^{n} \sum_{j=1}^{n} \log \frac{1}{1 + \exp(-z_{ij} (t \hat{\bg}_i^\top \hat{\bh}_j + b))} \\
&= \frac{1}{n} \sum_{i=1}^{n} \sum_{j=1}^{n} \underbrace{-\log \sigma(a_{ij})}_{\coloneqq \ell(\hat{\bg}_i, \hat{\bh}_j)},
\label{eq:siglip}
\end{align}
where $\sigma(a) = \frac{1}{1 + e^{-a}}$ denotes the sigmoid function, and $a_{ij} \coloneqq z_{ij}(t \hat{\bg}_i^\top \hat{\bh}_j + b)$, with labels $z_{ij} \in \{-1, 1\}$, a learnable temperature scaling parameter~$t$, and a learnable bias~$b$.

In order to derive the loss Hessian $\bLambda^\text{SigLIP}$, we first derive the component-wise loss gradient of $\ell$:
\[
\frac{\partial}{\partial \hat {\bg}_k} \ell(\hat{\bg}_i, \hat{\bh}_j) &\overset{\mathrm{i \neq k}}{=}
 0 \\
\frac{\partial}{\partial \hat{\bg}_k} \ell(\hat{\bg}_i, \hat {\bh}_j) &\overset{\mathrm{i = k}}{=} \frac{\partial}{\partial \hat{\bg}_k} - \log \sigma(a_{ij}
) \\
&= -\frac{1}{\sigma(a_{ij})} \frac{\partial \sigma(a_{ij})}{\partial a_{ij}}\frac{\partial a_{ij}}{\partial \hat{\bg}_{i}} \\
&= \left(\sigma(a_{ij}) - 1\right)z_{ij}t\hat{\bh}_j,
\]
which we utilise to derive the component-wise loss Hessian
\[
\frac{\partial^2}{\partial \hat{\bg}_k \partial \hat{\bg}_k^\top} \ell(\hat{\bg}_i, \hat{\bh}_j) &\overset{\mathrm{i \neq k}}{=} 0 \\
\frac{\partial^2}{\partial \hat{\bg}_k \partial \hat{\bg}_k^\top} \ell(\hat{\bg}_i, \hat{\bh}_j) &\overset{\mathrm{i = k}}{=} \frac{\partial}{\partial \hat{\bg}_k^\top} \left(\sigma(a_{ij})z_{ij}t\hat{\bh}_j - z_{ij}t\hat{\bh}_j\right) \\
&= z_{ij} t \hat h_k \frac{\partial \sigma(a_{ij})}{\partial a_{ij}} \frac{\partial a_{ij}}{\partial \hat{\bg}_k^\top} \\
&= t^2 \sigma(a_{ij})\left(1 - \sigma(a_{ij})\right)\hat{\bh}_j \hat{\bh}_j^\top.
\]
Finally, the likelihood Hessian for the SigLIP loss $\mathcal{L}_{\text{SigLIP}}$ can be expressed as
\[
    \bLambda_\img^{\text{SigLIP}} &= \frac{\partial^2}{\partial \hat{\bg}_i \partial \hat{\bg}_i^\top} \mathcal{L}(\hat{\bg}_{1:n}, \hat{\bh}_{1:n}) \\
    &= \frac{1}{n}\sum_{j=1}^n\sum_{i=1}^n \frac{\partial^2}{\partial \hat{\bg}_i \partial \hat{\bg}_i^\top}\ell(\hat{\bg}_i, \hat{\bh}_j) \\
    &= \frac{t^2}{n} \sum_{j=1}^n \sigma(a_{ij})\left(1 - \sigma(a_{ij})\right) \hat{\bh}_j \hat{\bh}_j^\top
\]
for the image encoder and as 
\[
\bLambda_\txt^{\text{SigLIP}} = \frac{t^2}{n} \sum_{i=1}^n \sigma(a_{ij})\left(1 - \sigma(a_{ij})\right) \hat{\bg}_i \hat{\bg}_i^\top
\]
for the text encoder.

\paragraph{Efficient Hessian Computation}  
At first glance, computing the Hessian appears prohibitively expensive: the loss Hessian $\bLambda$ has shape $|\mathcal{B}| \times |\mathcal{B}|$ with $|\mathcal{B}| \approx 32\text{k}$, while the embedding dimension is much smaller ($d \approx 512$). Forming $\bLambda$ explicitly is therefore impractical. By exploiting its low-rank structure and contracting with the Jacobians, however, the computation can be carried out efficiently without ever materializing $\bLambda$, making the GGN approximation feasible even for large batches.  
For example, for the image encoder with the InfoNCE loss in CLIP, the GGN block simplifies to
\[
    & \bJ_\img(\bx^\img_i) \transpose \bLambda_{\img} \bJ_\img^\text{InfoNCE} \\
&= 
\underbrace{\left( \frac{\bone}{\|\bh_i \|} - \frac{\bh_i \bh_i^\top}{\|\bh_i\|^3} \right)\vphantom{\left( \diag(\boldsymbol{\pi}) - \boldsymbol{\pi}\boldsymbol{\pi}^\top \right)}}_{\scriptstyle \bM \in \reals^{d\times d}}
\;\underbrace{\hat{\bG}^\top\vphantom{\left( \diag(\boldsymbol{\pi}) - \boldsymbol{\pi}\boldsymbol{\pi}^\top \right)}}_{\scriptstyle \in \reals^{d \times |\mathcal{B}|}}
\;\underbrace{\left(\diag(\boldsymbol{\pi}) - \boldsymbol{\pi}\boldsymbol{\pi}^\top \right)\vphantom{\left( \diag(\boldsymbol{\pi}) - \boldsymbol{\pi}\boldsymbol{\pi}^\top \right)}}_{\scriptstyle \in \reals^{|\mathcal{B}|\times|\mathcal{B}|}}
\;\underbrace{\hat{\bG}\vphantom{\left( \diag(\boldsymbol{\pi}) - \boldsymbol{\pi}\boldsymbol{\pi}^\top \right)}}_{\scriptstyle \in \reals^{|\mathcal{B}|\times d}}
\;\underbrace{\left( \frac{\bone}{\|\bh_i \|} - \frac{\bh_i \bh_i^\top}{\|\bh_i\|^3} \right)\vphantom{\left( \diag(\boldsymbol{\pi}) - \boldsymbol{\pi}\boldsymbol{\pi}^\top \right)}}_{\scriptstyle \bM \in \reals^{d\times d}}\\
    &= \bM \left( \hat{\bG}^\top\diag(\boldsymbol{\pi})\hat{\bG} - \hat{\bG}^\top \bpi\bpi^\top \hat{\bG} \right) \bM  \\
    &= \bM \bigg( \underbrace{\vphantom{(}\hat{\bG}^\top}_{\in \reals^{d \times |\mcB|}} \underbrace{(\bpi \odot \hat{\bG})}_{\in \reals^{|\mcB| \times d}} -  \underbrace{(\hat{\bG}^\top \bpi)}_{\in \reals^{d \times |\mcB|}} \underbrace{(\hat{\bG}^\top \bpi)^\top}_{\in \reals^{|\mcB| \times d}} \bigg) \bM 
\]
where $\odot$ denotes row-wise scaling of $\hat{\bG}$ by the vector $\bpi$.

\paragraph{Marginal likelihood}
To learn the prior precision parameter $\lambda$, we follow prior work (\eg, \citep{immer2021scalable}) and optimise the log marginal likelihood within each probabilistic model.
For the image projection layer $\bP$, denote the prior and posterior as below:
\[
&\text{prior}: \mathcal{N}(\boldsymbol{0}, \lambda_{\img} \bI) \\
&\text{posterior}: \mathcal{N}(\bP_{\text{MAP}}, \bSigma_{\img}) 
\]
The marginal likelihood is 
\begin{align}
  \log p(\bX^\img \mid \bX^\txt) \approx & \sum_{i=1}^n \log p(\bx_i^\img \mid \bX^\txt, \bP_{\text{MAP}}) \\
  &\qquad - \frac{1}{2} \left(\bP_{\text{MAP}}\transpose \lambda \bI \bP_{\text{MAP}}
  - \log \det (\bSigma_{\img}) + \log \det (\lambda_{\img} \bI) \right) \label{app:eq:marginal}
\end{align}

We can learn the prior precision $\lambda_{\img}$ using gradient-based optimisation.

\paragraph{Distribution over image and vision features}
For completeness, we will briefly derive the distribution over image and vision features.
In particular, for the image encoder let $\bP \sim \distMatrixNorm(\bP_{\text{MAP}}, \bB_{\img}^{-1}, \bA_{\img}^{-1})$, then:
\[
\bg &= \bP\phi(\bx^\img) \\ 
\intertext{with $\bP\phi(\bx^\img) \sim$} 
 &\distMatrixNorm(\bP_{\MAP}\phi(\bx^\img), \bB_{\img}^{-1}, \phi(\bx^\img)^\top\bA_{\img}^{-1}\phi(\bx^\img)) \nonumber \\
\bg \sim &\distNorm(\bP_{\MAP}\phi(\bx^\img), \left(\phi(\bx^\img)^\top\bA_{\img}^{-1}\phi(\bx^\img)\right) \bB_{\img}^{-1}) .
\]

\subsection{Distribution over cosine similarities} \label{app:distribution_cosine_similarity}
For the derivation of the distribution over cosine similarities, first recall the definition of the cosine similarity between two vectors, $\bg$ and $\bh$, which is given as $\cossim{\bg}{\bh} = \frac{\bg\transpose\bh}{\|\bg\|\|\bh\|}$.
Now, let $\bg$ and $\bh$ denote random vectors for the image and text embeddings, respectively.
Further, let us assume that their distribution follows a Gaussian distribution with mean $\bmu_{\bg} = (\mu_{\bg,1}, \dots, \mu_{\bg,d})$ and $\bmu_{\bh} = (\mu_{\bh,1}, \dots, \mu_{\bh,d})$ and diagonal covariance structure, \ie, $\bSigma_{\bg} = \diag(\sigma^2_{\bg,1}, \dots, \sigma^2_{\bg,d})$ and $\bSigma_{\bh} = \diag(\sigma^2_{\bh,1}, \dots, \sigma^2_{\bh,d})$.

Then the expected value of the cosine similarity is:
\[
	\EE[ \cossim{\bg}{\bh} ] &= \frac{ \EE[\bg\transpose\bh]  }{ \EE[\|\bg\|] \EE[\|\bh\|] }	 \\
  &= \frac{\sum^d_i \mu_{\bg,i}\mu_{\bh,i}}{ \EE[\|\bg\|] \EE[\|\bh\|] } .
\]
Note that computing $\EE[\|\bx\|]$ is intractable, and we, therefore, bound the expected value by application of the triangle inequality, \ie, 
\begin{equation}
  \EE[\|\bx\|] \leq \sqrt{\sum_{i} \mu^2_{\bx,i} + \sigma^2_{\bx,i} } ,
\end{equation}
where we use the fact that $\EE[x^2] = \mu^2_x + \sigma^2_x$.
Consequently, we obtain an approximation to the expected value of the cosine similarity given by:
\begin{equation}
	\EE[ \cossim{\bg}{\bh} ] \approx \frac{\sum^d_i \mu_{\bg,i}\mu_{\bh,i}}{\sqrt{\sum_i \mu^2_{\bg,i} + \sigma^2_{\bg,i} }\sqrt{\sum_i \mu^2_{\bh,i} + \sigma^2_{\bh,i} }}	.
\end{equation}
Next, we will derive the second moment (variance) of the cosine similarity of two random vectors.
First, note that the variance can be written as the difference between two expectations, \ie,
\begin{equation}
	\Var[\cossim{\bg}{\bh}] = \EE[\cossim{\bg}{\bh}^2] - \EE[\cossim{\bg}{\bh}]^2 ,
\end{equation}
where the second expectation corresponds to:
\begin{equation}
  \EE[\cossim{\bg}{\bh}]^2 \approx \frac{(\sum^d_i \mu_{\bg,i}\mu_{\bh,i})^2}{\sum_i \mu^2_{\bg,i} + \sigma^2_{\bg,i} \sum_i \mu^2_{\bh,i} + \sigma^2_{\bh,i} } .
\end{equation}
Next we can obtain $\EE[ \cossim{\bg}{\bh}^2 ]$ for which we will use the fact that $\EE[x^2] = \mu^2_x + \sigma^2_x$ again, \ie,
\begin{equation}
  \EE[ \cossim{\bg}{\bh}^2 ] = \frac{\EE[(\bg\transpose \bh)^2]}{\sum_i \mu^2_{\bg,i} + \sigma^2_{\bg,i} \sum_i \mu^2_{\bh,i} + \sigma^2_{\bh,i} } 
\end{equation}
where
\[
  \EE[(\bg\transpose \bh)^2] &= \sum_i\sum_{j} \mu_{\bg,i}\mu_{\bh,i}\mu_{\bg,j}\mu_{\bh,j} \\
	&+ \sum_i \sigma^2_{\bg,i}\mu^2_{\bh,i} + \mu^2_{\bg,i}\sigma^2_{\bh,i} + \sigma^2_{\bg,i}\sigma^2_{\bh,i} .
\]
Henceforth, we obtain the variance:
\begin{equation}
  \Var[\cossim{\bg}{\bh}] = \frac{\sum_i \sigma^2_{\bg,i}(\sigma^2_{\bh,i} + \mu^2_{\bh,i})+\sigma^2_{\bh,i}\mu^2_{\bg,i}}{\sum_i \mu^2_{\bg,i} + \sigma^2_{\bg,i} \sum_i \mu^2_{\bh,i} + \sigma^2_{\bh,i} } .
\end{equation}

\section{Active learning details} \label{app:teachnical_details}
We provide additional details on our active learning setup. Active learning provides a natural setting to evaluate the quality of uncertainty estimates, as it relies on selecting informative samples based on predictive uncertainty. We assess BayesVLM in this setting using acquisition functions from Bayesian active learning, combined with adaptive target region selection. Concretely, given a query set $\mathcal{X}_{\text{test}} = \{x_i^\star\}_{i=1}^{n_{\text{test}}}$ of unseen samples with unknown class labels, our goal is to select a support set $\{(x_j, y_j)\}_{j=1}^{m}$ of labeled examples such that predictive uncertainty on $\mathcal{X}_{\text{test}}$ is reduced. To this end, we first target the selection process toward the predictive distribution of the query set, and then select support candidates based on their estimated influence on predictive or model uncertainty.

We detail our method in three parts: \cref{app:subsec:targeted_selection} describes how we reduce the candidate pool by selecting samples that align with the target distribution; \cref{app:subsec:acquisition_functions} outlines the acquisition functions used for (targeted) active fine-tuning; and \cref{app:subsec:online_laplace} explains how we update the Laplace approximation in an online fashion during the EPIG acquisition process.

\begin{figure*}
	\begin{tikzpicture}
		\begin{axis}[
			no markers, 
			height=5cm, width=\textwidth,
			axis lines=left,
			axis line style={draw=black!60},
			xtick=\empty, ytick=\empty,
			xticklabels={,,}, yticklabels={,,},
			ylabel near ticks,
			xlabel near ticks,
			xlabel={$x$}, ylabel={$y$},
			ymin=0, ymax=5,
			]
			
			\draw[rectangle, draw] (axis cs:1.5,0.5) rectangle (axis cs:6.5,4);

			\addplot+[black!50, thick, smooth] coordinates {(0,1) (1,1) (2, 2) (4, 1) (6,3) (7,3) (9,4) (10,4)};
			\addplot+[black!50, thick, smooth, dashed,name path=A] coordinates {(0,1.5) (1,1.1) (2, 2.5) (4, 1.1) (6,3.5) (7,3.1) (9,4.5) (10,5)};
			\addplot+[black!50, thick, smooth, dashed,name path=B] coordinates {(0,0.5) (1,0.9) (2, 1.5) (4, 0.9) (6,2.5) (7,2.9) (9,3.5) (10,3)};
			\addplot[black!10] fill between[of=A and B];

			\coordinate (x1) at (axis cs:2,0);
			\coordinate (x2) at (axis cs:3,0);
			\coordinate (x3) at (axis cs:5,0);
			\coordinate (x4) at (axis cs:6,0);

			\coordinate (s1) at (axis cs:1.7,0);
			\coordinate (s2) at (axis cs:2.5,0);
			\coordinate (s3) at (axis cs:9.5,0);
			\coordinate (s4) at (axis cs:9,0);
			\coordinate (s5) at (axis cs:3.4,0);
			\coordinate (s6) at (axis cs:5.7,0);
			\coordinate (s7) at (axis cs:4.5,0);

			\draw[color=scCyan, dotted, thick] (x1) -- (axis cs:2,2);
			\draw[color=scCyan, dotted, thick] (x2) -- (axis cs:3,1.42);
			\draw[color=scCyan, dotted, thick] (x3) -- (axis cs:5,1.92);
			\draw[color=scCyan, dotted, thick] (x4) -- (axis cs:6,3);

			\draw[color=black!50, dotted] (s1) -- (axis cs:1.7, 2);
			\draw[color=black!50, dotted] (s2) -- (axis cs:2.5, 1.5);
			\draw[color=black!50, dotted, scRed] (s3) -- (axis cs:9.5, 4);
			\draw[color=black!50, dotted, scRed] (s4) -- (axis cs:9, 4);
			\draw[color=black!50, dotted] (s5) -- (axis cs:3.4, 1.2);
			\draw[color=black!50, dotted] (s6) -- (axis cs:5.7, 3);
			\draw[color=black!50, dotted] (s7) -- (axis cs:4.5, 1.5);

			\begin{scope}
				\path[clip] plot[smooth] coordinates{(axis cs:0,1.5) (axis cs:1,1.1) (axis cs:2, 2.5) (axis cs:4, 1.1) (axis cs:6,3.5) (axis cs:7,3.1) (axis cs:9,4.5) (axis cs:10,5) (axis cs:10,5) (axis cs:10,3) (axis cs:10,3) (axis cs:9,3.5) (axis cs:7,2.9) (axis cs:6,2.5) (axis cs:4, 0.9) (axis cs:2, 1.5) (axis cs:1,0.9) (axis cs:0,0.5) (axis cs:0,0.5) (axis cs:0,1.5)};

				\draw[color=scCyan, very thick] (axis cs:2,1) -- (axis cs:2,5);
				\draw[color=scCyan, very thick] (axis cs:3,1) -- (axis cs:3,5);
				\draw[color=scCyan, very thick] (axis cs:5,1) -- (axis cs:5,5);
				\draw[color=scCyan, very thick] (axis cs:6,1) -- (axis cs:6,5);

				\draw[color=black!50, thick] (axis cs:1.7,1) -- (axis cs:1.7,5);
				\draw[color=black!50, thick] (axis cs:2.5,1) -- (axis cs:2.5,5);
				\draw[color=black!50, thick] (axis cs:3.4,1) -- (axis cs:3.4,5);
				\draw[color=black!50, thick] (axis cs:4.5,1) -- (axis cs:4.5,5);
				\draw[color=black!50, thick] (axis cs:5.7,1) -- (axis cs:5.7,5);
				\draw[color=black!50, very thick, scRed] (axis cs:9,1) -- (axis cs:9,5);
				\draw[color=black!50, very thick, scRed] (axis cs:9.5,1) -- (axis cs:9.5,5);
			\end{scope}

			\node at (axis cs:4, 4.25) {\small Target Space};
			
		\end{axis}
		
		\foreach \x in {1,2,3,4} {
			\node[circle, fill=scCyan, minimum size=5.5pt, inner sep=0pt] at (x\x) {};
		}
		
		\foreach \x in {1,2,5,6,7} {
			\pic[thick, scBlue] at (s\x) {cross=2.5pt};
		}
		
		\foreach \x in {3,4} {
			\pic[thick, scRed] at (s\x) {cross=2.5pt};
		}
		
	\end{tikzpicture}
	\caption{Illustration of targeted support set selection. We aim to select an \textcolor{scBlue}{informative} support set that reduces the uncertainty over the predictions on the query set \protect\tikz{\protect\node at (0,0) [fill=scCyan, circle, thick,minimum size=5.5pt, inner sep=0pt] {};}. Only focusing on the epistemic uncertainties would not lead to a good selection, as we would select \textcolor{scRed}{uninformative} support set candidates \protect\tikz{\protect\pic[thick, scRed] at (0,0) {cross=2.5pt};} with high epistemic uncertainty. Hence, we target the selection process.}
	\label{app:fig:tageted_support_set_selection}
\end{figure*}

\subsection{Targeted selection}
\label{app:subsec:targeted_selection}
To target the active learning process towards relevant areas in the data space, we perform a $k$-nearest neighbours ($k$-NN) search around the test data.
The main idea behind our adaptive targeted region selection is illustrated in \cref{app:fig:tageted_support_set_selection}.

Specifically, we greedily acquire an intermediate candidate set $\mcT^\star \subseteq \trainset$ using $k$-NN selection based on the test set $\testset$.
For this, we need to compute a metric comparing the random feature projections. 
We assessed two different ways, first by computing the 2-Wasserstein distance between the distributions of the embeddings and the second by computing the expected cosine similarity based on \cref{app:distribution_cosine_similarity}.
Recall that for multivariate Gaussian distributions, the 2-Wasserstein distance exists in closed-form and is given as $W_2^2\rbra{\distNorm(\bmu_1, \bSigma_1), \distNorm(\bmu_2, \bSigma_2)}=$
\begin{equation}
    \|\bmu_1 - \bmu_2\|_2^2 + \tr\rbra{\bSigma_1 + \bSigma_2 - 2(\bSigma_1^{1/2}\bSigma_2\bSigma_1^{1/2})^{1/2}} ,
\end{equation}
where $\|\cdot\|_2$ denotes the Euclidean norm, $\tr(\cdot)$ is the trace operator, and $\bSigma^{1/2}$ is the matrix square root of $\bSigma$.
As computing the Wasserstein distance exactly is computationally and memory intensive due to the matrix square root, we approximate it by assuming both distributions to be isotropic.
Hence, simplifying to $W_2^2\rbra{\distNorm(\bmu_1, \bSigma_1), \distNorm(\bmu_2, \bSigma_2)}=$
\begin{equation}
    \sum_{i=1}^d (\mu_{1,i} - \mu_{2,i})^2 + \sigma^2_{1,i} + \sigma^2_{2,i} - 2\sigma_{1,i}\sigma_{2,i},
\end{equation}
where $\bSigma_1 = \diag(\sigma_{1,1}^2,\dots,\sigma_{1,d}^2)$ and $\bSigma_2$ is given respectively.

Based on a selected metric, we select the training samples closest to the test set in the joint embedding space, resulting in:
\[
\mcT = \bigcup_{\bg^\star \in \mcT^\star} {N_{k}(\bg^\star, \trainset)} ,
\]
with $N_{k}(\bg^\star, \trainset)$ denoting the set of $k$-nearest neighbours of $\bg^\star$ in the training set $\trainset$.
To ensure that we select $k$ distinct data points for each test sample, we perform an iterative search in which we discard already selected training samples and iteratively increase the search radius until $k$ distinct samples are found for each test datum. This process is illustrated in \cref{app:fig:nearest_neighbour_set_selection}.

\begin{figure*}[t]
	\centering
	
	\begin{tikzpicture}
		
		\newcommand{\drawTrainPoint}[3]{%
			\node[minimum width=0.4cm, minimum height=0.4cm, circle] (#1) at (#2, #3) {};
			\pic[thick] at (#2, #3) {cross=2.5pt};
		}

		\begin{axis}[
			no markers, 
			height=\textwidth*0.4, width=\textwidth*0.5,
			axis lines=left,
			axis line style={draw=black!60},
			xtick=\empty, ytick=\empty,
			xticklabels={,,}, yticklabels={,,},
			ylabel near ticks,
			xlabel near ticks,
			title={Stage 1: 1-NN Selection},
			xlabel={\small feature dimension $1$}, ylabel={feature dimension $2$},
			ymin=0, ymax=1,
			xmin=0, xmax=1,
			]
		\end{axis}
		
		\begin{scope}[xshift=1cm]
			\drawTrainPoint{x1}{1}{0.5}
			\drawTrainPoint{x2}{3}{1.75}
			\drawTrainPoint{x3}{2}{3.5}
			\drawTrainPoint{x4}{0.5}{2}

			\node (c1) at (2, 0.5) [fill, circle, thick, scRed] {};
			\node (c2) at (3, 3.5) [fill, circle, thick, scYellow] {};
			\node (c3) at (2.25, 2.7) [fill, circle, thick, scCyan] {};
			
			\draw[<->, opacity=1, scRed] (x1) -- (c1);
			\draw[<->, opacity=0.5, scRed] (x2) -- (c1);
			\draw[<->, opacity=0.1, scRed] (x3) -- (c1);
			\draw[<->, opacity=0.25, scRed] (x4) -- (c1);
			
			\draw[<->, opacity=0.1, scYellow] (x1) -- (c2);
			\draw[<->, opacity=0.5, scYellow] (x2) -- (c2);
			\draw[<->, opacity=1, scYellow] (x3) -- (c2);
			\draw[<->, opacity=0.25, scYellow] (x4) -- (c2);
			
			\draw[<->, opacity=0.2, scCyan] (x1) -- (c3);
			\draw[<->, opacity=0.75, scCyan] (x2) -- (c3);
			\draw[<->, opacity=1, scCyan] (x3) -- (c3);
			\draw[<->, opacity=0.5, scCyan] (x4) -- (c3);

			\draw[scRed, thick] (1,0.5) circle [radius=0.2cm];

			\draw[scYellow, thick] (2,3.5) circle [radius=0.2cm];

			\draw[scCyan, thick] (2,3.5) circle [radius=0.4cm];
			
			\node[align=center] (label1) at (0,3.5) {\textcolor{scRed}{\small Already} \\ \textcolor{scRed}{\small occupied}};
			\draw[-latex] (label1) -- (1.5,3.5);
			
		\end{scope}

		\begin{scope}[xshift=\linewidth*0.5]
			
			\begin{axis}[
				no markers, 
				height=\linewidth*0.4, width=\linewidth*0.5,
				axis lines=left,
				axis line style={draw=black!60},
				xtick=\empty, ytick=\empty,
				xticklabels={,,}, yticklabels={,,},
				ylabel near ticks,
				xlabel near ticks,
				title={Stage 2: 2-NN Selection},
				xlabel={feature dimension $1$}, ylabel={feature dimension $2$},
				ymin=0, ymax=1,
				xmin=0, xmax=1,
				]
			\end{axis}
			
			\begin{scope}[xshift=1cm]
				\drawTrainPoint{x1}{1}{0.5}
				\drawTrainPoint{x2}{3}{1.75}
				\drawTrainPoint{x3}{2}{3.5}
				\drawTrainPoint{x4}{0.5}{2}

				\node (c1) at (2, 0.5) [fill, circle, thick, scRed] {};
				\node (c2) at (3, 3.5) [fill, circle, thick, scYellow] {};
				\node (c3) at (2.25, 2.7) [fill, circle, thick, scCyan] {};
				
				\draw[<->, opacity=1, scRed] (x1) -- (c1);
				\draw[<->, opacity=0.5, scRed] (x2) -- (c1);
				\draw[<->, opacity=0.1, scRed] (x3) -- (c1);
				\draw[<->, opacity=0.25, scRed] (x4) -- (c1);
				
				\draw[<->, opacity=0.1, scYellow] (x1) -- (c2);
				\draw[<->, opacity=0.5, scYellow] (x2) -- (c2);
				\draw[<->, opacity=1, scYellow] (x3) -- (c2);
				\draw[<->, opacity=0.25, scYellow] (x4) -- (c2);
				
				\draw[<->, opacity=0.2, scCyan] (x1) -- (c3);
				\draw[<->, opacity=0.75, scCyan] (x2) -- (c3);
				\draw[<->, opacity=1, scCyan] (x3) -- (c3);
				\draw[<->, opacity=0.5, scCyan] (x4) -- (c3);

				\draw[scRed, thick] (1,0.5) circle [radius=0.2cm];

				\draw[scYellow, thick] (2,3.5) circle [radius=0.2cm];

				\draw[scCyan, thick] (3,1.75) circle [radius=0.2cm];
			\end{scope}
			
		\end{scope}
		
	\end{tikzpicture}
	
	\caption{Illustration of the nearest neighbour-based support set selection for adaptive targeted selection.
		The circles \protect\tikz{\protect\node at (0,0) [fill=black!30, circle, thick, minimum size=5.5pt, inner sep=0pt] {};} show test data points with uncertainty scores depicted through their colours: \textcolor{scRed}{high}, \textcolor{scYellow}{medium}, \textcolor{scCyan}{low}. 
		For each test datum we find the $k=1$ nearest neighbour from the support set candidates \protect\tikz[]{\protect\pic[thick] at (0,0) {cross=2.5pt};}.
		If the $k=1$ nearest neighbour is already selected, we increase $k$ for those with occupied neighbours and choose the second nearest neighbour, \ie, $k=2$. This recursion continues until every test datum has a selected support set candidate.
		The selected candidates are shown in coloured circles. Note that in the case of the \textcolor{scCyan}{blue} test datum, the closest support set candidate has already been chosen by the \textcolor{scYellow}{yellow}, and hence the second closest candidate is selected in the second stage. 
	} \label{app:fig:nearest_neighbour_set_selection}
\end{figure*}

\subsection{Acquisition functions} \label{app:subsec:acquisition_functions}
Given a labelled pool $\mcD_\text{train}$ and an unlabelled target set $\mcX_\text{test} = \{\bx \mid (\bx, y) \in \mcD_\text{test}\}$, the goal is to select $m$ maximally informative samples from $\mcD_\text{train}$ to reduce predictive uncertainty on $\mcX_\text{test}$. In this section, we provide a detailed explanation of the acquisition functions used for this purpose.

\paragraph{Naive random}
For the \textit{na\"ive random} acquisition function, we randomly sample $m$ data points from the train set $\mcD_{\text{train}}$ to form the support set $\mcS_{\text{ID}}$.

\paragraph{Targeted random}
For the \textit{targeted random} acquisition function, we randomly sample $m$ data points from the unlabelled test set $\mcX_{\text{test}}$ to form an intermediate support set $\mcT^*$.
According to~\cref{app:subsec:targeted_selection}, we then select the nearest neighbours to  $\mcT^*$ from the training set $\mcD_{\text{train}}$ based on the cosine similarity of the normalized image embeddings to form the support set $\mcT_{\text{t-ID}}$.

\paragraph{Targeted maximum entropy}
For the \textit{entropy} acquisition function, we compute the predictive entropy $\ent{y_i^* \mid \bx_i^*}$ for each data point $\bx_i^* \in \mcX_{\text{test}}$ and select the $m$ data points with the highest entropy. 
We use the predictive entropy on the MAP estimate of the model parameters to estimate the predictive entropy of the model:
\[
&\ent{y \mid \bx, \btheta_{\MAP}} \notag\\ 
&\quad= -\sum_{c=1}^{C} p(y = c | \bx, \btheta_{\MAP}) \log p(y = c | \bx, \btheta_{\MAP})
\]
According to~\cref{app:subsec:targeted_selection}, we then select the most similar data points from $\mcX_{\text{train}}$ to form the support set $\mcT_{\text{t-entropy}}$.

\paragraph{BALD}
We compute the BALD score~\citep{houlsby2011bayesian} for each data point in $\mcX_{\text{train}}$ and select the $m$ data points with the highest score.
The score is approximated using nested Monte Carlo sampling, as in~\citep{houlsby2011bayesian}.
\[
&\BALD(\bx) \\
&\quad= \EE_{p(y \mid \bx)}\sbra{ \ent{p(\btheta)} - \ent{p(\btheta \mid \bx, y)} } \\ \label{eq:bald}
&\quad= \EE_{p(\btheta \mid \mcD)} \sbra{ \ent{p(y \mid \bx, \btheta)} - \ent{p(y \mid \bx, \mcD)} }
\]
\paragraph{Targeted BALD}
We compute the BALD score~(\cref{eq:bald}) for each data point $\bx_i^* \in \mcX_{\text{test}}$ and select the $m$ data points with the highest score.
According to~\cref{app:subsec:targeted_selection}, we then select the most similar data points from $\mcX_{\text{train}}$ to form the support set $\mcT_{\text{t-BALD}}$.

\paragraph{EPIG}
The Expected Predictive Information Gain (EPIG) score~\citep{bickfordsmith2023prediction} calculates the expected mutual information between the model parameters and the predictive distribution resulting from the acquisition of a training data point.
This method is specifically designed to target relevant information, eliminating the need for a $k$-nearest neighbour search typically used in other acquisition functions.
The EPIG score is given by
\[
&\EPIG(\bx) \notag\\
&= \EE_{p_*(\bx^*) p(y \mid \bx)} 
    \big[\ent{p(y^* \mid \bx^*)} - \ent{p(y^* \mid \bx^*, \bx, y)}\big]\\
&= \EE_{p_*(\bx^*)}\sbra{\kl{p(y, y^* \mid \bx, \bx^*)}{ p(y \mid \bx) p(y^* \mid \bx^*)} } \\
&= \EE_{p_*(\bx^*)}\bigg[ \sum_{y\in \mcY}\sum_{y^*\in \mcY} p(y, y^* \mid \bx, \bx^*) \log\frac{p(y, y^* \mid \bx,\bx^*)}{p(y \mid \bx) p(y^* \mid \bx^*)}\bigg]
\]
where $p_{*}(\bx^{*})$ denotes the target input distribution. The EPIG score is approximated using Monte Carlo sampling, as detailed in\citep{bickfordsmith2023prediction}.
For the EPIG selection, we perform online updates to the model weights using the online Laplace as described in~\cref{app:subsec:online_laplace}.

\subsection{Online Laplace approximation}
\label{app:subsec:online_laplace}
We use an online Laplace approximation to efficiently update the posterior distribution over the image projection matrix $\bP$ during active learning. Instead of recomputing the posterior from scratch after each support set update, we incrementally refine both the MAP estimate and the Kronecker-factored Hessian approximation using the newly selected datapoint. Concretely, we perform a gradient step to update $\bP_{\text{MAP}}$, and adjust the Kronecker factors $\bA_{\img}$ and $\bB_{\img}$ based on the contribution of the new sample. This yields a computationally efficient approximation to the posterior over $\bP$ conditioned on the growing support set. Additionally, the prior precision can optionally be re-estimated after each update step, as commonly done in online Laplace methods~\citep{immer2021scalable,lin2023online}. In the following, we outline the structure of the Laplace approximation and describe how it is updated online during EPIG-based support set construction.
 
Recall that we obtain from our post-hoc Laplace approximation the Kronecker factorized Hessian approximation $\bH_{\img} \approx (\sqrt{\tau}\bA_{\img} + \sqrt{\lambda}\bI) \otimes (\sqrt{\tau}\bB_{\img} + \sqrt{\lambda}\bI)$ with
\[
  \bA_\img &= \frac{1}{\sqrt{n}} \sum_{i=1}^n \phi(\bx^\img_i)\phi(\bx^\img_i)\transpose \quad \text{and} \\
  \bB_\img &= \frac{1}{\sqrt{n}} \sum^n_{i=1} \bJ_\img(\bx^\img_i)\transpose \bLambda_{\img} \, \bJ_\img(\bx^\img_i),
\]
approximating a posterior distribution over the projection weights:
\[
\bP &\sim \mcM\mcN(\bP_\text{MAP}, \widetilde{\bB}_\img^{-1}, \widetilde{\bA}_\img^{-1}) \\
\bQ &\sim \mcM\mcN(\bQ_\text{MAP}, \widetilde{\bB}_\txt^{-1}, \widetilde{\bA}_\txt^{-1})
\]
with $\widetilde{\bA}$, $\widetilde{\bB}$ denoting the Kronecker factors after applying $\tau$ and $\lambda$.

Further, utilizing \cref{app:distribution_cosine_similarity} in combination with the generalized probit approximation (as described, for instance, in \citep{daxberger2021laplace}), we obtain an analytical form for the predictive posterior distribution $p(y \mid \bx, \mcD)$ of our few-shot classifier.

Our goal with EPIG is to iteratively construct a support set $\mcT_t$, where $t$ denotes the current number of selected training data points. We construct $\mcT_t$ by greedily selecting the training datum that maximises the expected information gain on the predictive distribution in the target domain:
\[
   \bx_{t+1} &= \argmax_{\bx \in \mcD_\text{train}} \EPIG(\bx \mid \mcT_{t}) \\
   &= \argmax_{\bx \in \mcD_\text{train}} \EE_{p_*(\bx^*) p(y \mid \bx)} 
    \big[\ent{p(y^* \mid \bx^*, \mcT_{t})} - \ent{p\left(y^* \mid \bx^*, \mcT_{t} \cup \{(\bx, y)\}\right)}\big]
\]
and obtain the corresponding label $y_{t+1}$, forming the support set $\mcT_{t+1} = \mcT_{t} \cup \{(\bx_{t+1}, y_{t+1})\}$.

The integration of $\{(\bx_{t+1}, y_{t+1})\}$ into the few-shot training set changes the posterior distribution over the image projection. To obtain the updated posterior distribution 
\[
\bP \mid \mcT_{t+1} &\sim \mcM\mcN(\bP_\text{MAP}, \widetilde{\bB}_{\img,t+1}^{-1}, \widetilde{\bA}_{\img,t+1}^{-1}),
\]
we utilise the following online updates to the projection weights and the Laplace approximation:
\[
\bP_{\text{MAP},t+1} &= \bP_{\text{MAP},t} - \gamma \grad_{\bP} \mcL(\bx^\img_{t+1}, \bX^\txt) \\
\bA_{\img, t+1} &= \frac{\sqrt{n+t}\bA_{\img, t} + \beta \bA_{\bx_{t+1}}}{\sqrt{n + t+1}} \\
\bB_{\img, t+1} &= \frac{\sqrt{n+t}\bB_{\img, t} + \beta \bB_{\bx_{t+1}}}{\sqrt{n + t+1}}.
\]
where $\gamma \geq 0$ and $\beta \geq 0$ are hyperparameters and
\[
\bA_{\bx_{t+1}} &= \phi(\bx^\img_{t+1})\phi(\bx^\img_{t+1})\transpose \\
\bB_{\bx_{t+1}} &= \bJ_\img(\bx^\img_{t+1})\transpose \bLambda_{\img} \bJ_\img(\bx^\img_{t+1}).
\]

\section{Experimental details} \label{app:experiment_details}
This section details the experimental setup used in our study. In \cref{app:subsec:exp_details_pretrained_models}, we describe the pre-trained vision-language models and checkpoints used. \cref{app:subsec:exp_details_hessian_estimation} explains how we computed the Hessian matrices required for the Laplace approximation. In \cref{app:subsec:exp_details_hessian_param_estimation}, we outline how we selected the Laplace parameters, such as the pseudo-data count~$\tau$ and prior precision~$\lambda$. \cref{app:subsec:exp_details_active_learning} describes our active learning setup, including dataset preparation, selection strategies, and training hyperparameters (see \cref{app:tab:finetuning_params}).

We run experiments on a compute cluster with NVIDIA P100 16GB, V100 32 GB, and A100 80GB GPUs. We used V100 or A100 GPUs for the Huge model variants and the ImageNet experiments. 

\subsection{Pre-trained vision-language models}\label{app:subsec:exp_details_pretrained_models}
In this work, we used the OpenCLIP \citep{ilharcoopenclip2021} implementations of CLIP \citep{radford2021learning}, which was published under the MIT license. We present additional experimental results on the HuggingFace implementation of SigLIP \citep{zhai2023sigmoid}, which was originally published under the Apache2 license.

For PCME++, we used the CLIP ViT-B/16 checkpoint provided by the authors with uncertainty adapter trained on CC-3M \citep{sharma2018conceptual}, CC-12M \citep{changpinyo2021cc12m}, and Redcaps \citep{desai2021redcaps}. For ProLIP, we used the ViT-B/16 model checkpoint released by the authors at \url{https://github.com/naver-ai/prolip}, which shares the same backbone architecture.

\subsection{Estimation of the Hessian matrices}\label{app:subsec:exp_details_hessian_estimation}
We estimated the Hessians separately for the CLIP image and text encoders using the pre-training dataset LAION-400M \citep{schuhmann2022laion} published under MIT license. For this estimation, we randomly sampled a subset of $327.680$ data points. The pre-training dataset was filtered to exclude NSFW content.
For the Laplace approximation, we used the GGN approximation of the Hessian matrices as described in \cref{app:laplace_approximation} and estimated the covariance matrices $\bA$ and $\bB$ for the image and text encoders. 

\subsection{Estimation of the Hessian parameters}\label{app:subsec:exp_details_hessian_param_estimation}

To ensure that both zero-shot and active learning experiments rely on a well-calibrated posterior covariance, we estimated the pseudo-data count~\citep{ritter2018scalable}, $\tau$, by performing a grid search over values in $\tau \in [1, 5, 10, 15, \ldots, 200]$. In the active learning experiments, the step size was reduced to $2$. The optimal value of $\tau$ was selected by minimizing the negative log predictive density (NLPD) on a random subset of ImageNet consisting of 100 classes and 1097 test data points in total as a proxy. \cref{app:fig:gridearch_pseudo_data_count} presents two plots illustrating the NLPD as a function of the pseudo-data count for SigLIP-Base and CLIP-Base, respectively. Once this optimal $\tau$ was identified, we further optimised the prior precision, $\lambda$, using the marginal likelihood on the LAION-400M~\citep{schuhmann2022laion} dataset.

\begin{figure}[h!]
    \label{app:fig:gridearch_pseudo_data_count}
    \centering
    \begin{tikzpicture}
        \begin{groupplot}[
            group style={
                group size=1 by 2, %
                horizontal sep=0.5cm, %
                vertical sep=0.5cm %
            },
            width=0.85\linewidth, 
            height=0.15\linewidth, 
            scale only axis,
            tick label style={font=\tiny}, 
            axis lines=middle, 
            axis line style={->},
            ylabel near ticks, 
            xlabel near ticks, 
            yticklabel style={rotate=45}, 
            xticklabel style={rotate=45},
            yticklabel style={/pgf/number format/fixed, /pgf/number format/precision=2},
            scaled y ticks=false,
            every axis plot/.append style={thick},
            ylabel style={yshift=-0.1cm},
            yticklabel shift={-0.15cm}
        ]

        \nextgroupplot[
            xmin=0,
            xmax=14,
            ylabel={\scriptsize NLPD $\leftarrow$},
        ]
        \addplot [thick, gray]
        table{

        1.0   1.617
        2.0   1.567
        3.0   1.556
        4.0   1.555
        5.0   1.557
        6.0   1.559
        7.0   1.562
        8.0   1.565
        9.0   1.567
        10.0  1.570
        11.0  1.572
        12.0  1.575
        13.0  1.577
        14.0  1.579
        };

        \nextgroupplot[
            xmin=5,
            xmax=50,
            ymax=1.1,
            ylabel={\scriptsize NLPD $\leftarrow$},
            xlabel={\scriptsize pseudo data-count $\tau$}
        ]
        \addplot [thick, gray, dashed]
        table{

         1     2.26920
         5     1.16064
        10     1.06922
        15     1.05145
        20     1.04659
        25     1.04542
        30     1.04619
        35     1.04699
        40     1.04786
        45     1.04871
        50     1.04953
        55     1.05029
        60     1.05029
        };
        \end{groupplot}
    \end{tikzpicture}

    \caption{Grid search over the pseudo-data count parameter $\tau$ for CLIP-Base \legendline{gray} and SigLIP-Base \legendline{gray, dashed}. The optimal NLPD for CLIP-Base is identified at $\tau = 10$, while the optimal NLPD for SigLIP-Base is identified at $\tau = 100$.}
\end{figure}

\subsection{Active learning experiments}\label{app:subsec:exp_details_active_learning}
We conducted active learning experiments on the OfficeHome data set \citep{venkateswara2017officehome}, which consists of the domains: art, clipart, product, and real-world, as well as on the ImageNet dataset with the domains ImageNet-R~\citep{hendrycks2021imagenetR} and ImageNet-Sketch~\citep{wang2019learning}. For these experiments, all training sets from the respective domains were combined into a single, large training set, and the projection layer of either CLIP or SigLIP was fine-tuned for a specific domain. Data selection was performed based on the acquisition functions described in \cref{app:subsec:acquisition_functions}. Performance was evaluated at the checkpoint corresponding to the lowest NLPD on a domain-specific validation set. We also performed a grid search for the online learning parameters of the EPIG acquisition rule, selecting the EPIG learning rate $\gamma$ from the range [1e-5, 1e-4, 1e-3, 1e-2] and the EPIG Hessian update scale $\beta$ from [1, 10, 100, 1000], based on the NLPD on the domain-specific validation set. Details on the training hyperparameter settings are given in \cref{app:tab:finetuning_params}.
\begin{table}[h]
  \centering
  \caption{Active fine-tuning hyperparameters.}
  \begin{tabular}{@{}ll@{}}
    \toprule
    config & value \\
    \midrule
    optimiser & AdamW \\
    learning rate & $1e{-5}$ \\
    weight decay & $5e{-2}$ \\
    optimiser momentum & $\beta_1, \beta_2 = 0.9, 0.999$\\
    batch size & $32$ \\
    epochs & $100$ \\
    \bottomrule
  \end{tabular}
  \label{app:tab:finetuning_params}
\end{table}

\section{Additional results} \label{app:additional_results}
In this Appendix, we provide additional results to support the findings in the main paper. Specifically, we detail (i) the approximation quality of the Gaussian approximation to the distribution over cosine similarities in \cref{app:subsec:approximation_quality}, (ii) the active learning experiments in \cref{app:subsec:active_learning_experiments}, (iii) the ablation of the $k$-NN distance metric in \cref{app:subsec:knn_ablation}, (iv) the influence of the number of data points used for Hessian estimation in \cref{app:hessian_trace}, (v) the runtime overhead and inference costs of BayesVLM compared to the baselines in \cref{app:runtime}, and (vi) additional zero-shot learning results in \cref{app:zero_shot_results} to demonstrate the generality of our approach.

\subsection{Approximation quality}\label{app:subsec:approximation_quality}
To assess the approximation quality of the Gaussian approximation to the distribution over cosine similarities, we generated $500$ samples for the image and text feature distributions for a given input. For the resulting samples, we then computed the respective cosine similarity for each pair and performed kernel density estimation with a Gaussian kernel and length scale of $0.3$ on the similarity scores. We added increasing shifts to the distribution mean to evaluate the change in the approximation quality under varying cosine similarity values. The results are depicted in \cref{fig:cosine_similarity}.
We can observe that our approximation through ProbCosine results qualitatively in a low approximation error.
\begin{figure}[h]
	\centering
	\pgfplotsset{
	tick label style={font=\scriptsize}, 
	axis lines = middle, 
	axis line style={->}, 
	xlabel near ticks, 
	every axis plot/.append style={thick},
	xlabel={\scriptsize Cosine Similarity},
	ylabel={\scriptsize Density},
	extra x ticks=0,
}

 	\caption{Approximation quality of the Gaussian approximation (ProbCosine) (\protect\tikz[baseline=-0.5ex]{\protect\draw[dashed, thick, scCyan] (0,0) -- (0.5,0);}) to the distribution over cosine similarities compared to KDE over samples (\protect\tikz[baseline=-0.5ex]{\protect\draw[thick, black] (0,0) -- (0.5,0);}) for image-text pairs with increasing Euclidean distance between their feature projection means ($\bmu_{\bg}, \bmu_{\bh}$).\looseness-1 }
	\label{fig:cosine_similarity}
\end{figure}

\subsection{Active learning experiments}\label{app:subsec:active_learning_experiments}
We report the active learning results for CLIP-Huge in~\cref{fig:results_activelearning_clip_huge} and for SigLIP-Base in~\cref{fig:results_activelearning_siglip_base}. For CLIP-Huge, we observe that active learning based on our post-hoc uncertainties consistently improves upon random and entropy-based selection. For SigLIP-Base, we observe significant improvements in terms of accuracy and NLPD on the ImageNet version with active learning based on our post-hoc uncertainties.

\subsection{Ablation of the $k$-NN distance metric}\label{app:subsec:knn_ablation}
We performed an ablation study on the $k$-NN distance metric for our proposed targeted selection in \cref{sec:active_learning} while fixing the \textit{online LA learning rate} $\gamma = 1\text{e}{-4}$ and the \textit{online LA pseudo-data count} $\beta = 10$. We evaluate performance using two distance metrics: Wasserstein and cosine similarity. Results are reported for EPIG and BALD, with Wasserstein (\textit{solid lines}) and cosine (\textit{dashed lines}) metrics. While Wasserstein-based k-NN selection demonstrates improved performance for datasets such as OfficeHome-Art and ImageNet-R, no clear trend is observed across the other datasets.

\subsection{Number of data points for Hessian estimation} \label{app:hessian_trace}
We assessed the influence of the number of data points used to estimate the Hessian by estimating the trace of the Hessian using a varying number of data points for 10 random subsets of the Laion400m data set.
Fluctuations of the estimated trace can be understood as an indicator that the Hessian estimates are unreliable.
\cref{app:fig:hessian-nbr-samples} shows that the trace of both the Hessian over the image projection and the text projection quickly converges to a stable value with low fluctuations, as seen by the low variance and stable mean. Moreover, the results indicate that 10 mini-batches suffice to obtain a reliable estimate of the Hessians.

\begin{figure}[h!]
\begin{minipage}[t]{0.5\columnwidth}
\centering
\scriptsize $\bSigma_\img$
\end{minipage}
\begin{minipage}[t]{0.49\columnwidth}
\centering
\scriptsize $\bSigma_\txt$
\end{minipage}
\resizebox{0.505\columnwidth}{!}{
\begin{tikzpicture}[scale=0.45]
	\begin{axis}[
		width=\columnwidth,
		height=0.5\columnwidth,
		outer sep=0pt,
        axis lines = middle, axis line style={->},
        tick label style={font=\tiny},
		ylabel={\scriptsize Trace}, ylabel near ticks,
        scale only axis,
        ylabel shift = 7pt,
		y tick label style={
			/pgf/number format/.cd,
			fixed,
			fixed zerofill,
			precision=2,
			/tikz/.cd},
		xlabel={\scriptsize \# data points = $x \times$ batch size}, xlabel near ticks,
        xtick={4096,8192,16384,32768,65536,98304,131072,163840,196608,229376,262144,294912,327680,360448,393216,425984,458752,491520},
		xticklabels={,,,1, 2, 3, 4, 5, 6, 7, 8, 9, 10, 11, 12, 13, 14, 15}, scaled x ticks=false,
	]
	    \addplot+[name path=upper,draw=none,mark=none] coordinates {
			(4096,31444471400)
			(8192,31374778781)
			(16384,31099206095)
			(32768,31360861485)
			(65536,31174831189)
			(98304,31098506232)
			(131072,31103753387)
			(163840,31121513423)
			(196608,31156789092)
			(229376,31169980464)
			(262144,31125074064)
			(294912,31113558714)
			(327680,31139895705)
			(360448,31073574521)
			(393216,31097483917)
			(425984,31050662992)
			(458752,31079142841)
			(491520,31080643153)
	    };
	    \addplot+[name path=lower,draw=none,mark=none] coordinates {
			(4096,30629063884)
			(8192,30405482502)
			(16384,30633489149)
			(32768,30859677190)
			(65536,30861971576)
			(98304,30849608711)
			(131072,30820646945)
			(163840,30874078667)
			(196608,30937995010)
			(229376,30918113642)
			(262144,30889878589)
			(294912,30944714874)
			(327680,30944959283)
			(360448,30953604043)
			(393216,30966182463)
			(425984,30961574300)
			(458752,30933510317)
			(491520,30989677589)
	    };
	    \addplot+[fill=scBlue!30] fill between[of=upper and lower];
	    \addplot+ [
	        scBlue, sharp plot, no marks, thick
	    ] coordinates {
			(4096,31036767642)
			(8192,30890130637)
			(16384,30866347622)
			(32768,31110269338)
			(65536,31018401382)
			(98304,30974057472)
			(131072,30962200166)
			(163840,30997796045)
			(196608,31047392051)
			(229376,31044047053)
			(262144,31007476326)
			(294912,31029136794)
			(327680,31042427494)
			(360448,31013584282)
			(393216,31031833190)
			(425984,31006120346)
			(458752,31006326579)
			(491520,31035160371)
	    };
	\end{axis}
\end{tikzpicture}
}
\resizebox{0.485\columnwidth}{!}{
\begin{tikzpicture}[scale=0.45]
	\begin{axis}[
		width=\columnwidth,
		height=0.5\columnwidth,
        outer sep=0pt,
		scale only axis,
        axis lines = middle, axis line style={->},
        tick label style={font=\tiny},
        ylabel = {},
        ylabel near ticks,
        ylabel shift = 5pt,
		y tick label style={
			/pgf/number format/.cd,
			fixed,
			fixed zerofill,
			precision=2,
			/tikz/.cd},
		xlabel={\scriptsize \# data points = $x \times$ batch size}, xlabel near ticks,
		xtick={4096,8192,16384,32768,65536,98304,131072,163840,196608,229376,262144,294912,327680,360448,393216,425984,458752,491520},
		xticklabels={,,,1, 2, 3, 4, 5, 6, 7, 8, 9, 10, 11, 12, 13, 14, 15}, scaled x ticks=false,
	]
	    \addplot+[name path=upper,draw=none,mark=none] coordinates {
			(4096,28675927550)
			(8192,28532102476)
			(16384,28299534493)
			(32768,28687949894)
			(65536,28460713530)
			(98304,28468145664)
			(131072,28468612958)
			(163840,28439123563)
			(196608,28406953611)
			(229376,28484236813)
			(262144,28423459659)
			(294912,28421501406)
			(327680,28444779746)
			(360448,28374607356)
			(393216,28408813983)
			(425984,28353633066)
			(458752,28463262330)
			(491520,28395501864)
	    };
	    \addplot+[name path=lower,draw=none,mark=none] coordinates {
			(4096,27934307638)
			(8192,27755964698)
			(16384,27750342089)
			(32768,27986239008)
			(65536,28137430265)
			(98304,28170245017)
			(131072,28189700669)
			(163840,28202206409)
			(196608,28248479300)
			(229376,28244561190)
			(262144,28151024821)
			(294912,28219244066)
			(327680,28216543211)
			(360448,28232429264)
			(393216,28289310714)
			(425984,28302735778)
			(458752,28232524780)
			(491520,28272075684)
	    };
	    \addplot[fill=scBlue!30] fill between[of=upper and lower];
	    \addplot+[
        scBlue, sharp plot, no marks, thick
        ] coordinates {
			(4096,28305117594)
			(8192,28144033587)
			(16384,28024938291)
			(32768,28337094451)
			(65536,28299071898)
			(98304,28319195341)
			(131072,28329156813)
			(163840,28320664986)
			(196608,28327716454)
			(229376,28364399002)
			(262144,28287242240)
			(294912,28320372736)
			(327680,28330661478)
			(360448,28303518310)
			(393216,28349062349)
			(425984,28328184422)
			(458752,28347893555)
			(491520,28333788774)
	    };
	\end{axis}
\end{tikzpicture}
}
\caption{Hessian (Trace) vs.\ number of samples. Error bars indicate $\pm 1$ standard deviation over 10 random subsets of Laion400m.}
\label{app:fig:hessian-nbr-samples}
\end{figure}

\subsection{Runtime overhead}
\label{app:runtime}

We compared our approach against the vanilla CLIP. We report runtimes on CIFAR-100 with CLIP-Huge/-Base on a Tesla V100 with 3 warmup steps and averaged over 1000 runs (batch size 1) in \cref{tab:runtime}. Indicating minimal computational overhead. 
In addition, we present inference costs in GFLOPs, comparing the original VLM (deterministic) against TTA and BayesVLM in \cref{tab:flops}.
We find that BayesVLM results in comparable inference costs, while TTA has an 80-fold increase in the inference cost.

\begin{table}[h]
    \centering
    \small
    \caption{Average runtime measured in seconds on CIFAR-100.}    \label{tab:runtime}
    \vspace*{-0.5em}
    \begin{tabular}{c|ro|c}
        \toprule
        Model & Vanilla & BayesVLM & rel. increase  \\ \midrule
        CLIP-Huge & 43.8178 & 43.9712 & 0.35\% \\
        CLIP-Bas & 9.4498 & 9.8929 & 4.69\% \\
        \bottomrule
    \end{tabular}
\end{table}

\begin{table}[h]
    \centering
    \small
    \caption{Inference computational cost per image on CIFAR-100 (in GFLOPs $10^9$).}
    \label{tab:flops}
    \vspace*{-0.5em}

\begin{tabular}{l|rro}
\toprule
Model & Vanilla & \sc TTA~\citep{farina2024frustratingly} & \sc BayesVLM \\
\midrule
CLIP-Base & 8.83 & 687.78 & 8.84 \\
CLIP-Lage & 162.06 & 12638.00 & 162.07 \\
CLIP-Huge & 334.71 & 26098.06 & 334.72 \\
SigLIP-Base & 47.00 & 3652.05 & 47.03 \\
\bottomrule
\end{tabular} \end{table}

\subsection{Interpreting probabilistic cosine similarities}
\label{app:subsec:interpreting_cossim}
We qualitatively assessed the distribution obtained by ProbCosine on a randomly selected test example from the OfficeHome clipart domain, evaluating the mean and variance of the cosine similarity under increasing corruption in both image and text domains. Text corruption was introduced by randomly replacing characters with `x', and image corruption by randomly adding grey squares. %
\cref{fig:perturbation_plot} shows the mean and variance of cosine similarities as corruption increases. We observe that the expected cosine similarity generally decreases and variance increases with more corruption, indicating that our approximation effectively captures model uncertainties under distribution shift.
 Note that we observe a slight increase in the cosine similarity after one character has been replaced, indicating that performing predictions solely on the expected cosine similarity can be problematic.
 In this case, the variance over cosine similarities can capture the change in the input, highlighting the importance of capturing and propagating the model uncertainties.

 \begin{figure}[h]
  \centering
  \resizebox{0.7\textwidth}{!}{
  \begin{tikzpicture}

    \def\nodewidth{10mm}

    \node[align=center, text width=12cm, font=\Large\strut] at (3.5, 1) {Expected Value};
    \node[align=center, text width=12cm, font=\Large\strut] at (10.5, 1) {Variance};

    \foreach \a [count=\i] in {0,1,2,3,4,5} {
      \node[minimum size=\nodewidth, inner sep=0pt, draw=white] at (\i, 0) {\includegraphics[width=\nodewidth]{figs/perturbation_plot/input_\a.jpg}};
    }

    \def\meanmaxval{30.66}
    \def\meanminval{17.77}
    \def\meanrangeval{(\meanmaxval-\meanminval)}

    \def\varmaxval{1.16}
    \def\varminval{0.72}
    \def\varrangeval{(\varmaxval-\varminval)}

    \def\rangeval{(\maxval-\minval)}

    \def\meanvalues{{{28.667324, 24.377195, 22.944693, 25.072762, 23.980007, 19.805763}, {30.662502, 28.089537, 27.190496, 30.056845, 29.621847, 24.638433}, {26.113857, 22.81041, 21.354885, 22.767145, 22.596605, 18.271564}, {28.06377, 24.358717, 22.911673, 24.593325, 24.247713, 19.534058}, {24.839317, 22.16473, 20.44263, 22.845757, 23.146006, 17.921112}, {19.88999, 20.707981, 19.58666, 21.409832, 21.706045, 17.771017}}}

    \def\variancevalues{{{0.72650445, 0.7961286, 0.8068778, 0.78780437, 0.8230217, 0.89644986}, {0.8148311, 0.8847842, 0.8980044, 0.8795547, 0.9149752, 0.98888284}, {0.94943386, 1.019018, 1.0361282, 1.0189433, 1.0542376, 1.1279718}, {0.9094317, 0.9791411, 0.99509305, 0.97752655, 1.0128663, 1.0866681}, {0.9604758, 1.0298133, 1.0472734, 1.0302707, 1.065446, 1.1389413}, {0.9873369, 1.0568107, 1.0750171, 1.0581907, 1.0934464, 1.167125}}}

    \foreach \y in {1,...,6} {
      \foreach \x in {1,...,6} {
        \pgfmathsetmacro\meanintensity{100*(\meanvalues[\x-1][\y-1]-\meanminval)/\meanrangeval}
        \node[fill=scCyan!\meanintensity, minimum size=\nodewidth, draw=gray!30] at (\y,-\x) {};
      }
    }
    
    \shade[left color=white, right color=scCyan] (0.5,-7.3) rectangle +(6,0.3);
    \node[below] at (0.75,-7.8) {\meanminval};
    \node[below] at (6.25,-7.8) {\meanmaxval};

    \foreach \a [count=\i from 8] in {0,1,2,3,4,5} {
      \node[minimum size=\nodewidth, inner sep=0pt] at (\i, 0) {\includegraphics[width=\nodewidth]{figs/perturbation_plot/input_\a.jpg}};
    }

    \foreach \y in {1,...,6} {
      \foreach \x in {1,...,6} {
        \pgfmathsetmacro\varintensity{100*(\variancevalues[\x-1][\y-1]-\varminval)/\varrangeval}
        \node[fill=scCyan!\varintensity, minimum size=\nodewidth, draw=gray!30] at (\y+7,-\x) {};
      }
    }

    \shade[left color=white, right color=scCyan] (7.5,-7.3) rectangle +(6,0.3);
    \node[below] at (7.75,-7.8) {\varminval};
    \node[below] at (13.25,-7.8) {\varmaxval};
  
    \foreach \a/\b [count=\i from 2] in {
      {An image of a cup},
      {A\textcolor{scRed}{x} image of a cup},
      {A\textcolor{scRed}{x} ima\textcolor{scRed}{x}e of a cup},
      {A\textcolor{scRed}{x} ima\textcolor{scRed}{x}e\textcolor{scRed}{x}of a cup},
      {A\textcolor{scRed}{x} ima\textcolor{scRed}{x}e\textcolor{scRed}{x}\textcolor{scRed}{x}f a cup},
      {A\textcolor{scRed}{x} ima\textcolor{scRed}{x}e\textcolor{scRed}{x}\textcolor{scRed}{x}f a\textcolor{scRed}{x}cup}} {
      \node[minimum size=\nodewidth, align=right, font=\Large] at (-1.5,-\i+1) {\a};
    }
  \end{tikzpicture}
  }
  \caption{Illustration of ProbCosine under increasing corruption. The mean similarity decreases and variance increases with higher levels of corruption, demonstrating effective uncertainty estimation under distribution shift.
  } \label{fig:perturbation_plot}
\end{figure}

\subsection{Additional zero-shot results}\label{app:zero_shot_results}

Here, we report additional zero-shot results for CLIP-huge and SigLIP-base in \cref{tab:uq_results_clip_large} and \cref{tab:uq_results_siglip}, respectively. We also report the results for applying our \emph{ProbCosine} to the ProLIP~\citep{prolip} model in \cref{tab:uq_results_prolip}.

\paragraph{CLIP-huge results} In \cref{tab:uq_results_clip_large},
BayesVLM consistently matches or outperforms baseline methods in terms of accuracy (ACC), negative log predictive density (NLPD), and expected calibration error (ECE). 
We note that TTA achieves worse ACC than CLIP on CIFAR-10 and -100, which we believe is due to test-time augmentation in \citep{farina2024frustratingly} not being optimised for small images (32x32), where the classes could become unrecognisable with additional cropping in the augmentation. 

\paragraph{SigLIP-base results} In \cref{tab:uq_results_siglip}, for SigLIP, BayesVLM still yields improvements on several benchmarks (despite using proxy data for Hessian estimation, as the original training set is not publicly available), demonstrating robustness to such settings. Despite mismatched training and Hessian estimation datasets, BayesVLM remains competitive, especially on CIFAR-10, UCF101, and SUN397, effectively improving calibration without sacrificing predictive performance. 
We note that TTA achieves better ACC on some benchmarks, which is sensible since the model will get better at predicting the correct class average with more chances (augmentations). 

\paragraph{ProbCosine combined with ProLIP results} In \cref{tab:uq_results_prolip}, we observe that applying ProbCosine to ProLIP improves zero-shot performance across classification benchmarks and metrics.

\begin{table*}[h]
  \centering
  \small  %
  \caption{{\bfseries Zero-shot Results:} Quantitative evaluation of uncertainty estimation across multiple data sets in the zero-shot setting for the OpenCLIP ViT-L-14 model. Our proposed BayesVLM consistently outperforms baseline methods across accuracy (ACC, in \%), negative log predictive density (NLPD), and expected calibration error (ECE, in \%) metrics.}
  \label{tab:uq_results_clip_large} 
  \vspace*{-0.5em}  
  \setlength{\tabcolsep}{9pt}  %
\resizebox{\textwidth}{!}{

\begin{tabular}{l|l|ccccccc}
\toprule
Metrics & Methods & \sc Flowers-102 & \sc Food-101 & \sc CIFAR-10 & \sc CIFAR-100 & \sc ImageNet-R & \sc UCF101 & \sc SUN397 \\
\hline
    \multirow{4}{*}{ACC $\uparrow$} & CLIP~\citep{radford2021learning} & \val{\bf}{72.04}{0.5723} & \val{}{86.60}{0.2144} & \val{\bf}{95.57}{0.2058} & \val{\bf}{76.74}{0.4225} & \val{}{85.51}{0.4065} & \val{\bf}{69.60}{0.7479} & \val{}{71.48}{0.3205}\\
 & CLIP (temp. scaling) & \val{\bf}{72.04}{0.5723} & \val{}{86.60}{0.2144} & \val{\bf}{95.57}{0.2058} & \val{\bf}{76.74}{0.4225} & \val{}{85.51}{0.4065} & \val{\bf}{69.60}{0.7479} & \val{}{71.48}{0.3205}\\
 & TTA~\citep{farina2024frustratingly}  & \val{}{71.85}{0.5735} & \val{\bf}{87.11}{0.2109} & \val{}{92.83}{0.2580} & \val{}{71.56}{0.4511} & \val{\bf}{87.64}{0.3800} & \val{\bf}{70.10}{0.7443} & \val{\bf}{71.99}{0.3187}\\
 \rowcolor{scCyan!30}
\cellcolor{white} & BayesVLM
  & \val{\bf}{72.42}{0.4469}   %
  & \val{\bf}{87.20}{0.3341}   %
  & \val{\bf}{95.49}{0.2075}   %
  & \val{\bf}{76.77}{0.4223}   %
  & \val{}{85.63}{0.3508}      %
  & \val{\bf}{70.26}{0.4571}   %
  & \val{}{71.12}{0.4532}      %
\\\hline
\multirow{4}{*}{NLPD $\downarrow$} & CLIP~\citep{radford2021learning} & \val{}{1.75}{0.0479} & \val{}{0.48}{0.0083} & \val{\bf}{0.15}{0.0072} & \val{}{0.90}{0.0178} & \val{}{0.58}{0.0174} & \val{}{1.36}{0.0391} & \val{}{1.02}{0.0128}\\
 & CLIP (temp. scaling) & \val{\bf}{1.51}{0.0378} & \val{}{0.47}{0.0065} & \val{\bf}{0.15}{0.0056} & \val{\bf}{0.87}{0.0143} & \val{}{0.59}{0.0144} & \val{\bf}{1.20}{0.0298} & \val{\bf}{0.96}{0.0098}\\
 & TTA~\citep{farina2024frustratingly}  & \val{}{1.74}{0.0472} & \val{}{0.49}{0.0086} & \val{}{0.24}{0.0080} & \val{}{1.15}{0.0188} & \val{\bf}{0.49}{0.0160} & \val{}{1.34}{0.0390} & \val{}{1.06}{0.0133}\\
 \rowcolor{scCyan!30}
\cellcolor{white} & BayesVLM
  & \val{}{1.62}{0.0335}   %
  & \val{\bf}{0.45}{0.0110} %
  & \val{\bf}{0.15}{0.0061} %
  & \val{\bf}{0.87}{0.0155} %
  & \val{}{0.57}{0.0132}    %
  & \val{}{1.22}{0.0204}    %
  & \val{}{0.98}{0.0154}    %
\\\hline
\multirow{4}{*}{ECE $\downarrow$} & CLIP~\citep{radford2021learning} & $9.47$ & $3.07$ & $0.97$ & $5.73$ & $2.13$ & $10.72$ & $8.60$\\ 
 & CLIP (temp. scaling) & $4.90$ & $3.00$ & $1.35$ & $2.55$ & $5.21$ & $\bf3.40$ & $\bf1.48$\\
 & TTA~\citep{farina2024frustratingly}  & $11.96$ & $3.77$ & $1.21$ & $3.92$ & $\bf1.92$ & $11.87$ & $10.53$\\  \rowcolor{scCyan!30}
\cellcolor{white} & BayesVLM
  & $\bf4.66$  %
  & $\bf1.00$  %
  & $\bf0.62$  %
  & $\bf1.91$  %
  & $2.15$     %
  & $5.37$     %
  & $3.89$     %
\\
\bottomrule
\end{tabular}
     }
\end{table*}

\begin{table*}[h]
  \centering
  \small  %
  \caption{{\bfseries Zero-shot results:} Quantitative evaluation of uncertainty estimation across multiple data sets in the zero-shot setting for the SigLIP-Base model~\citep{zhai2023sigmoid}.
  Our proposed BayesVLM often performs competitively to baseline methods across accuracy (ACC, in \%), negative log predictive density (NLPD), and expected calibration error (ECE, in \%) metrics.
  }
  \label{tab:uq_results_siglip}     
  \vspace*{-0.5em} 
  \setlength{\tabcolsep}{9pt}  %
\resizebox{\textwidth}{!}{

\begin{tabular}{l|l|ccccccc}
\toprule
Metrics & Methods & \sc Flowers-102 & \sc Food-101 & \sc CIFAR-10 & \sc CIFAR-100 & \sc ImageNet-R & \sc UCF101 & \sc SUN397 \\
\hline
    \multirow{4}{*}{ACC $\uparrow$} & SigLIP~\citep{zhai2023sigmoid} & \val{\bf}{82.31}{0.4867} & \val{}{88.81}{0.1984} & \val{\bf}{93.20}{0.2517} & \val{\bf}{71.27}{0.4525} & \val{}{89.71}{0.3509} & \val{\bf}{59.61}{0.7978} & \val{\bf}{67.55}{0.3323}\\
 & SigLIP (temp. scaling) & \val{\bf}{82.31}{0.4867} & \val{}{88.81}{0.1984} & \val{\bf}{93.20}{0.2517} & \val{\bf}{71.27}{0.4525} & \val{}{89.71}{0.3509} & \val{\bf}{59.61}{0.7978} & \val{\bf}{67.55}{0.3323}\\
 & TTA~\citep{farina2024frustratingly}  & \val{\bf}{82.66}{0.4828} & \val{\bf}{89.24}{0.1950} & \val{}{87.96}{0.3254} & \val{}{60.95}{0.4879} & \val{\bf}{90.91}{0.3320} & \val{\bf}{60.14}{0.7960} & \val{}{66.82}{0.3342}\\
\rowcolor{scCyan!30} \cellcolor{white} & BayesVLM
  & \val{\bf}{82.44}{0.3805}   %
  & \val{}{88.84}{0.3148}      %
  & \val{\bf}{93.16}{0.2524}   %
  & \val{\bf}{71.22}{0.4527}   %
  & \val{}{89.72}{0.3037}      %
  & \val{\bf}{59.69}{0.4905}   %
  & \val{\bf}{67.44}{0.4686}   %
\\\hline
\multirow{4}{*}{NLPD $\downarrow$} & SigLIP~\citep{zhai2023sigmoid} & \val{}{0.88}{0.0285} & \val{}{0.38}{0.0061} & \val{\bf}{0.21}{0.0063} & \val{\bf}{1.08}{0.0168} & \val{}{0.41}{0.0139} & \val{}{1.90}{0.0438} & \val{}{1.12}{0.0117}\\
 & SigLIP (temp. scaling) & \val{\bf}{0.84}{0.0246} & \val{}{0.40}{0.0054} & \val{}{0.22}{0.0057} & \val{}{1.09}{0.0152} & \val{}{0.43}{0.0125} & \val{\bf}{1.77}{0.0376} & \val{}{1.12}{0.0102}\\
 & TTA~\citep{farina2024frustratingly}  & \val{\bf}{0.85}{0.0276} & \val{\bf}{0.37}{0.0064} & \val{}{0.36}{0.0073} & \val{}{1.62}{0.0209} & \val{\bf}{0.37}{0.0134} & \val{}{1.88}{0.0430} & \val{}{1.18}{0.0124}\\
\rowcolor{scCyan!30}
\cellcolor{white} & BayesVLM
  & \val{}{0.86}{0.0210}   %
  & \val{}{0.39}{0.0091}   %
  & \val{\bf}{0.21}{0.0061}   %
  & \val{\bf}{1.08}{0.0163} %
  & \val{}{0.41}{0.0114}   %
  & \val{}{1.82}{0.0248}   %
  & \val{\bf}{1.12}{0.0154} %
\\\hline
\multirow{4}{*}{ECE $\downarrow$} & SigLIP~\citep{zhai2023sigmoid} & $\bf4.31$ & $1.66$ & $\bf0.92$ & $1.97$ & $\bf1.36$ & $12.72$ & $3.82$\\
 & SigLIP (temp. scaling) & $6.28$ & $5.54$ & $2.94$ & $4.47$ & $4.85$ & $\bf6.69$ & $2.83$\\
 & TTA~\citep{farina2024frustratingly}  & $4.56$ & $\bf0.72$ & $3.14$ & $2.09$ & $1.59$ & $13.47$ & $6.81$\\
\rowcolor{scCyan!30}
\cellcolor{white} & BayesVLM
  & $4.87$  %
  & $3.49$  %
  & $1.38$  %
  & $\bf1.52$  %
  & $2.79$  %
  & $9.60$  %
  & $\bf1.14$  %
\\

\bottomrule
\end{tabular}
     }
\end{table*}

\begin{table*}[h]
  \centering
  \small  %
  \caption{{\bfseries Can \emph{ProbCosine} improve the zero-shot performance of pre-trained probabilistic models? Yes.} Applying \emph{ProbCosine} to the ProLIP~\citep{prolip} model improves zero-shot performance across classification benchmarks and metrics.}
  \label{tab:uq_results_prolip} 
  \vspace*{-0.5em}
  \setlength{\tabcolsep}{9pt}  %
\resizebox{\textwidth}{!}{

\begin{tabular}{l|l|ccccccc}
\toprule
Metrics & Methods & \sc Flowers-102 & \sc Food-101 & \sc CIFAR-10 & \sc CIFAR-100 & \sc ImageNet-R & \sc UCF101 & \sc SUN397 \\
\hline
   & Mean \citep{prolip} & \val{\bf}{77.83}{0.0053} & \val{\bf}{90.38}{0.0019} & \val{\bf}{96.52}{0.0018} & \val{\bf}{82.41}{0.0038} & \val{}{84.76}{0.0042} & \val{\bf}{69.94}{0.0033} & \val{\bf}{65.93}{0.0077}\\
\rowcolor{scCyan!30} \multirow{-2}{*}{\cellcolor{white}ACC $\uparrow$} & ProbCosine (\cref{eq:cosine_sim_dist}) & \val{\bf}{77.74}{0.0053} & \val{\bf}{90.35}{0.0019} & \val{\bf}{96.52}{0.0018} & \val{\bf}{82.48}{0.0038} & \val{\bf}{84.91}{0.0041} & \val{\bf}{69.99}{0.0033} & \val{\bf}{65.82}{0.0077}\\
 \hline
& Mean \citep{prolip} & \val{}{1.36}{0.0411} & \val{\bf}{0.33}{0.0060} & \val{\bf}{0.11}{0.0052} & \val{}{0.64}{0.0148} & \val{\bf}{0.59}{0.0164} & \val{}{1.05}{0.0120} & \val{}{1.28}{0.0316}\\
\rowcolor{scCyan!30} \multirow{-2}{*}{\cellcolor{white}NLPD $\downarrow$} & ProbCosine (\cref{eq:cosine_sim_dist}) & \val{\bf}{1.28}{0.0376} & \val{}{0.34}{0.0055} & \val{\bf}{0.11}{0.0047} & \val{\bf}{0.63}{0.0133} & \val{}{0.60}{0.0151} & \val{\bf}{1.02}{0.0108} & \val{\bf}{1.24}{0.0286}\\
 \hline
 & Mean \citep{prolip} & $5.31$ & $\bf0.79$ & $\bf0.60$ & $3.38$ & $\bf1.08$ & $5.99$ & $7.99$\\
\rowcolor{scCyan!30} \multirow{-2}{*}{\cellcolor{white}ECE $\downarrow$} & ProbCosine (\cref{eq:cosine_sim_dist}) & $\bf3.60$ & $2.57$ & $0.76$ & $\bf1.40$ & $3.53$ & $\bf2.23$ & $\bf4.73$\\

\bottomrule
\end{tabular}
     }
\end{table*}

\begin{figure*}[h]
    \centering

    \pgfplotsset{width=.85\textwidth, height=0.7\textwidth, scale only axis,%
        tick label style={font=\tiny}, axis lines = middle, axis line style={->}, xmax = 200,
        ylabel near ticks, xlabel near ticks, y tick label style={rotate=45}, x tick label style={rotate=45},
        yticklabel style={ /pgf/number format/fixed, /pgf/number format/precision=2},
        scaled y ticks=false,
        xtick = {0, 10, 25, 50, 75, 100, 150, 200}, every axis plot/.append style={thick},
        ylabel = {\scriptsize Weighted ACC $\rightarrow$},
        xlabel = {}, %
        ylabel style={yshift=-0.1cm},
        yticklabel shift={-0.15cm}
    }
    
    \begin{subfigure}[t]{.18\textwidth}
      \raggedleft
      \pgfplotsset{title = {\scriptsize \bfseries OH-Art}}

     \end{subfigure}

    \vspace*{-0.5em}    

    \caption{{\bfseries Active learning results (CLIP Huge):} We present results for EPIG \legendline{scCyan}, BALD \legendline{scBlue}, Entropy (targeted) \legendline{scPurple}, Entropy \legendline{scRed}, Random selection (targeted) \legendline{black}, Random selection \legendline{gray, dashed} on the OfficeHome dataset (OH) and ImageNet variants (IN). We observe that active learning based on our post-hoc uncertainties consistently improves upon random and entropy-based selection.}
    \label{fig:results_activelearning_clip_huge}
\end{figure*}

\begin{figure*}[h]
    \centering

    \pgfplotsset{width=.85\textwidth, height=0.7\textwidth, scale only axis,%
        tick label style={font=\tiny}, axis lines = middle, axis line style={->}, xmax = 200,
        ylabel near ticks, xlabel near ticks, y tick label style={rotate=45}, x tick label style={rotate=45},
        yticklabel style={ /pgf/number format/fixed, /pgf/number format/precision=2},
        scaled y ticks=false,
        xtick = {0, 10, 25, 50, 75, 100, 150, 200}, every axis plot/.append style={thick},
        ylabel = {\scriptsize Weighted ACC $\rightarrow$},
        xlabel = {}, %
        ylabel style={yshift=-0.1cm},
        yticklabel shift={-0.15cm}
    }
    
    \begin{subfigure}[t]{.18\textwidth}
      \raggedleft
      \pgfplotsset{title = {\scriptsize \bfseries OH-Art}}

     \end{subfigure}
    
    \vspace*{-0.5em}  

    \caption{{\bfseries Active learning results (SigLIP Base):} We present results for EPIG \legendline{scCyan}, BALD \legendline{scBlue}, Entropy (targeted) \legendline{scPurple}, Entropy \legendline{scRed}, Random selection (targeted) \legendline{black}, Random selection \legendline{gray, dashed} on the OfficeHome dataset (OH) and ImageNet variants (IN).}
    \label{fig:results_activelearning_siglip_base}
\end{figure*}

\begin{figure*}[h]
    \centering

    \pgfplotsset{
        width=.75\textwidth, 
        height=0.5\textwidth, 
        scale only axis,
        tick label style={font=\tiny}, 
        axis lines = middle, 
        axis line style={->}, 
        xmax = 200,
        ylabel near ticks, xlabel near ticks, y tick label style={rotate=45}, x tick label style={rotate=45},
        yticklabel style={ /pgf/number format/fixed, /pgf/number format/precision=2},
        scaled y ticks=false,
        xtick = {0, 10, 25, 50, 75, 100, 150, 200}, every axis plot/.append style={thick},
        ylabel = {\scriptsize Weighted ACC $\rightarrow$},
        xlabel = {}, %
        ylabel style={yshift=-0.1cm},
        yticklabel shift={-0.15cm}
    }
    
    \begin{subfigure}[t]{.18\textwidth}
      \raggedleft
      \pgfplotsset{title = {\scriptsize \bfseries OH-Art}}
\begin{tikzpicture}
\begin{axis}[
]
\addplot [thick, scCyan]
table {%
10 0.853909492492676
25 0.864197492599487
50 0.86831271648407
75 0.882715940475464
100 0.884773731231689
150 0.882716059684753
200 0.882716059684753
};
\addlegendentry{EPIG wasserstein}
\addplot [thick, scBlue]
table {%
10 0.847736597061157
25 0.85185182094574
50 0.86831271648407
75 0.874485492706299
100 0.888888835906982
150 0.893004059791565
200 0.888888835906982
};
\addlegendentry{BALD (test) wasserstein}

\addplot [thick, scCyan, dashed]
table {%
10 0.853909492492676
25 0.858024716377258
50 0.855967044830322
75 0.870370388031006
100 0.880658388137817
150 0.878600835800171
200 0.872427940368652
};
\addlegendentry{EPIG cosine}
\addplot [thick, scBlue, dashed]
table {%
10 0.841563701629639
25 0.855967044830322
50 0.874485492706299
75 0.876543164253235
100 0.882715940475464
200 0.882716059684753
};
\addlegendentry{BALD (test) cosine}

\legend{}
\end{axis}

\end{tikzpicture}
     \end{subfigure}
    \hfill
    \begin{subfigure}[t]{.18\textwidth}
      \raggedleft
      \pgfplotsset{title = {\scriptsize \bfseries OH-ClipArt}, ylabel = {}, ylabel style={yshift=0}}
\begin{tikzpicture}
\begin{axis}[
]
\addplot [thick, scCyan]
table {%
10 0.745704412460327
25 0.753722786903381
50 0.772050380706787
75 0.778923273086548
100 0.786941647529602
150 0.79266881942749
200 0.809851169586182
};
\addlegendentry{EPIG wasserstein}
\addplot [thick, scBlue]
table {%
10 0.754868268966675
25 0.768614053726196
50 0.767468452453613
75 0.7697594165802
100 0.775486946105957
150 0.785796165466309
200 0.793814420700073
};
\addlegendentry{BALD (test) wasserstein}

\addplot [thick, scCyan, dashed]
table {%
10 0.747995376586914
25 0.760595679283142
50 0.760595679283142
75 0.775486826896667
100 0.788087010383606
150 0.801832795143127
200 0.810996532440186
};
\addlegendentry{EPIG cosine}
\addplot [thick, scBlue, dashed]
table {%
10 0.756013751029968
25 0.760595679283142
50 0.770905017852783
75 0.777777791023254
100 0.783505201339722
150 0.791523456573486
200 0.804123759269714
};
\addlegendentry{BALD (test) cosine}

\legend{}
\end{axis}

\end{tikzpicture}
     \end{subfigure}
    \hfill
    \begin{subfigure}[t]{.18\textwidth}
      \raggedleft
      \pgfplotsset{title = {\scriptsize\bfseries OH-Product}, ylabel = {}}
\begin{tikzpicture}
\begin{axis}[
]
\addplot [thick, scCyan]
table {%
10 0.924549579620361
25 0.934684753417969
50 0.940315246582031
100 0.953828811645508
200 0.958333492279053
};
\addlegendentry{EPIG wasserstein}
\addplot [thick, scBlue]
table {%
10 0.922297239303589
25 0.926801800727844
50 0.938063025474548
75 0.948198080062866
100 0.949324250221252
150 0.954954981803894
200 0.958333253860474
};
\addlegendentry{BALD (test) wasserstein}

\addplot [thick, scCyan, dashed]
table {%
10 0.920045137405396
25 0.935810804367065
50 0.940315246582031
75 0.942567586898804
100 0.948198080062866
150 0.957207202911377
200 0.960585594177246
};
\addlegendentry{EPIG cosine}
\addplot [thick, scBlue, dashed]
table {%
10 0.922297239303589
25 0.930180191993713
50 0.938063144683838
75 0.948198199272156
100 0.948198080062866
150 0.951576471328735
200 0.960585594177246
};
\addlegendentry{BALD (test) cosine}

\legend{}
\end{axis}

\end{tikzpicture}
     \end{subfigure}
    \hfill
    \begin{subfigure}[t]{.18\textwidth}
      \raggedleft
      \pgfplotsset{title = {\scriptsize\bfseries IN-R}, ylabel = {}}
\begin{tikzpicture}
\begin{axis}[
]
\addplot [thick, scCyan]
table {%
50 0.732500076293945
100 0.737499952316284
150 0.741999864578247
200 0.743499994277954
300 0.750999927520752
400 0.757166743278503
500 0.756333351135254
};
\addlegendentry{EPIG wasserstein}
\addplot [thick, scBlue]
table {%
50 0.728000044822693
100 0.734333276748657
150 0.736666679382324
200 0.737499952316284
300 0.743666648864746
400 0.749333381652832
500 0.750166654586792
};
\addlegendentry{BALD (test) wasserstein}

\addplot [thick, scCyan, dashed]
table {%
50 0.736166715621948
100 0.73633337020874
150 0.739333152770996
200 0.739666700363159
300 0.744666695594788
400 0.750166654586792
500 0.75516664981842
};
\addlegendentry{EPIG cosine}
\addplot [thick, scBlue, dashed]
table {%
50 0.728499889373779
100 0.733833312988281
150 0.736833333969116
200 0.737333297729492
300 0.744333267211914
400 0.751499891281128
500 0.753999948501587
};
\addlegendentry{BALD (test) cosine}

\legend{}
\end{axis}

\end{tikzpicture}
     \end{subfigure}
    \hfill
    \begin{subfigure}[t]{.18\textwidth}
      \raggedleft
      \pgfplotsset{title = {\scriptsize\bfseries IN-Sketch}, ylabel = {}}
\begin{tikzpicture}
\begin{axis}[
]
\addplot [thick, scCyan]
table {%
50 0.908466815948486
100 0.918764352798462
150 0.923341035842896
200 0.925629377365112
300 0.931350111961365
400 0.932494282722473
500 0.931350111961365
};
\addlegendentry{EPIG wasserstein}
\addplot [thick, scBlue]
table {%
50 0.909610986709595
100 0.914187669754028
150 0.924485206604004
200 0.925629377365112
300 0.925629377365112
400 0.93478262424469
500 0.93478262424469
};
\addlegendentry{BALD (test) wasserstein}

\addplot [thick, scCyan, dashed]
table {%
50 0.909610986709595
100 0.917620182037354
150 0.924485206604004
200 0.933638453483582
300 0.935926795005798
500 0.938215136528015
};
\addlegendentry{EPIG cosine}
\addplot [thick, scBlue, dashed]
table {%
50 0.908466815948486
100 0.909610986709595
150 0.921052694320679
200 0.921052694320679
300 0.926773548126221
400 0.935926795005798
500 0.937070846557617
};
\addlegendentry{BALD (test) cosine}

\legend{}
\end{axis}

\end{tikzpicture}
     \end{subfigure}
    
    \vspace*{-1.75em}

    \pgfplotsset{
        ylabel = {\scriptsize NLPD $\leftarrow$},
        xlabel = {\scriptsize Subset Size},
        title = {}
    }
    \begin{subfigure}[t]{.18\textwidth}
        \raggedleft
\begin{tikzpicture}
\begin{axis}[
]
\addplot [thick, scCyan]
table {%
10 0.564424276351929
25 0.556679487228394
50 0.503969311714172
75 0.496453166007996
100 0.502012252807617
150 0.491081476211548
200 0.49254846572876
};
\addlegendentry{EPIG wasserstein}
\addplot [thick, scBlue]
table {%
10 0.591114521026611
25 0.547877907752991
50 0.512692213058472
75 0.504598140716553
100 0.472466230392456
150 0.463337659835815
200 0.469667196273804
};
\addlegendentry{BALD (test) wasserstein}

\addplot [thick, scCyan, dashed]
table {%
10 0.586488962173462
25 0.566072940826416
50 0.519615292549133
75 0.509848713874817
100 0.512712001800537
150 0.505711793899536
200 0.506946802139282
};
\addlegendentry{EPIG cosine}
\addplot [thick, scBlue, dashed]
table {%
10 0.596570134162903
25 0.544293761253357
50 0.524789333343506
75 0.501524448394775
100 0.479892015457153
150 0.48262619972229
200 0.491323947906494
};
\addlegendentry{BALD (test) cosine}

\legend{}
\end{axis}

\end{tikzpicture}
     \end{subfigure}
    \pgfplotsset{
        ylabel = {},
    }
    \hfill
    \begin{subfigure}[t]{.18\textwidth}
      \raggedleft
\begin{tikzpicture}
\begin{axis}[
]
\addplot [thick, scCyan]
table {%
10 0.979279041290283
25 0.950736522674561
50 0.870826005935669
75 0.829813241958618
100 0.789095640182495
150 0.73612380027771
200 0.694782137870789
};
\addlegendentry{EPIG wasserstein}
\addplot [thick, scBlue]
table {%
10 0.92957067489624
25 0.911185145378113
50 0.89618444442749
75 0.861356616020203
100 0.820641994476318
150 0.779101967811584
200 0.740799427032471
};
\addlegendentry{BALD (test) wasserstein}

\addplot [thick, scCyan, dashed]
table {%
10 0.979983329772949
25 0.937260866165161
50 0.899082779884338
75 0.83654522895813
100 0.806858777999878
150 0.711852192878723
200 0.70304536819458
};
\addlegendentry{EPIG cosine}
\addplot [thick, scBlue, dashed]
table {%
10 0.932507753372192
25 0.908894777297974
50 0.887145638465881
75 0.859245777130127
100 0.822076201438904
150 0.785528659820557
200 0.730416178703308
};
\addlegendentry{BALD (test) cosine}

\legend{}
\end{axis}

\end{tikzpicture}
     \end{subfigure}
    \hfill
    \begin{subfigure}[t]{.18\textwidth}
      \raggedleft
\begin{tikzpicture}
\begin{axis}[
]
\addplot [thick, scCyan]
table {%
10 0.271091103553772
25 0.250711679458618
50 0.246048808097839
75 0.219184160232544
100 0.170548558235168
150 0.162109136581421
200 0.15934956073761
};
\addlegendentry{EPIG wasserstein}
\addplot [thick, scBlue]
table {%
10 0.280627131462097
25 0.265828609466553
50 0.225194573402405
75 0.2214435338974
100 0.22330105304718
150 0.229358673095703
200 0.169040560722351
};
\addlegendentry{BALD (test) wasserstein}

\addplot [thick, scCyan, dashed]
table {%
10 0.264061212539673
25 0.251425504684448
50 0.232800960540771
75 0.224288821220398
100 0.217337131500244
150 0.158744096755981
200 0.14521324634552
};
\addlegendentry{EPIG cosine}
\addplot [thick, scBlue, dashed]
table {%
10 0.264463901519775
25 0.257837414741516
50 0.226011514663696
75 0.208601593971252
150 0.214790940284729
200 0.152443051338196
};
\addlegendentry{BALD (test) cosine}

\legend{}
\end{axis}

\end{tikzpicture}
     \end{subfigure}
    \hfill
    \begin{subfigure}[t]{.18\textwidth}
      \raggedleft
\begin{tikzpicture}
\begin{axis}[
]
\addplot [thick, scCyan]
table {%
50 1.10605347156525
100 1.06875205039978
150 1.06122159957886
200 1.05446350574493
300 1.03197038173676
400 1.00024354457855
500 0.982278108596802
};
\addlegendentry{EPIG wasserstein}
\addplot [thick, scBlue]
table {%
50 1.13370680809021
100 1.10756802558899
150 1.0868992805481
200 1.0792441368103
300 1.0501264333725
400 1.02029883861542
500 1.01088523864746
};
\addlegendentry{BALD (test) wasserstein}

\addplot [thick, scCyan, dashed]
table {%
50 1.09824335575104
100 1.08387160301208
150 1.05635917186737
200 1.05388712882996
300 1.03476548194885
400 1.01887691020966
500 1.00846600532532
};
\addlegendentry{EPIG cosine}
\addplot [thick, scBlue, dashed]
table {%
50 1.13601529598236
100 1.10418081283569
150 1.08924376964569
200 1.08107721805573
300 1.04451358318329
400 1.01224613189697
500 1.00331032276154
};
\addlegendentry{BALD (test) cosine}

\legend{}
\end{axis}

\end{tikzpicture}
     \end{subfigure}
    \hfill
    \begin{subfigure}[t]{.18\textwidth}
      \raggedleft
\begin{tikzpicture}
\begin{axis}[
]
\addplot [thick, scCyan]
table {%
50 0.389440298080444
100 0.365748405456543
150 0.354536890983582
200 0.346180438995361
300 0.260512590408325
400 0.25058126449585
500 0.250988721847534
};
\addlegendentry{EPIG wasserstein}
\addplot [thick, scBlue]
table {%
50 0.397465229034424
100 0.38827657699585
150 0.361380815505981
200 0.352616906166077
300 0.301052093505859
400 0.275495409965515
500 0.267449975013733
};
\addlegendentry{BALD (test) wasserstein}

\addplot [thick, scCyan, dashed]
table {%
50 0.37133526802063
100 0.369364261627197
150 0.347434282302856
200 0.280353665351868
300 0.26447606086731
400 0.264344811439514
500 0.256273508071899
};
\addlegendentry{EPIG cosine}
\addplot [thick, scBlue, dashed]
table {%
50 0.403494358062744
100 0.38582170009613
150 0.364645600318909
200 0.356545567512512
300 0.30580723285675
400 0.281247615814209
500 0.271267652511597
};
\addlegendentry{BALD (test) cosine}

\legend{}
\end{axis}

\end{tikzpicture}
     \end{subfigure}

    \vspace*{-1em}  

    \caption{Ablation study on the $k$-NN distance metric, fixing the \textit{online LA learning rate} $\gamma = 1\text{e}{-4}$ and the \textit{online LA pseudo-data count} $\beta = 10$. Results are shown for EPIG (Wasserstein) \legendline{scCyan}, EPIG (cosine) \legendline{scCyan, dashed}, BALD (Wasserstein) \legendline{scBlue}, and BALD (cosine) \legendline{scBlue, dashed}. As shown in \cref{fig:results_knn_method_ablation}, Wasserstein-based $k$-NN selection demonstrates improved performance for datasets such as \textit{OH-Art} and \textit{IN-R}, while no clear trend is observed across the other datasets.}
    \label{fig:results_knn_method_ablation}
\end{figure*}

\subsection{Robustness wrt to pseudo-data count}
\label{app:subsec:pseudo_count}

To examine the influence of the pseudo–data count~$\tau$ on BayesVLM,
we evaluated BayesVLM with the CLIP-Base configuration on four datasets
(\textsc{Flowers-102}, \textsc{Food-101}, \textsc{CIFAR-10}, \textsc{CIFAR-100})
while varying $\tau \in \{1,3,5,7,9\}$.
For each setting, we report classification accuracy (ACC~$\uparrow$),
negative log predictive density (NLPD~$\downarrow$),
and expected calibration error (ECE~$\downarrow$)
in \Cref{tab:pseudo_count_results}.

The results show that performance is stable across the tested range of
pseudo–data counts.  
Accuracy and calibration metrics vary only slightly with~$\tau$,
indicating that the method is robust to this hyperparameter.
In the main manuscript, we used $\tau = 4$ as the default setting.
We find that slight improvements can be observed when moving the pseudo–data
count away from $\tau = 4$, but the value found on the proxy dataset
($\tau = 4$) provides reasonable performance in general.

\begin{table}[h]
\centering
\small
\caption{Zero-shot CLIP-Base performance versus pseudo-count on four datasets.}
\label{tab:pseudo_count_results}
\vspace*{-0.5em}
\resizebox{\textwidth}{!}{%
\begin{tabular}{c
                |ccc
                |ccc
                |ccc
                |ccc}
\toprule
\multirow{2}{*}{Pseudo Count} &
\multicolumn{3}{c|}{\textsc{Flowers-102}} &
\multicolumn{3}{c|}{\textsc{Food-101}} &
\multicolumn{3}{c|}{\textsc{CIFAR-10}} &
\multicolumn{3}{c}{\textsc{CIFAR-100}} \\
& ACC$\uparrow$ & NLPD$\downarrow$ & ECE$\downarrow$
& ACC$\uparrow$ & NLPD$\downarrow$ & ECE$\downarrow$
& ACC$\uparrow$ & NLPD$\downarrow$ & ECE$\downarrow$
& ACC$\uparrow$ & NLPD$\downarrow$ & ECE$\downarrow$ \\
\midrule
1 & 69.04$\pm$0.59 & 1.75$\pm$0.04 & 3.98
  & 80.62$\pm$0.25 & 0.67$\pm$0.01 & 0.85
  & 93.58$\pm$0.25 & 0.20$\pm$0.01 & 0.70
  & 73.82$\pm$0.44 & 0.95$\pm$0.02 & 2.58 \\
3 & 69.43$\pm$0.59 & 1.81$\pm$0.05 & 3.78
  & 80.45$\pm$0.25 & 0.68$\pm$0.01 & 1.67
  & 93.60$\pm$0.24 & 0.20$\pm$0.01 & 1.10
  & 73.80$\pm$0.44 & 0.95$\pm$0.02 & 4.31 \\
5 & 69.54$\pm$0.59 & 1.83$\pm$0.05 & 3.97
  & 80.37$\pm$0.25 & 0.68$\pm$0.01 & 2.31
  & 93.60$\pm$0.24 & 0.20$\pm$0.01 & 1.20
  & 73.82$\pm$0.44 & 0.96$\pm$0.02 & 4.82 \\
7 & 69.36$\pm$0.59 & 1.84$\pm$0.05 & 4.50
  & 80.34$\pm$0.25 & 0.68$\pm$0.01 & 2.63
  & 93.60$\pm$0.24 & 0.21$\pm$0.01 & 1.25
  & 73.81$\pm$0.44 & 0.96$\pm$0.02 & 5.10 \\
9 & 69.43$\pm$0.59 & 1.85$\pm$0.05 & 4.60
  & 80.33$\pm$0.25 & 0.68$\pm$0.01 & 2.81
  & 93.61$\pm$0.24 & 0.21$\pm$0.01 & 1.27
  & 73.79$\pm$0.44 & 0.96$\pm$0.02 & 5.29 \\
\bottomrule
\end{tabular}%
}
\end{table}

\subsection{Robustness wrt the Number of Negative Samples}
\label{app:subsec:batch_size}
To assess the impact of negative samples on the likelihood approximation, we vary the batch size $K \in \{32768, 8192, 2048\}$ and estimate the posterior from 1–5 random batches, reporting mean and standard deviation over five trials.
Because the posterior depends on the number of negatives only through the Hessian $\bB$-factor (\cf~\cref{eq:kfacggn}),
we present the normalised trace $(\frac{\tr(\bB_{i\times K})}{\tr(\bB_{5\times K})})$.
Ideally, this ratio equals one with zero variance.
Results for the image and text surrogate models are shown in
\cref{app:fig:ablation_batch_size_image} and \cref{app:fig:ablation_batch_size_text}, respectively.

The batch size $32768$ yields mean values near one for all numbers of random batches and maintains low standard deviations, indicating reliable estimates for both modalities.

\begin{figure}[h]
\centering
\begin{subfigure}[t]{0.48\textwidth}
\centering
    \setlength{\figurewidth}{.85\textwidth}
    \setlength{\figureheight}{.3\textwidth}
    \pgfplotsset{%
        scale only axis,
        tick label style={font=\tiny}, axis lines = middle, axis line style={->}, 
        ylabel near ticks, 
        xlabel near ticks, 
        y tick label style={rotate=45}, 
        yticklabel style={/pgf/number format/fixed, 
                          /pgf/number format/precision=2,
                          /pgf/number format/fixed zerofill},
        scaled y ticks=false,
        xticklabels=\empty,
        every axis plot/.append style={thick},
        ylabel style={yshift=-0.1cm},
        xlabel style={yshift=+0.1cm},
        yticklabel shift={-0.15cm},
        xticklabel style={scale=.8},
        yticklabel style={scale=.8},
        xlabel={\scriptsize\# Datapoints ($i \times K$)},
        ylabel={\scriptsize$\frac{\tr(\bB_{i\times K})}{\tr(\bB_{5\times K})}$},
        xmin=0.8, xmax=5.2,
        ymin=0.96, ymax=1.015,
        xtick={1,2,3,4,5},
        xticklabels={1$\times$K,2$\times$K,3$\times$K,4$\times$K,5$\times$K},
    }
    \centering
\begin{tikzpicture}

\begin{axis}[
height=\figureheight,
width=\figurewidth,
]

\addplot[
    scRed, dashed, very thick
]
    coordinates {
        (1,0.971)
        (2,0.978)
        (3,0.989) 
        (4,0.993) 
        (5,1.000) 
    };
    
\addplot[
    scRed, name path=A, thin
]
    coordinates {
        (1,0.971 + 0.0436)
        (2,0.978 + 0.0140)
        (3,0.989 + 0.0117)
        (4,0.993 + 0.0233)
        (5,1.000 + 0.0189)
    };

\addplot[
    scRed, name path=B, thin
]
    coordinates {
        (1,0.971 - 0.0436)
        (2,0.978 - 0.0140)
        (3,0.989 - 0.0117)
        (4,0.993 - 0.0233)
        (5,1.000 - 0.0189)
    };

\addplot [scRed, opacity=0.1] fill between [of=A and B];

\addplot[
    scYellow, name path=A, thin
]
    coordinates {
        (1,0.991 + 0.0163)
        (2,0.998 + 0.00847)
        (3,0.998 + 0.0108)
        (4,1.002 + 0.00798)
        (5,1.000 + 0.00549)
    };
    
\addplot[
    scYellow, name path=B, thin
]
    coordinates {
        (1,0.991 - 0.0163)
        (2,0.998 - 0.00847)
        (3,0.998 - 0.0108)
        (4,1.002 - 0.00798)
        (5,1.000 - 0.00549)
    };
    
\addplot[
    scYellow, very thick, dash dot
]
    coordinates {
        (1,0.991) 
        (2,0.998)
        (3,0.998)
        (4,1.002)
        (5,1.000)
    };
\addplot [scYellow, opacity=0.1] fill between [of=A and B];

\addplot[
    scCyan, name path=A, thin
]
    coordinates {
        (1,1.003 + 0.00712)
        (2,1.002 + 0.00520)
        (3,1.000 + 0.00379)
        (4,1.000 + 0.00360)
        (5,1.000 + 0.00282)
    };
    
\addplot[
    scCyan, name path=B, thin
]
    coordinates {
        (1,1.003 - 0.00712)
        (2,1.002 - 0.00520)
        (3,1.000 - 0.00379)
        (4,1.000 - 0.00360)
        (5,1.000 - 0.00282)
    };
    
\addplot[
    scCyan, very thick
]
    coordinates {
        (1,1.003) 
        (2,1.002) 
        (3,1.000) 
        (4,1.000) 
        (5,1.000) 
    }; 
    
\addplot [scCyan, opacity=0.1] fill between [of=A and B];

\end{axis}
\end{tikzpicture}     \vspace{-1em}
    \subcaption{Image modality}
    \label{app:fig:ablation_batch_size_image}
\end{subfigure}\hfill
\begin{subfigure}[t]{0.48\textwidth}
\centering
    \setlength{\figurewidth}{.85\textwidth}
    \setlength{\figureheight}{.3\textwidth}
    \pgfplotsset{%
        scale only axis,
        tick label style={font=\tiny}, axis lines = middle, axis line style={->}, 
        ylabel near ticks, 
        xlabel near ticks, 
        y tick label style={rotate=45}, 
        yticklabel style={/pgf/number format/fixed, 
                          /pgf/number format/precision=2,
                          /pgf/number format/fixed zerofill},
        scaled y ticks=false,
        xticklabels=\empty,
        every axis plot/.append style={thick},
        ylabel style={yshift=-0.1cm},
        xlabel style={yshift=+0.1cm},
        yticklabel shift={-0.15cm},
        xticklabel style={scale=.8},
        yticklabel style={scale=.8},
        xlabel={\scriptsize\# Datapoints ($i \times K$)},
        ylabel={\scriptsize$\frac{\tr(\bB_{i\times K})}{\tr(\bB_{5\times K})}$},
        xmin=0.8, xmax=5.2,
        ymin=0.95, ymax=1.05,
        xtick={1,2,3,4,5},
        xticklabels={1$\times$K,2$\times$K,3$\times$K,4$\times$K,5$\times$K},
    }
    \centering
\begin{tikzpicture}
\begin{axis}[
    height=\figureheight,
    width=\figurewidth,
    ymin=0.95, ymax=1.05,
]

\addplot[scRed, very thick, dashed]
coordinates {
  (1,0.981) (2,0.996) (3,1.003) (4,0.996) (5,1.000)
};
\addplot[scRed, name path=Rhi, thin]
coordinates {
  (1,0.981+0.0871) (2,0.996+0.0518) (3,1.003+0.0362)
  (4,0.996+0.0371) (5,1.000+0.0277)
};
\addplot[scRed, name path=Rlo, thin]
coordinates {
  (1,0.981-0.0871) (2,0.996-0.0518) (3,1.003-0.0362)
  (4,0.996-0.0371) (5,1.000-0.0277)
};
\addplot[scRed, opacity=0.1] fill between[of=Rhi and Rlo];

\addplot[scYellow, very thick, dash dot]
coordinates {
  (1,0.992) (2,0.992) (3,0.994) (4,0.998) (5,1.000)
};
\addplot[scYellow, name path=Yhi, thin]
coordinates {
  (1,0.992+0.0209) (2,0.992+0.0157) (3,0.994+0.0123)
  (4,0.998+0.00937) (5,1.000+0.00546)
};
\addplot[scYellow, name path=Ylo, thin]
coordinates {
  (1,0.992-0.0209) (2,0.992-0.0157) (3,0.994-0.0123)
  (4,0.998-0.00937) (5,1.000-0.00546)
};
\addplot[scYellow, opacity=0.1] fill between[of=Yhi and Ylo];

\addplot[scCyan, very thick]
coordinates {
  (1,1.006) (2,1.005) (3,1.003) (4,1.001) (5,1.000)
};
\addplot[scCyan, name path=Chi, thin]
coordinates {
  (1,1.006+0.00447) (2,1.005+0.00448) (3,1.003+0.00457)
  (4,1.001+0.00345) (5,1.000+0.00241)
};
\addplot[scCyan, name path=Clo, thin]
coordinates {
  (1,1.006-0.00447) (2,1.005-0.00448) (3,1.003-0.00457)
  (4,1.001-0.00345) (5,1.000-0.00241)
};
\addplot[scCyan, opacity=0.1] fill between[of=Chi and Clo];

\end{axis}
\end{tikzpicture}
     \vspace{-1em}
    \subcaption{Text modality}
    \label{app:fig:ablation_batch_size_text}
\end{subfigure}
    \caption{Normalized trace of the image Hessian $B$-factor for varying base batch sizes
$K$ (2048 \legendline{scRed, dashed}, 8192 \legendline{scYellow, dash dot},
32768 \legendline{scCyan}) and 1–5 random batches.
Error bars show $\pm1$\,std over five trials.}
\label{app:fig:ablation_batch_size}
\end{figure}

\subsection{Additional experiments on CC12M}
We find that BayesVLM provided robust uncertainty estimates for CLIP even when estimated on the proxy dataset, cf., \Cref{app:tab:uq_results_base_laion_vs_cc12m}.
\begin{table*}[h]
  \centering
  \small  %
  \caption{\textbf{Does BayesVLM work in closed-source data settings? Yes.} With OpenCLIP ViT-B-32 trained on LAION-400M and BayesVLM estimated on the proxy dataset CC12M, we find that results are robust and show only slight degradation; statistically significant differences are \textbf{bold} ($p=0.05$).} 
  \label{app:tab:uq_results_base_laion_vs_cc12m}     
  \setlength{\tabcolsep}{9pt}  %
\resizebox{\textwidth}{!}{
\begin{tabular}{l|l|ccccccc}
\toprule
Metrics & Dataset & \sc Flowers-102 & \sc Food-101 & \sc CIFAR-10 & \sc CIFAR-100 & \sc ImageNet-R & \sc UCF101 & \sc SUN397 \\
\hline
\multirow{2}{*}{ACC $\uparrow$}
  & LAION-400M
    & \val{\bf}{68.87}{0.4630} & \val{}{80.43}{0.3968} & \val{}{93.62}{0.2444} & \val{}{73.63}{0.4406} & \val{}{74.45}{0.4361} & \val{}{61.43}{0.4868} & \val{}{66.96}{0.4703} \\
  & CC12M
    & \val{}{68.12}{0.4660} & \val{}{80.35}{0.3974} & \val{}{93.57}{0.2453} & \val{}{73.78}{0.4398} & \val{}{74.32}{0.4369} & \val{}{61.46}{0.4867} & \val{}{66.81}{0.4709} \\
\hline
\multirow{2}{*}{NLPD $\downarrow$}
  & LAION-400M
    & \val{\bf}{1.73}{0.0320} & \val{}{0.68}{0.0126} & \val{}{0.20}{0.0067} & \val{}{0.95}{0.0152} & \val{\bf}{1.03}{0.0177} & \val{}{1.44}{0.0183} & \val{\bf}{1.12}{0.0155} \\
  & CC12M
    & \val{}{1.77}{0.0330} & \val{}{0.68}{0.0129} & \val{}{0.20}{0.0067} & \val{}{0.95}{0.0152} & \val{}{1.03}{0.0180} & \val{}{1.44}{0.0185} & \val{}{1.13}{0.0162} \\
\hline
\multirow{2}{*}{ECE $\downarrow$}
  & LAION-400M
    & $4.22$ & $1.69$ & $0.72$ & $1.92$ & $1.78$ & $\bf3.77$ & $\bf2.06$ \\
  & CC12M
    & $\bf3.84$ & $\bf0.99$ & $\bf0.70$ & $\bf1.43$ & $\bf1.39$ & $3.83$ & $3.89$\\
\bottomrule
\end{tabular} }
\end{table*}

\subsection{Robustness \wrt distribution shift of the Proxy Dataset}
\label{app:subsec:proxy_shift}
Most OpenCLIP models are trained on the \textsc{LAION-400M} dataset, which allows us to directly estimate the Hessians on this distribution. However, some large CLIP variants are trained on closed-source data, requiring the use of a proxy dataset. As shown in \Cref{app:tab:uq_results_base_laion_vs_cc12m}, BayesVLM remains effective in a simulated closed-source setting, where we estimate the Hessians using the CC12M dataset. 

\begin{figure}[h]
\centering

\setlength{\tabcolsep}{2pt}
\renewcommand{\arraystretch}{1.0}

\begin{tabular}{c c c c c c c}

\rotatebox{90}{\small\textsc{Grayscale}} & 
\includegraphics[width=0.14\linewidth]{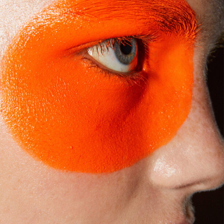} &
\includegraphics[width=0.14\linewidth]{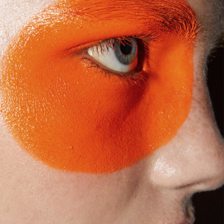} &
\includegraphics[width=0.14\linewidth]{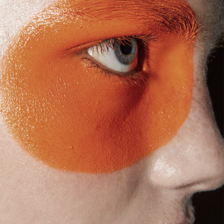} &
\includegraphics[width=0.14\linewidth]{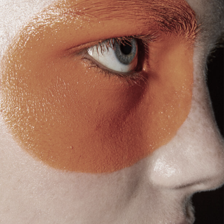} &
\includegraphics[width=0.14\linewidth]{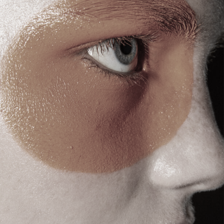} &
\includegraphics[width=0.14\linewidth]{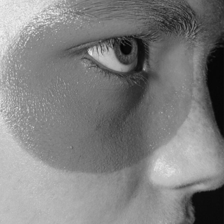} \\
[-3pt]
& \scriptsize 0.0 & \scriptsize 0.2 & \scriptsize 0.4 &
  \scriptsize 0.6 & \scriptsize 0.8 & \scriptsize 1.0 \\
\rotatebox{90}{\small\textsc{Compression}} &
\includegraphics[width=0.14\linewidth]{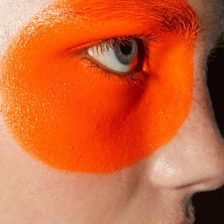} &
\includegraphics[width=0.14\linewidth]{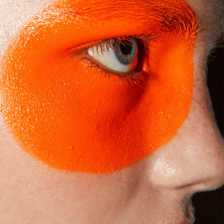} &
\includegraphics[width=0.14\linewidth]{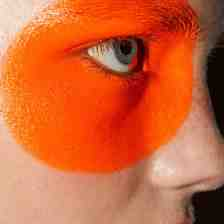} &
\includegraphics[width=0.14\linewidth]{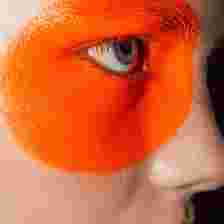} &
\includegraphics[width=0.14\linewidth]{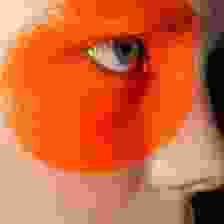} &
\includegraphics[width=0.14\linewidth]{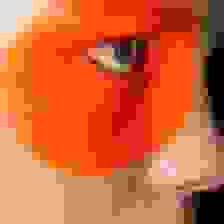} \\
[-3pt]
& \scriptsize 100 & \scriptsize 50 & \scriptsize 25 &
  \scriptsize 10 & \scriptsize 5 & \scriptsize 1 \\

\rotatebox{90}{\small\textsc{Blur}} &
\includegraphics[width=0.14\linewidth]{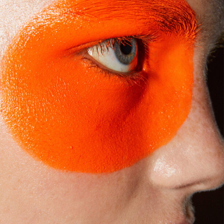} &
\includegraphics[width=0.14\linewidth]{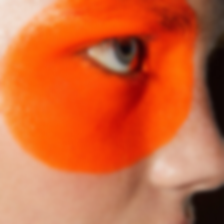} &
\includegraphics[width=0.14\linewidth]{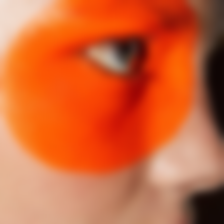} &
\includegraphics[width=0.14\linewidth]{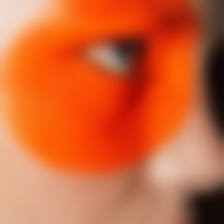} &
\includegraphics[width=0.14\linewidth]{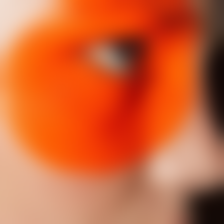} &
\includegraphics[width=0.14\linewidth]{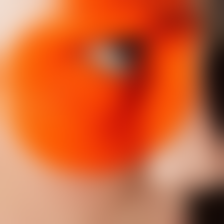} \\
[-3pt]
& \scriptsize 0 & \scriptsize 10 & \scriptsize 20 &
  \scriptsize 30 & \scriptsize 40 & \scriptsize 50 \\

\end{tabular}

\caption{Visualisation of proxy-data augmentations across degradation levels. Rows correspond to augmentation types (\textsc{Grayscale}, \textsc{JPEG Compression}, \textsc{Gaussian Blur}). Columns show increasing degradation severity, parameterised by intensity, JPEG quality, or blur radius, respectively.}
\label{app:fig:la_augmentations}
\end{figure}

To further analyse robustness, we conduct additional experiments in which we progressively distort the images of the \textsc{LAION-400M} dataset using different augmentations before estimating the Laplace Hessians: \textsc{Grayscale} (interpolating between the RGB variant and the Grayscaled variant with intensity coefficients in $
\{0, 0.2, 0.4, 0.6, 0.8, 1.0\}$), \textsc{JPEG Compression} (using increasingly lower JPEG compression quality values in $\{100, 50, 25, 10, 5, 1\}$, and \textsc{Gaussian Blur} (with increasing radius in $\{0, 10, 20, 30, 40, 50\}$). We provide examples in~\Cref{app:fig:la_augmentations}). This setup simulates a controlled proxy-distribution shift. Importantly, all zero-shot evaluations are performed on the original (non-augmented) benchmarks.

The zero-shot results on \textsc{Food-101}, \textsc{CIFAR-10}, and \textsc{ImageNet-R} in \Cref{app:tab:food_aug,app:tab:cifar10_aug,app:tab:imagenet_r_aug} show that augmentations that alter image style while preserving semantic content lead to a gradual increase in ECE that closely tracks the augmentation intensity. In contrast, \textsc{Gaussian Blur} (which directly impairs object recognizability, especially at intensities 0.6 to 1.0) causes substantially stronger degradation in both ECE and NLPD. Overall, these findings indicate that BayesVLM is robust to mild stylistic shifts in the proxy data, but its performance deteriorates once the augmentations begin to compromise semantic information relevant to the model.

\begin{table}[h]
\centering
\small
\caption{Zero-shot CLIP-Base performance on \textsc{Food-101} versus augmentation intensity for three augmentation types.}
\label{app:tab:food_aug}
\vspace*{-0.5em}
\resizebox{\textwidth}{!}{%
\begin{tabular}{
    cccc | cccc | cccc
}
\toprule
\multicolumn{4}{c|}{\textsc{Grayscale}} &
\multicolumn{4}{c|}{\textsc{JPEG Compression}} &
\multicolumn{4}{c}{\textsc{Gaussian Blur}} \\
Intensity & ACC$\uparrow$ & NLPD$\downarrow$ & ECE$\downarrow$
& Quality & ACC$\uparrow$ & NLPD$\downarrow$ & ECE$\downarrow$
& Radius & ACC$\uparrow$ & NLPD$\downarrow$ & ECE$\downarrow$ \\
\midrule
0.0 & 80.44$\pm$0.397 & 0.68$\pm$0.013 & 1.69
    & 100 & 80.44$\pm$0.397 & 0.68$\pm$0.013 & 1.69
    & 0 & 80.44$\pm$0.397 & 0.68$\pm$0.013 & 1.69 \\

0.2 & 80.45$\pm$0.397 & 0.68$\pm$0.013 & 1.71
    & 50 & 80.48$\pm$0.396 & 0.68$\pm$0.012 & 2.21
    & 10 & 80.61$\pm$0.395 & 0.68$\pm$0.012 & 2.38 \\

0.4 & 80.43$\pm$0.397 & 0.68$\pm$0.013 & 1.73
    & 25 & 80.54$\pm$0.396 & 0.68$\pm$0.012 & 2.26
    & 20 & 80.58$\pm$0.396 & 0.68$\pm$0.012 & 3.25 \\

0.6 & 80.48$\pm$0.396 & 0.68$\pm$0.013 & 1.85
    & 10 & 80.59$\pm$0.396 & 0.68$\pm$0.012 & 2.23
    & 30 & 80.48$\pm$0.396 & 0.69$\pm$0.011 & 5.69 \\

0.8 & 80.54$\pm$0.396 & 0.68$\pm$0.012 & 2.06
    & 5 & 80.58$\pm$0.396 & 0.68$\pm$0.012 & 2.74
    & 40 & 80.43$\pm$0.397 & 0.72$\pm$0.011 & 8.93 \\

1.0 & 80.56$\pm$0.396 & 0.68$\pm$0.012 & 2.62
    & 1 & 80.51$\pm$0.396 & 0.68$\pm$0.012 & 4.38
    & 50 & 80.40$\pm$0.397 & 0.75$\pm$0.011 & 11.52 \\
\bottomrule
\end{tabular}
}
\end{table}

\begin{table}[h]
\centering
\small
\caption{Zero-shot CLIP-Base performance on \textsc{CIFAR-10} versus augmentation intensity for three augmentation types.}
\label{app:tab:cifar10_aug}
\vspace*{-0.5em}
\resizebox{\textwidth}{!}{%
\begin{tabular}{cccc | cccc | cccc}
\toprule
\multicolumn{4}{c|}{\textsc{Grayscale}} &
\multicolumn{4}{c|}{\textsc{JPEG Compression}} &
\multicolumn{4}{c}{\textsc{Gaussian Blur}} \\
Intensity & ACC$\uparrow$ & NLPD$\downarrow$ & ECE$\downarrow$
& Quality & ACC$\uparrow$ & NLPD$\downarrow$ & ECE$\downarrow$
& Radius & ACC$\uparrow$ & NLPD$\downarrow$ & ECE$\downarrow$ \\
\midrule

0.0 & 93.61$\pm$0.245 & 0.203$\pm$0.007 & 0.72
    & 100 & 93.61$\pm$0.245 & 0.203$\pm$0.007 & 0.72
    & 0 & 93.61$\pm$0.245 & 0.203$\pm$0.007 & 0.72 \\

0.2 & 93.61$\pm$0.245 & 0.203$\pm$0.007 & 0.72
    & 50 & 93.60$\pm$0.245 & 0.203$\pm$0.007 & 1.02
    & 10 & 93.57$\pm$0.245 & 0.203$\pm$0.007 & 0.72 \\

0.4 & 93.61$\pm$0.245 & 0.203$\pm$0.007 & 0.71
    & 25 & 93.59$\pm$0.245 & 0.203$\pm$0.007 & 0.96
    & 20 & 93.57$\pm$0.245 & 0.203$\pm$0.007 & 0.65 \\

0.6 & 93.61$\pm$0.245 & 0.203$\pm$0.007 & 0.68
    & 10 & 93.59$\pm$0.245 & 0.203$\pm$0.007 & 0.92
    & 30 & 93.58$\pm$0.245 & 0.204$\pm$0.007 & 0.84 \\

0.8 & 93.58$\pm$0.245 & 0.203$\pm$0.007 & 0.69
    & 5 & 93.57$\pm$0.245 & 0.203$\pm$0.007 & 0.89
    & 40 & 93.58$\pm$0.245 & 0.207$\pm$0.007 & 1.88 \\

1.0 & 93.58$\pm$0.245 & 0.203$\pm$0.007 & 0.80
    & 1 & 93.60$\pm$0.245 & 0.205$\pm$0.007 & 1.35
    & 50 & 93.58$\pm$0.245 & 0.212$\pm$0.007 & 2.73 \\

\bottomrule
\end{tabular}
}
\end{table}

\begin{table}[h]
\centering
\small
\caption{Zero-shot CLIP-Base performance on \textsc{ImageNet-R} versus augmentation intensity for three augmentation types.}
\label{app:tab:imagenet_r_aug}
\vspace*{-0.5em}
\resizebox{\textwidth}{!}{
\begin{tabular}{cccc | cccc | cccc}
\toprule
\multicolumn{4}{c|}{\textsc{Grayscale}} &
\multicolumn{4}{c|}{\textsc{JPEG Compression}} &
\multicolumn{4}{c}{\textsc{Gaussian Blur}} \\
Intensity & ACC$\uparrow$ & NLPD$\downarrow$ & ECE$\downarrow$
& Quality & ACC$\uparrow$ & NLPD$\downarrow$ & ECE$\downarrow$
& Radius & ACC$\uparrow$ & NLPD$\downarrow$ & ECE$\downarrow$ \\
\midrule

0.0 & 74.49$\pm$0.436 & 1.031$\pm$0.018 & 1.71
    & 100 & 74.48$\pm$0.436 & 1.031$\pm$0.018 & 1.71
    & 0 & 74.49$\pm$0.436 & 1.031$\pm$0.018 & 1.71 \\

0.2 & 74.51$\pm$0.436 & 1.031$\pm$0.018 & 1.70
    & 50 & 74.53$\pm$0.436 & 1.031$\pm$0.017 & 2.05
    & 10 & 74.67$\pm$0.435 & 1.029$\pm$0.017 & 2.04 \\

0.4 & 74.51$\pm$0.436 & 1.031$\pm$0.018 & 1.61
    & 25 & 74.56$\pm$0.436 & 1.031$\pm$0.017 & 1.93
    & 20 & 74.69$\pm$0.435 & 1.031$\pm$0.017 & 3.01 \\

0.6 & 74.48$\pm$0.436 & 1.030$\pm$0.018 & 1.72
    & 10 & 74.63$\pm$0.436 & 1.030$\pm$0.018 & 1.94
    & 30 & 74.76$\pm$0.434 & 1.046$\pm$0.016 & 5.70 \\

0.8 & 74.47$\pm$0.436 & 1.030$\pm$0.018 & 1.89
    & 5 & 74.47$\pm$0.436 & 1.030$\pm$0.018 & 1.89
    & 40 & 74.71$\pm$0.435 & 1.077$\pm$0.016 & 9.28 \\

1.0 & 74.56$\pm$0.436 & 1.029$\pm$0.018 & 1.95
    & 1 & 74.72$\pm$0.435 & 1.032$\pm$0.017 & 3.23
    & 50 & 74.75$\pm$0.434 & 1.109$\pm$0.015 & 12.20 \\

\bottomrule
\end{tabular}
}
\end{table}

\subsection{Additional Zero-shot Results on Adversarial ImageNet Variants}
\label{app:subsec:zeroshot_imagenet_adversarial}
To further assess the robustness of BayesVLM under substantial distribution shift, we evaluate the CLIP-Base model on two challenging adversarial ImageNet variants: \textsc{ImageNet-A} and \textsc{ImageNet-Sketch}. These datasets introduce a severe covariate shift. \textsc{ImageNet-A} contains naturally adversarial, small, or off-centre objects, whereas \textsc{ImageNet-Sketch} imposes strong stylistic changes that alter low-level statistics.

\Cref{app:tab:zero_shot_adversarial_imgnet} reports zero-shot accuracy, negative log predictive density (NLPD), and expected calibration error (ECE) for CLIP, temperature scaling, test-time augmentation (TTA), and BayesVLM. Across both datasets, BayesVLM does not show signs of epistemic underestimation. On \textsc{ImageNet-A}, BayesVLM matches TTA in terms of calibration (ECE~0.23 vs.\ 0.23) while improving over CLIP. On \textsc{ImageNet-Sketch}, BayesVLM achieves the best calibration among all methods (ECE~0.08), outperforming temperature scaling and TTA. Although the NLPD on \textsc{ImageNet-A} is slightly higher than with temperature scaling, BayesVLM still provides an improvement over CLIP.

Notably, TTA yields large accuracy gains, especially on \textsc{ImageNet-A}, which likely arise from improved viewpoint coverage rather than improved uncertainty modelling: using 64 crops increases the likelihood of capturing an informative region of the image. In contrast, BayesVLM preserves single-pass inference efficiency while improving predictive calibration. Since BayesVLM is complementary to TTA, both approaches can be combined to obtain improvements in accuracy and uncertainty quality simultaneously.

\begin{table}[h]
\centering
\small
\caption{Zero-shot performance on \textsc{ImageNet-A} and \textsc{ImageNet-Sketch}.}
\label{app:tab:zero_shot_adversarial_imgnet}
\vspace{0.5em}
\begin{tabular}{lccc|ccc}
\toprule
& \multicolumn{3}{c}{\textsc{ImageNet-A}}
& \multicolumn{3}{c}{\textsc{ImageNet-Sketch}} \\
\cmidrule(lr){2-4} \cmidrule(lr){5-7}
Method 
& Acc (\%) $\uparrow$ & NLPD $\downarrow$ & ECE $\downarrow$
& Acc (\%) $\uparrow$ & NLPD $\downarrow$ & ECE $\downarrow$ \\
\midrule

CLIP~\citep{radford2021learning}
& \val{}{25.87}{0.4379} & \val{}{3.77}{0.0312} & 0.33
& \val{}{50.98}{0.4999} & \val{}{2.37}{0.0303} & 0.16 \\

CLIP (temp. scaling)
& \val{}{25.87}{0.4379} & \val{}{3.22}{0.0226} & 0.18
& \val{}{50.98}{0.4999} & \val{}{2.29}{0.0286} & 0.13 \\

TTA~\citep{farina2024frustratingly}
& \val{}{38.67}{0.4870} & \val{}{2.77}{0.0279} & 0.23
& \val{}{52.40}{0.4994} & \val{}{2.36}{0.0300} & 0.15 \\

\rowcolor{scCyan!30}
BayesVLM
& \val{}{26.67}{0.4422} & \val{}{3.36}{0.0257} & 0.23
& \val{}{50.59}{0.5000} & \val{}{2.24}{0.0259} & 0.08 \\

\bottomrule
\end{tabular}
\end{table}

\subsection{Empirical Assessment of the Independence Assumption}
\label{app:subsec:independence_assumption}

Our approach relies on an independence assumption between the parameters of the image and text projection layers when constructing the surrogate probabilistic models. This assumption follows from adopting modality-wise surrogate models satisfying the common \iid assumption and is crucial for maintaining computational tractability. As detailed in \cref{app:exact-model}, relaxing this assumption would require estimating and storing the full joint Hessian of the projection parameters, which is computationally infeasible in practice.

To empirically assess the strength of cross-modal dependence in the loss curvature, we estimate the relative magnitude of the off-diagonal Hessian block $\bH_{PQ}$ compared to the modality-specific blocks $\bH_{PP}$ and $\bH_{QQ}$. Specifically, we compute the ratio of Frobenius norms
\begin{equation}
R = 
\frac{\|\bH_{PQ}\|_F}
{\tfrac{1}{2}\left(\|\bH_{PP}\|_F + \|\bH_{QQ}\|_F\right)},
\end{equation}
where $R \approx 1$ would indicate strong cross-modal dependence, and smaller values correspond to weaker coupling in the curvature.

Computing the full Hessian is intractable due to its size ($\mathbb{R}^{512^2 \times 512^2}$, approximately $1.6$~TiB in Float32). We therefore approximate the curvature using a Monte Carlo estimate of randomly sampled gradient components per block. The Hessian of the contrastive loss with respect to the projection parameters is approximated via the empirical Fisher, $\bH \approx \mathbb{E}[g g^\top]$, where $g$ denotes the per-batch gradient. While the empirical Fisher is less accurate than the GGN approximation used in the main method, it provides a computationally feasible proxy for estimating relative curvature magnitudes.

For each Hessian block, we estimate the Frobenius norm by averaging the squared magnitudes of the sampled entries. The ratio $R$ is computed across three random seeds and multiple sample sizes. The results are reported in \cref{tab:independence_ratio}.

\begin{table}[h]
\centering
\caption{Estimated curvature ratio $R$ for different numbers of sampled gradient components.}
\label{tab:independence_ratio}
\begin{tabular}{lc}
\toprule
Number of random samples & $R$ \\
\midrule
50k  & $0.363 \pm 0.002$ \\
100k & $0.374 \pm 0.035$ \\
\bottomrule
\end{tabular}
\end{table}

The results indicate only moderate cross-modal dependence in the Hessian (loss curvature). Hence, while secondary interactions between modalities exist, the dominant curvature structure is captured by the modality-wise blocks. This provides empirical support for the independence approximation adopted in BayesVLM.
Importantly, the approximation does not eliminate the multimodal coupling learned during pre-training. The curvature is evaluated at the MAP solution obtained via multimodal contrastive learning, and the resulting posterior still reflects the cross-modal alignment encoded in the model. In practice, our method replaces the exact posterior with a tractable approximation that preserves the model's principal geometric structure while discarding interaction terms of moderate magnitude.
Finally, the entanglement of modalities in the resulting uncertainties can be qualitatively observed in \cref{app:subsec:interpreting_cossim}, where corruption of one modality affects the predictive uncertainty of the joint model, indicating that cross-modal dependencies remain reflected in the predictive distribution. 
\end{document}